%% file: mean-shift-review.tex
\documentclass[letterpaper]{article}
\usepackage[margin=1in,dvips]{geometry}
\usepackage{graphicx,psfrag,amsmath,amsthm,amssymb}
\usepackage{url}
\usepackage{tipa}
\usepackage[numbers,sort&compress]{natbib}

\input{MACPdef}
\input{MACPdef-ams}

\graphicspath{{grf/}}

\AtBeginDvi{}

\title{A review of mean-shift algorithms for clustering\thanks{To appear in: \emph{CRC Handbook of Cluster Analysis}, edited by Roberto Rocci, Fionn Murtagh, Marina Meil\u{a} and Christian Hennig.}}
\author{Miguel {\'A}.\ Carreira-Perpi{\~n}{\'a}n \\
  Electrical Engineering and Computer Science, University of California, Merced \\
  {\url{http://eecs.ucmerced.edu}}
}
\date{March 2, 2015}

\begin{document}

\maketitle

\begin{abstract} 
  A natural way to characterize the cluster structure of a dataset is by finding regions containing a high density of data. This can be done in a nonparametric way with a kernel density estimate, whose modes and hence clusters can be found using mean-shift algorithms. We describe the theory and practice behind clustering based on kernel density estimates and mean-shift algorithms. We discuss the blurring and non-blurring versions of mean-shift; theoretical results about mean-shift algorithms and Gaussian mixtures; relations with scale-space theory, spectral clustering and other algorithms; extensions to tracking, to manifold and graph data, and to manifold denoising; $K$-modes and Laplacian $K$-modes algorithms; acceleration strategies for large datasets; and applications to image segmentation, manifold denoising and multivalued regression.
\end{abstract} 

\section{Introduction}
\label{ch4.1sec:intro}

\begin{figure}[t]
  \centering
  \psfrag{F1}[][]{\textbf{A}}
  \psfrag{F2}[][]{\textbf{B}}
  \psfrag{F3}[][]{\textbf{C}}
  \psfrag{F4}[][]{\textbf{D}}
  \begin{tabular}{@{}c@{\hspace{0\textwidth}}c@{}}
    \includegraphics[width=0.50\textwidth]{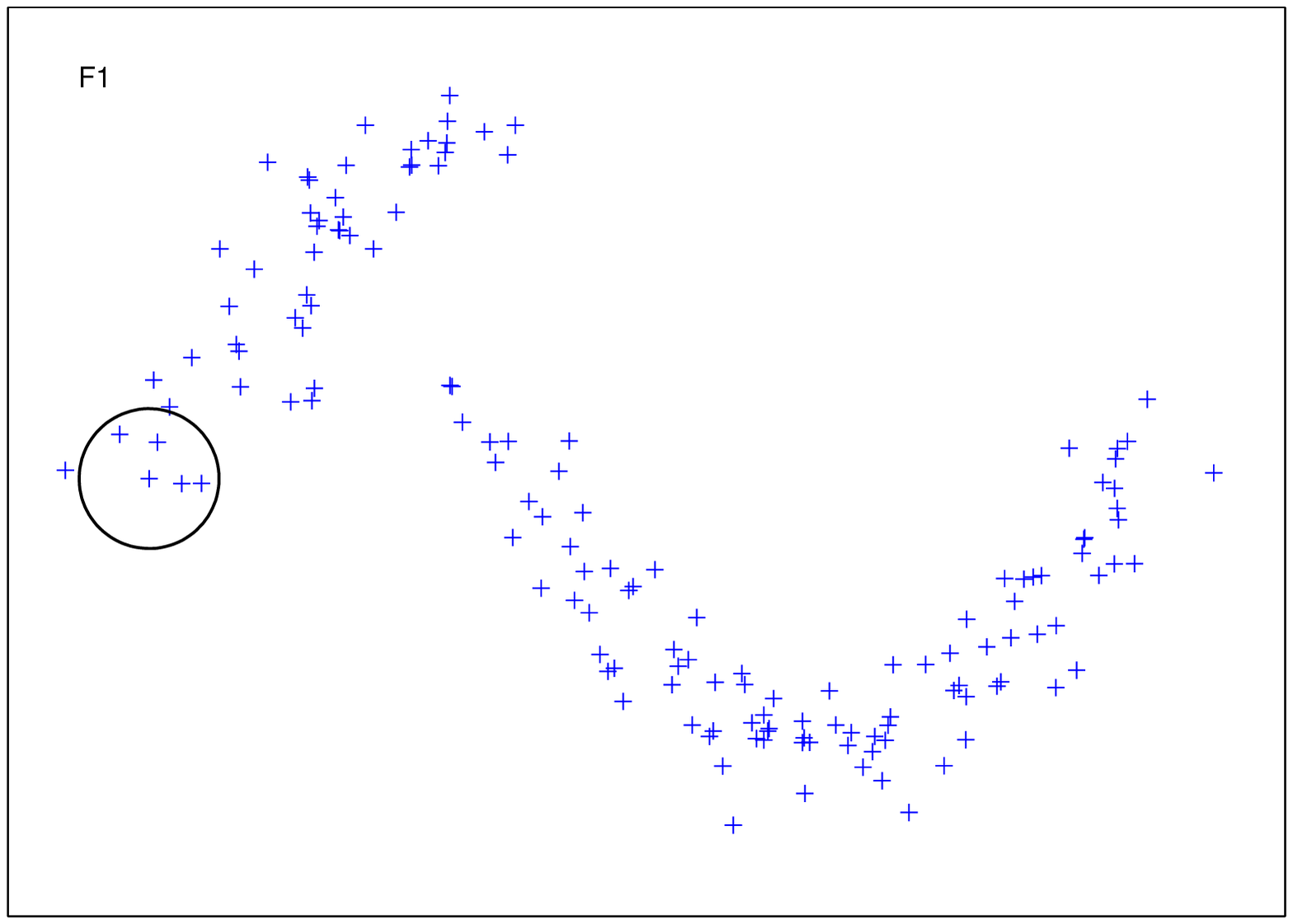} &
    \includegraphics[width=0.50\textwidth]{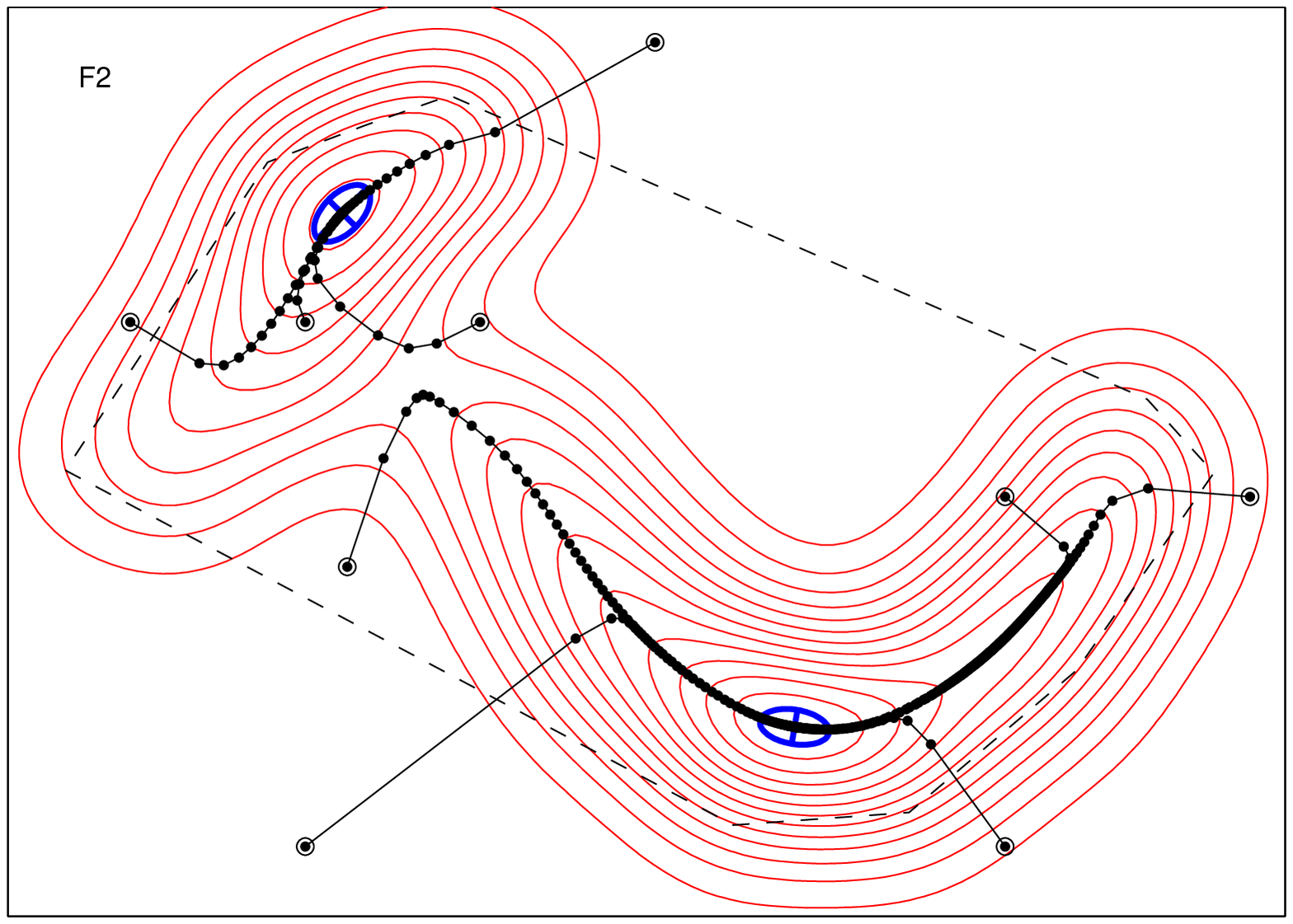} \\
    \includegraphics[width=0.50\textwidth]{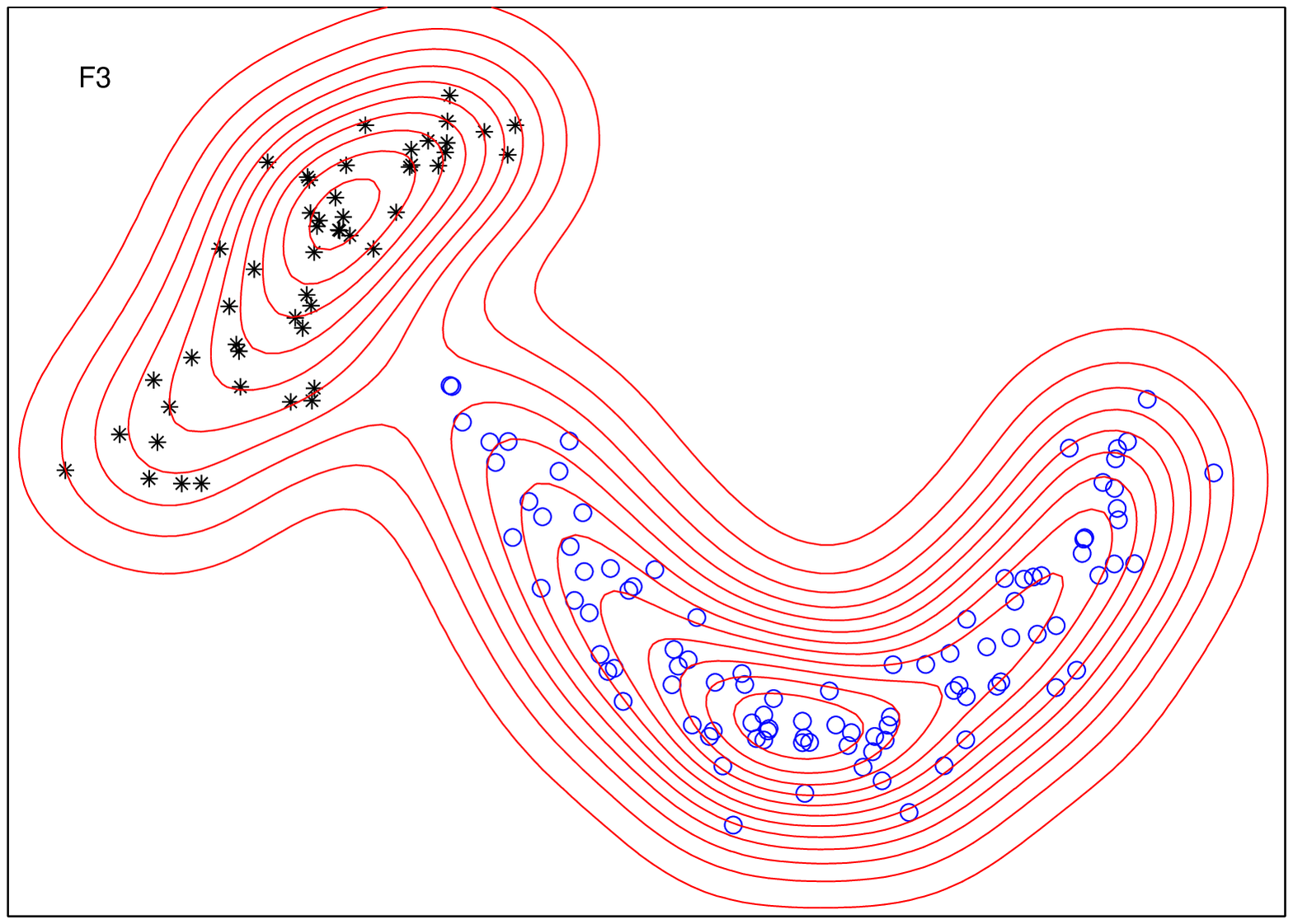} &
    \includegraphics[width=0.50\textwidth]{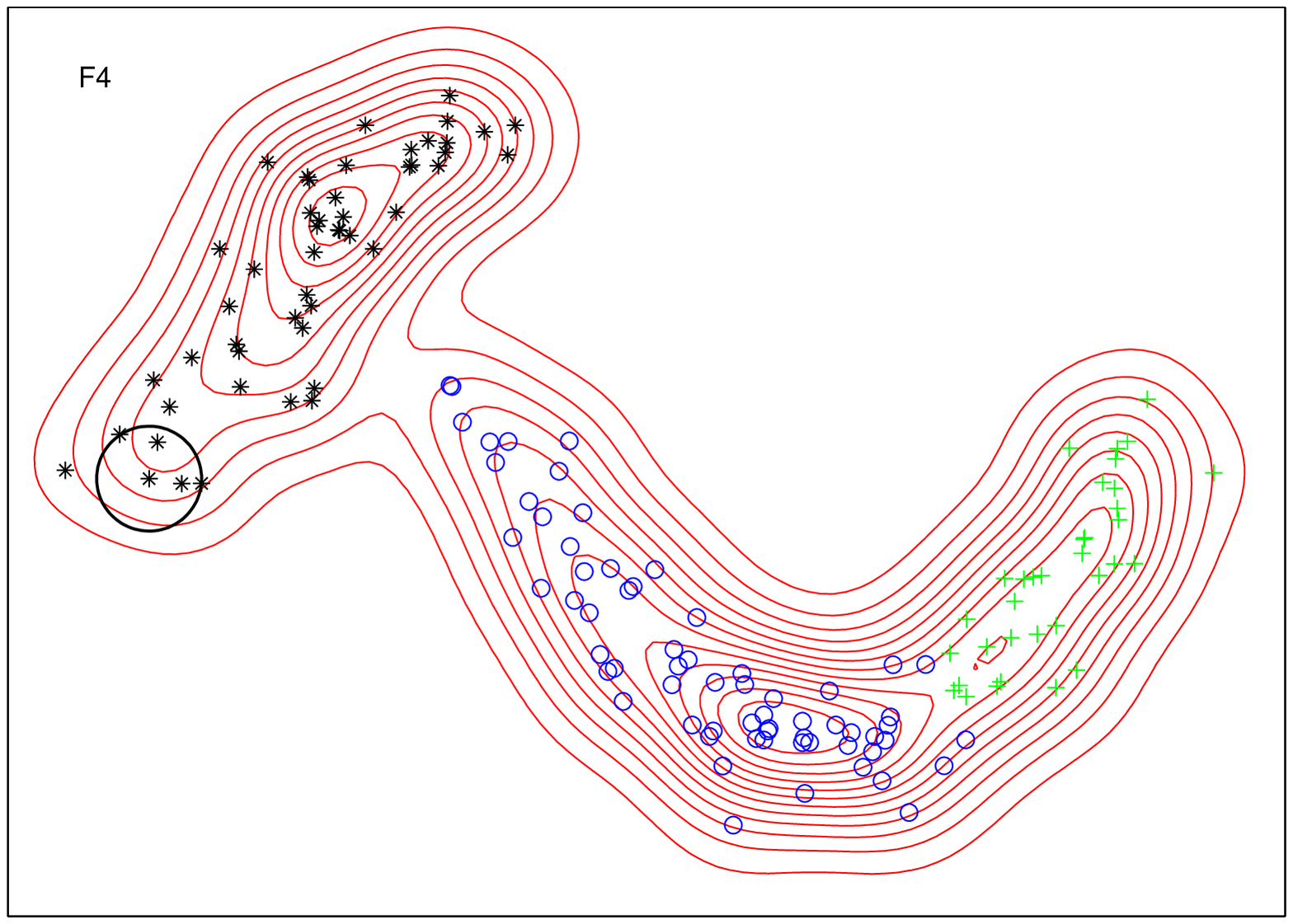}
  \end{tabular}
  \caption{Illustration in 2D of complex-shaped clusters, the kernel density estimate (KDE) and the mean-shift results. \textbf{A}: the dataset. The circle has a radius equal to the bandwidth $\sigma$ used in the KDE. \textbf{B}: a contour plot of the Gaussian KDE $p(\x)$ with bandwidth $\sigma$. The KDE has two modes, located at the center of the blue ellipses. Each ellipse indicates the eigenvectors (rescaled to improve visibility) of the Jacobian $\J(\x^*)$ of eq.~\eqref{e:MS-Jacobian} at a mode $\x^*$. The dotted-line polygon is the convex hull of the data points. The paths followed by Gaussian MS for various starting points are shown. \textbf{C}: the resulting clustering using mean-shift, with each point color-coded according to the cluster it belongs to. \textbf{D}: the resulting clustering and KDE contours using a smaller bandwidth (indicated by the radius of the black circle). Now, the KDE defines three clusters.}
  \label{f:2Dcontours}
\end{figure}

One intuitive way of defining clusters is to assume that the data points are a sample of a probability density function, and then to define the clusters through this density. For example, fig.~\ref{f:2Dcontours} shows a 2D dataset and a density estimate for it, whose contours clearly suggest that there are two clusters of a complex shape. The first step, then, is to learn an estimate of the density for the data points. This can be done with a parametric model, such as a Gaussian mixture, typically trained with an EM algorithm to maximize the likelihood \cite{MclachKrishn97a}. Such an approach is often computationally efficient and can give good results with clusters of elliptical shape, but it has several disadvantages. The likelihood function will typically have local optima, and finding a global optimum is, in general, very difficult; thus, the result is dependent on the initialization, and in practice a user will try different initializations (usually random restarts). The selection of the model (what kernel and how many components) is left to the user, as well as the number of clusters to find. And when the clusters have complex shapes, as for example in image segmentation, many components will be required to approximate them well, increasing the training time and the number of local optima.

We focus on \emph{nonparametric, kernel density estimates (KDE)}. A KDE is a generalization of histograms to define density estimates in any dimension that are smooth. They simplify the mathematical and computational treatment of densities and, crucially, enable one to use continuous optimization to find maxima of the density. With a kernel such as the Gaussian kernel, a KDE requires a single user parameter, the \emph{bandwidth} (also referred to as \emph{scale}). Given the bandwidth, the KDE is uniquely determined and, as seen below, so will be its clusters, which can take complex nonconvex shapes. Hence, the user need not select the number of clusters or try random restarts. We will focus on clusters defined by the modes of the KDE (although this is not the only way to define the clusters). A \emph{mode} is a local maximum of the density. A natural algorithm to find modes of a KDE is the \emph{mean-shift} iteration, essentially a local average, described in section~\ref{ch4.1sec:existing}. The basic idea in mean-shift clustering is to run a mean-shift iteration initialized at every data point and then to have each mode define one cluster, with all the points that converged to the same mode belonging to the same cluster. Section~\ref{ch4.1sec:theory} reviews theoretical results regarding the number and location of modes of a KDE, the convergence of mean-shift algorithms and the character of the cluster domains. Section~\ref{ch4.1sec:relations} discusses relations of mean-shift algorithms with spectral clustering and other algorithms. Sections~\ref{ch4.1sec:extensions} and~\ref{ch4.1sec:denoising} describe extensions of mean-shift for clustering and manifold denoising, respectively. One disadvantage of mean-shift algorithms is their computational cost, and section~\ref{ch4.1sec:accel} describes several accelerations. Section~\ref{ch4.1sec:kmodes} describes another family of KDE-based clustering algorithms which are a hybrid of $K$-means and mean-shift, the $K$-modes and Laplacian $K$-modes algorithms, which find exactly $K$ clusters and a mode in each, and work better with high-dimensional data. Section~\ref{ch4.1sec:apps} shows applications in image segmentation, inverse problems, denoising, and other areas. We assume multivariate, continuous (i.e., not categorical) data throughout.

\clearpage

\section{Problem formulation and mean-shift algorithms}
\label{ch4.1sec:existing}

In mean-shift clustering, the input to the algorithm are the data points (multivariate, continuous feature vectors) and the bandwidth or scale. Call $\{\x_n\}^N_{n=1} \subset \bbR^D$ the data points to be clustered. We define a kernel density estimate \cite{WandJones94a}
\begin{equation}
  \label{e:kde}
  p(\x) = \frac{1}{N} \sum^N_{n=1}{K\bigg( \norm{\frac{\x-\x_n}{\sigma}}^2 \bigg)} \qquad \x\in\bbR^D
\end{equation}
with bandwidth $\sigma > 0$ and kernel $K(t)$, e.g.\ $K(t) = e^{-t/2}$ for the Gaussian kernel or $K(t) = 1-t$ if $t \in [0,1)$ and $0$ if $t \ge 1$ for the Epanechnikov kernel. Many of the results below carry over to kernels where each point has its own weight and its own bandwidth, which can be an isotropic, diagonal or full covariance matrix. To simplify the presentation we focus on the case where all points have the same, scalar bandwidth $\sigma$ (the isotropic, homoscedastic case) and the same weight $\frac{1}{N}$ unless otherwise noted. This is the case found most commonly in practice. Also, we mostly focus on Gaussian kernels, which are easier to analyze and give rise to simpler formulas.

We can derive a simple iterative scheme $\x^{(\tau+1)} = \f(\x^{(\tau)})$ for $\tau = 0,1,2\dots$ to find a mode of $p$ by equating its gradient to zero and rearranging terms (section~\ref{ch4.1sec:accel} discusses other ways to find modes). We obtain 
\begin{equation}
  \label{e:MS}
  \f(\x) = \sum^N_{n=1}{ \frac{K'\big( \norm{\smash{\frac{\x-\x_n}{\sigma}}}^2 \big)}{\sum^N_{n'=1}{K'\big( \norm{\smash{\frac{\x-\x_{n'}}{\sigma}}}^2 \big)}} \x_n }
\end{equation}
where $K' = dK/dt$ and the vector $\f(\x) - \x$ is the \emph{mean shift}, since it averages the individual shifts $\x_n-\x$ with weights as above. For a Gaussian kernel, $K' \propto K$ and this simplifies to the following, elegant form (where, by Bayes' theorem, $p(n|\x) = p(\x|n) p(n)/p(\x)$ is the posterior probability of the component centered at $\x_n$ given point \x) \cite{Carreir00b}:
\begin{equation}
  \label{e:GMS}
  p(n|\x) = \frac{\exp{\bigl(-\frac{1}{2} \norm{(\x-\x_n)/\sigma}^2\bigr)}}{\sum^N_{n'=1}{\exp{\bigl(-\frac{1}{2} \norm{(\x-\x_{n'})/\sigma}^2\bigr)}}} \qquad
  \f(\x) = \sum^N_{n=1}{p(n|\x) \x_n}.
\end{equation}
As discussed below, under mild conditions this scheme converges to modes of $p$ from nearly any initial $\smash{\x^{(0)}}$. Intuitively, each step moves the iterate $\smash{\x^{(\tau)}}$ to a local average of the data, in that data points closer to $\smash{\x^{(\tau)}}$ have larger weight, and this increases the density. Eq.~\ref{e:MS} is called the \emph{mean-shift iteration}, and it can be used in two distinct ways: to find modes, and to filter (or smooth) a dataset. This gives rise to two different clustering algorithms, as follows. We will refer to them as \emph{mean shift (MS)} (where modes are found) and \emph{blurring mean shift (BMS)} (where the dataset is filtered).

\subsection{Two basic types of mean-shift algorithms: MS and BMS}
\label{ch4.1sec:MSalg}

\begin{figure}[t]
  \begin{center}
    \begin{tabular}{@{}c@{\hspace{0.04\linewidth}}c@{}}
      \begin{minipage}[t]{0.47\linewidth}
        \textbf{A}. Gaussian mean-shift (MS) algorithm \\
        \setlength{\fboxsep}{4pt}
        \framebox[\textwidth][l]{%
          \begin{minipage}[c]{0.95\textwidth}
            \begin{tabbing}
              m \= m \= m \= m \= m \= \kill
              \underline{\textbf{for}} $n \in \{1,\dots,N\}$ \+ \\
              $\x \leftarrow \x_n$ \\
              \underline{\textbf{repeat}} \+ \\
              $\forall n$: $p(n|\x) \leftarrow \frac{\exp{\bigl(-\frac{1}{2} \norm{(\x-\x_n)/\sigma}^2\bigr)}}{\sum^N_{n'=1}{\exp{\bigl(-\frac{1}{2} \norm{(\x-\x_{n'})/\sigma}^2\bigr)}}}$ \\
              $\x \leftarrow \sum^N_{n=1}{p(n|\x) \x_n}$ \- \\
              \underline{\textbf{until}} stop \\
              $\z_n \leftarrow \x$ \- \\
              \underline{\textbf{end}} \\
              connected-components($\{\z_n\}^N_{n=1}$,$\epsilon$)
            \end{tabbing}
          \end{minipage}%
        }
      \end{minipage} &
      \begin{minipage}[t]{0.47\linewidth}
        \textbf{B}. Gaussian blurring mean-shift (BMS) algorithm \\
        \setlength{\fboxsep}{4pt}
        \framebox[\textwidth][l]{%
          \begin{minipage}[c]{0.95\textwidth}
            \begin{tabbing}
              m \= m \= m \= m \= m \= \kill
              \underline{\textbf{repeat}} \+ \\
              \underline{\textbf{for}} $m \in \{1,\dots,N\}$ \+ \\
              $\forall n$: $p(n|\x) \leftarrow \frac{\exp{\bigl(-\frac{1}{2} \norm{(\x_m-\x_n)/\sigma}^2\bigr)}}{\sum^N_{n'=1}{\exp{\bigl(-\frac{1}{2} \norm{(\x_m-\x_{n'})/\sigma}^2\bigr)}}}$ \\
              $\y_m \leftarrow \sum^N_{n=1}{p(n|\x_m) \x_n}$ \- \\
              \underline{\textbf{end}} \\
              $\forall m$: $\x_m \leftarrow \y_m$ \- \\
              \underline{\textbf{until}} stop \\
              connected-components($\{\x_n\}^N_{n=1}$,$\epsilon$)
            \end{tabbing}
          \end{minipage}%
        }
      \end{minipage} \\ \\[-1ex]
      \begin{minipage}[t]{0.47\linewidth}
        \textbf{C}. Gaussian MS algorithm in matrix form \\
        \setlength{\fboxsep}{4pt}
        \framebox[\textwidth][l]{%
          \begin{minipage}[c]{0.95\textwidth}
            \begin{tabbing}
              m \= m \= m \= m \= m \= \kill
              $\Z = \X$ \\
              \underline{\textbf{repeat}} \+ \\
              $\W = \bigl( \exp{\bigl(-\frac{1}{2} \norm{(\z_m-\x_n)/\sigma}^2\bigr)} \bigl)_{nm}$ \\
              $\D = \diag{\sum^N_{n=1}{w_{nm}}}$ \\
              $\Q = \W \D^{-1}$ \\
              $\Z = \X \Q$ \- \\
              \underline{\textbf{until}} stop \\
              connected-components($\{\z_n\}^N_{n=1}$,$\epsilon$)
            \end{tabbing}
          \end{minipage}%
        }
      \end{minipage} &
      \begin{minipage}[t]{0.47\linewidth}
        \textbf{D}. Gaussian BMS algorithm in matrix form \\
        \setlength{\fboxsep}{4pt}
        \framebox[\textwidth][l]{%
          \begin{minipage}[c]{0.95\textwidth}
            \begin{tabbing}
              m \= m \= m \= m \= m \= \kill
              \underline{\textbf{repeat}} \+ \\
              $\W = \bigl( \exp{\bigl(-\frac{1}{2} \norm{(\x_m-\x_n)/\sigma}^2\bigr)} \bigl)_{nm}$ \\
              $\D = \diag{\sum^N_{n=1}{w_{nm}}}$ \\
              $\PP = \W \D^{-1}$ \\
              $\X = \X \PP$ \- \\
              \underline{\textbf{until}} stop \\
              connected-components($\{\x_n\}^N_{n=1}$,$\epsilon$)
            \end{tabbing}
          \end{minipage}%
        }
      \end{minipage}
    \end{tabular}
    \caption{Pseudocode for MS (left) and BMS (right), in both loop and matrix forms, for the Gaussian kernel. In all cases, the input is a dataset $\x_1,\dots,\x_N \in \bbR^D$ and a bandwidth $\sigma > 0$. See section~\ref{ch4.1sec:MSalg} for the stopping criterion and the connected-components threshold distance between points $\epsilon$.}
    \label{f:code}
  \end{center}
\end{figure}

\paragraph{Clustering by mean-shift (MS): find modes}

Here, we declare each mode of $p$ as representative of one cluster, and assign data point $\x_n$ to the mode it converges to under the mean-shift iteration, $\f^{\infty}(\x_n)$. The algorithm is given in fig.~\ref{f:code}\textbf{A} for the Gaussian kernel. We can also estimate error bars for each mode from the local Hessian \cite{Carreir00a,Carreir01a}, given in eq.~\eqref{e:MS-Jacobian}, which is related to the local covariance.

Some practical problems need to be solved. Firstly, some points (minima and saddle points) do not converge to modes. It is unlikely that this will happen with a finite sample, but if so such points can be detected by examining the Hessian or by a postprocessing step that checks for small clusters.

Second, the mean-shift iteration is stopped after a finite number of steps, for example when the relative change in the value of \x\ is smaller than a set tolerance \texttt{tol} $> 0$. This means that data points that in theory would converge to the same mode actually stop at numerically slightly different points. A postprocessing step is necessary to merge these into a unique mode. This can be done by finding the connected components%
\footnote{The connected-components algorithm is described in appendix~\ref{s:conncomp}.}
of a graph that has $N$ vertices, one for every convergence point, and has an edge between any pair of vertices lying within a small distance $\epsilon > 0$. The graph need not be explicitly constructed. The user should set \texttt{tol} small enough to converge to the modes with good accuracy, while limiting the computational cost incurred; and $\epsilon$ should be set quite larger than \texttt{tol}, but smaller than the distance between different true modes.

\paragraph{Clustering by blurring mean-shift (BMS): smooth the data}

Here, each point $\x_m$ of the dataset actually moves to the point $\f(\x_m)$ given by eq.~\eqref{e:MS}. That is, given the dataset $\X = \{\x_1,\dots,\x_N\}$, for each $\x_m \in \X$ we obtain a new point $\tilde{\x}_m$ by applying one step of the mean-shift algorithm: $\tilde{\x}_m = \f(\x_m)$. Thus, one iteration of blurring mean-shift results in a new dataset $\tilde{\X}$ which is a blurred (smoothed) version of \X. By iterating this process we obtain a sequence of datasets $\X^{(0)},\X^{(1)},\dots$ (and a sequence of kernel density estimates $p^{(0)}(\x),p^{(1)}(\x),\dots$) where $\X^{(0)}$ is the original dataset and $\X^{(\tau)}$ is obtained by blurring $\X^{(\tau-1)}$ with one mean-shift step (see fig.~\ref{f:BMS_cameraman}).

As will be shown below, Gaussian BMS can be seen as an iterated filtering (in the signal processing sense) that eventually leads to a dataset with all points coincident for any starting dataset and bandwidth. However, before that happens, the dataset quickly collapses into meaningful, tight clusters which depend on $\sigma$ (see fig.~\ref{f:BMS_cameraman}), and then these point-like clusters continue to move towards each other relatively slowly. A stopping criterion that detects this situation quite reliably is based on whether the entropy of the dataset changes \cite{Carreir06b} (a simpler criterion would be to stop when the update to \X\ is small, but this does not always give good clusters). As with MS clustering, a connected-components postprocessing step merges the points into actual clusters. The BMS algorithm is given in fig.~\ref{f:code}\textbf{B}.

\paragraph{Similarities and differences between MS and BMS}

Although both MS and BMS are based on the same mean-shift iteration, they are different algorithms and can produce different clustering results. Specifically, given a value of the bandwidth, the number of clusters resulting from MS and BMS is usually different. However, the collection of clusterings produced over a range of bandwidths can be quite similar.

BMS is quite faster than MS in number of iterations and in runtime, particularly if using the accelerated BMS algorithm (section~\ref{ch4.1sec:accel}), which introduces essentially no approximation error. However, MS (and also BMS) can be considerably accelerated if a small clustering error is tolerated (section~\ref{ch4.1sec:accel}).

In MS, the $N$ optimizations (one for each data point) proceed independently, but they could be done synchronously, as in BMS, without altering the result. However, practically this is wasteful, because the number of iterations required varies considerably among points, and a synchronous scheme would have to run the largest number of iterations. Conversely, it is possible to run BMS with asynchronous iterations, for example moving points as soon as their update is computed. However, this makes the result dependent on the order in which points are picked, and is unlikely to be faster than the accelerated algorithm described below.

\paragraph{Choice of bandwidth}

The fundamental parameter in mean-shift algorithms is the bandwidth $\sigma$, which determines the number of clusters. The statistics literature has developed various ways to estimate the bandwidth of a KDE \cite{Silver86a,WandJones94a}, mostly in the 1D setting, for example based on minimizing a suitable loss function (such as the mean integrated squared error), or more heuristic rules (such as making the bandwidth proportional to the average distance of each point to its $k$th nearest neighbor). While these bandwidth estimators are useful and can give reasonable results, they should be used with caution, because the bandwidth that gives the best density estimate (in a certain sense) need not give the best clustering---clustering and density estimation are, after all, different problems. Besides, clustering is by nature exploratory, and it is best to explore a range of bandwidths. Computationally, this is particularly easy to do in MS (see the scale-space discussion below).

It is also possible to use a different bandwidth $\sigma_n$ for each point (called \emph{adaptive KDE}), which can help with areas where points are sparse, for example. A good way to do this is with \emph{entropic affinities} \cite{HintonRoweis03a,VladymCarreir13a}, where the user sets a global number of neighbors $k$ and then, for each data point $n = 1,\dots,N$, the bandwidth $\sigma_n$ is computed so that point $n$ has a distribution over neighbors with a perplexity (log-entropy) $k$, i.e., each point sets its own bandwidth to have $k$ effective neighbors. One could then vary $k$ to achieve different clusterings. Other ways to construct adaptive KDEs for MS have been proposed \cite{Comanic03a}. The mean-shift update with adaptive bandwidths $\sigma_1,\dots,\sigma_N$ has the form \cite{Carreir07a}:
\begin{equation}
  \label{e:GMS-adaptive}
  q(n|\x) = \frac{p(n|\x) \sigma^{-2}_n}{\sum^N_{n'=1}{p(n'|\x) \sigma^{-2}_{n'}}} \qquad
  \f(\x) = \sum^N_{n=1}{q(n|\x) \x_n}
\end{equation}
where the $q(m|\x)$ values are the posterior probabilities $p(m|\x)$ reweighted by the inverse variance and renormalized (compare with the single-bandwidth equation~\eqref{e:GMS}).

Mean-shift has also been used to track moving objects (``blobs'') through a sequence of images \cite{Comanic_03a}. Since the blobs can change size, using a fixed bandwidth (scale) becomes problematic. Collins \cite{Collin03b} used ideas from scale-space theory (see below) to select a scale that adapts to the blob size. An ``image feature'' is defined as a point in scale-space where a certain differential operator achieves a local maximum (i.e., a mode) with respect to both space and scale. This mode is then tracked by a two-step mean-shift procedure that convolves in image space with a filterbank of spatial difference-of-Gaussian filters, and convolves in scale space with an Epanechnikov kernel, in an efficient way.

\paragraph{Choice of kernel}

Different kernels give rise to different versions of the mean-shift and blurring mean-shift algorithms. Much previous work (including \cite{FukunagHostet75a,Cheng95a}) uses the Epanechnikov kernel for computational efficiency, since the kernel evaluations involve only pairs of neighboring points (at distance $< \sigma$) rather than all pairs of points (though the neighbors must still be found at each iteration), and convergence occurs in a finite number of iterations. However, in practice, the Gaussian kernel produces better results than the Epanechnikov kernel \cite{ComanicMeer02a}, which generates KDEs that are only piecewise differentiable and can contain spurious modes (see below).

\paragraph{Hierarchical mean-shift and scale-space clustering}

The behavior of modes and other critical points (minima, saddles) as a function of the bandwidth (or scale) is the basis of \emph{scale-space theory} in computer vision \cite{Witkin83a,Koender84a,Lindeb94a}. Here, one studies the evolution of an image under Gaussian convolution (blurring). If we represent the image as a sum of delta functions centred at the pixel locations and with value equal to the grayscale, convolving this with an isotropic Gaussian kernel of scale $\sigma$ gives a KDE (with weighted components). As the scale increases, the image blurs, structures in the image such as edges or objects lose detail, and for large scales the image tends to uniformly gray. Some important structures, such as objects, can be associated with modes of the KDE, and the lifetime of a mode---defined by the scale interval between the creation and destruction (merging with another mode) of that mode---is taken as an indication of its importance in the image.

An ideal convolution kernel should blur structure in the image but never create new structure (which would then reflect properties of the kernel rather than of the image). The result would then be a tree of modes, where there are $N$ modes for $\sigma=0$ (one mode at each component) and modes merge as $\sigma$ increases until there is a single mode. Unfortunately, this is only true for the Gaussian kernel and only in dimension one (section~\ref{ch4.1sec:theory}). In dimension two, i.e., with images, new modes can be created as the scale decreases. However, in practice these creation events seem rare and short-lived, in that the mode created usually merges with another mode or critical point at a slightly larger scale. Thus, they are generally ignored. Computationally, one starts with a mode at each pixel (or data point) for $\sigma=0$ and then tracks the location of the modes as in a numerical continuation method \cite{NocedalWright06a}, by running mean-shift at a new, larger scale using as initial points the modes at the previous scale. By construction, this results in a tree---although, unlike in agglomerative clustering, the resulting set of clusterings at each scale need not be nested.

Another notion of lifetime, topological persistence, has been explored in computational geometry \cite{Edelsb_02a}, and used to define a hierarchical mean-shift image segmentation \cite{ParisDurand07a,Chazal_13a}. The tree of modes has also been proposed in the statistical literature as a tool for visualization of KDEs \cite{MinnotScott93a}, but this is practical only for small datasets in 1D or 2D. The mode tree is sensitive to small changes in the data and gives no way to differentiate between important modes and those caused by, for example, outliers, so it can help to combine several trees constructed by jittering or resampling the original dataset \cite{Minnot_98a}.

\paragraph{Matrix formulation of BMS and generalizations of BMS}

As indicated in fig.~\ref{f:code}\textbf{D}, we can equivalently write each Gaussian BMS iteration in matrix form \cite{Carreir06b} as $\X \leftarrow \X \PP$ in terms of the random-walk matrix $\PP = \W \D^{-1}$, an $N \times N$ stochastic matrix with elements $p_{nm} = p(n|\x_m) \in (0,1)$ and $\smash[t]{\sum^N_{n=1}{p_{nm}}} = 1$. Here, $\X = (\x_1,\dots,\x_N)$ is the $D \times N$ matrix of data points, $\W = \smash[b]{\bigl( \exp{\bigl(-\frac{1}{2} \norm{(\x_m-\x_n)/\sigma}^2\bigr)} \bigl)}_{nm}$ is a Gaussian affinity matrix, which defines a weighted graph where each $\x_n$ is a vertex, and $\D = \diag{\smash{\sum^N_{n=1}{w_{nm}}}}$ is the degree matrix of that graph. This establishes a connection with spectral clustering, which we describe in section~\ref{ch4.1sec:relations}.

Also, we can regard the iteration $\X \leftarrow \X\PP$ as a smoothing, or more generally a filtering \cite{Taubin95a,Desbrun_99a,Taubin00a,Carreir08a}, of (each dimension of) the data \X\ with an inhomogeneous filter \PP, where the filter depends on the data and is updated at each iteration as well, hence resulting in a nonlinear filtering. This in turn suggests that one could use other filters constructed as a function $\phi(\PP)$, where $\phi$ is a scalar function, so it modifies the spectrum of \PP. Carreira-Perpi{\~n}{\'a}n \cite{Carreir08a} studied several such filters, including explicit, implicit, power and exponential ones, depending on a step size parameter $\eta$, resulting in generalized Gaussian blurring mean-shift algorithms. He gave convergence conditions on $\phi$ and $\eta$ and found that the different filters tend to find similar clusters, i.e., over a range of bandwidths they can obtain the same clustering (at possibly different $\sigma$ values). However, their runtime varies widely. Implicit filters (which involve solving a linear system) or power filters (which involve iterating a matrix product) have a strong clustering effect in each iteration, but their iterations themselves are more costly. When one considers both the number of iterations and the cost of each iteration, the method found fastest was a slightly overrelaxed explicit function of the form $\phi(\PP) = (1-\eta)\I + \eta \PP$ with $\eta \approx 1.25$. However, its runtime was very close to that of the standard BMS ($\eta=1$). An interesting extension of this work would be to be able to design the function $\phi$ so that the resulting generalized BMS algorithm is optimal (in some sense) for clustering.

It is also possible to write MS in matrix form (fig.~\ref{f:code}\textbf{C}), where we write the random-walk matrix with the symbol \Q\ to differentiate it from the standard random-walk matrix \PP, since $q_{nm} = p(\x_n|\z_m)$ is defined on two sets of points (\X\ and \Z) while \PP\ is defined on one ($\X=\Z$). However, the matrix form implies that the MS updates for all points are synchronous, which, as mentioned before, is slow.

\paragraph{Mapping new points to clusters (out-of-sample mapping)}

Having run MS or BMS on a dataset, how should we deal with new data points not in the original dataset? The purist option (and the only one for BMS) is to run the clustering algorithm from scratch on the entire dataset (old and new points), but this is computationally costly. A faster option with MS is to use the original KDE and simply run MS on each of the new points, assigning them to the mode they converge to. This reflects the fact that MS clusters not just the points in the dataset, but (implicitly) the whole space. However, this point of view implies no new clusters are created when new points arrive.

\paragraph{Advantages and disadvantages of mean-shift algorithms}

The advantages of mean-shift algorithms stem from the nonparametric nature of the KDE: (1) It makes no model assumptions (other than using a specific kernel), unlike Gaussian mixture models or $K$-means, for example. (2) It is able to model complex clusters having nonconvex shape (although this does not imply that all shapes can be modeled well), unlike $K$-means. (3) The user need only set one parameter, the bandwidth, which has an intuitive physical meaning of local scale, and this determines automatically the number of clusters. This is often more convenient than having to select the number of clusters explicitly. (4) It has no local minima, thus the clustering it defines is uniquely determined by the bandwidth, without the need to run the algorithm with different initializations. (5) Outliers, which can be very problematic for Gaussian mixtures and $K$-means, do not overly affect the KDE (other than creating singleton clusters).

Using KDEs and equating modes with clusters as in mean-shift also has some disadvantages. The most important one is that KDEs break down in high dimensions, where the number of clusters changes abruptly from one for large $\sigma$ to many, with only a minute decrease in $\sigma$. Indeed, most successful applications of mean-shift have been in low-dimensional problems, in particular image segmentation (using a few features per pixel, such as color in LAB space). A way of using modes in high-dimensional spaces is the $K$-modes algorithm described in section~\ref{ch4.1sec:kmodes}. 

Other possible disadvantages, depending on the case, are as follows. (1) In some applications (e.g.\ figure-ground or medical image segmentation) the user may seek a specific number of clusters. However, in mean-shift clustering we have no direct control over the number of clusters: to obtain $K$ clusters, one has to search over $\sigma$. This is computationally costly (and sometimes not well defined, since the number of clusters might not be a monotonic function of $\sigma$). (2) We do not differentiate between meaningful and non-meaningful modes. For example, outliers will typically create their own mode; or, the density in a cluster may genuinely contain multiple modes (especially with clusters that have a nonconvex or manifold structure, as in fig.~\ref{f:2Dcontours}\textbf{D}). Some of these problems may be partially corrected by postprocessing the results from mean-shift (e.g.\ to remove low-density modes and clusters with few points, which are likely outliers), or by the $K$-modes algorithm. Finally, mean-shift is slow computationally, and this is addressed in section~\ref{ch4.1sec:accel}.

\subsection{Origins of the mean-shift algorithm}
\label{ch4.1sec:origins}

The mean-shift algorithm is so simple that it has probably been discovered many times. In 1975, Fukunaga and Hostetler \cite{FukunagHostet75a} were perhaps the first to propose its idea and also introduced the term ``mean shift''. They derived the blurring version of the algorithm (BMS) for a KDE with the Epanechnikov kernel as gradient ascent on $\log{p(\x)}$ with a variable step size, without proving convergence or giving a stopping criterion. They observed that it could be used for clustering and dimensionality reduction (or denoising), since points converge locally to cluster centroids or medial lines for appropriate values of the bandwidth. Since 1981, the algorithm was also independently known in the statistics literature as ``mean update algorithm'' (see \cite{ThompsTapia90a} pp.~167ff and references therein). The term ``blurring process'' is due to Cheng \cite{Cheng95a}, who discussed the convergence of both the blurring (BMS) and the non-blurring (MS) versions of mean-shift. Carreira-Perpi{\~n}{\'a}n \cite{Carreir00b}, motivated by the problem of finding all the modes of a Gaussian mixture for multivalued regression \cite{Carreir00a,Carreir01a}, independently rediscovered the algorithm for the Gaussian kernel and proved its convergence for arbitrary covariance matrices \cite{Carreir01a}. Since the early 2000s, the non-blurring mean-shift received much attention thanks to the work of Comaniciu and Meer \cite{ComanicMeer02a}, who demonstrated its success in image filtering, image segmentation and later in tracking \cite{Comanic_03a}. This was followed by work by many researchers in both theoretical, computational and application issues. Algorithms similar to mean-shift have appeared in scale-space clustering \cite{WilsonSpann90a,Wong93a,ChakravGhosh96a,Robert97a}, in clustering by deterministic annealing \cite{Rose98a} and in pre-image finding in kernel-based methods \cite{Schoel_99b}.

\subsection{Theoretical results about mean-shift algorithms and Gaussian mixtures}
\label{ch4.1sec:theory}

Although MS and BMS are defined by very simple iterative schemes, they exhibit some remarkable and somewhat counterintuitive properties regarding whether they converge at all, how fast they converge, and the character of the convergence domains. The geometry of the modes of a KDE is also surprising. The relevant literature is scattered over different areas, including computer vision, statistics and machine learning. We give a summary of results, with particular attention to Gaussian kernels, without proof. More details can be found in the references cited.

\paragraph{Geometry of the modes of a Gaussian mixture}

Intuitively, one might expect that a sum of $N$ unimodal kernels with bandwidth $\sigma$ as in eq.~\eqref{e:kde} would have at most $N$ modes, and that the number of modes should decrease monotonically as $\sigma$ increases from zero and the different components coalesce. In general, this is only true for the Gaussian kernel in dimension one. Motivated by scale-space theory, several papers \cite{Koender84a,Babaud_86a,YuillePoggio86a} showed that, in dimension one, the Gaussian kernel is the only kernel that does not create modes as the scale increases. It is easy to see that, in KDEs with non-gaussian kernels (Epanechnikov, Cauchy, etc.), modes can appear as $\sigma$ increases \cite{CarreirWilliam03a}. The creation of modes need not occur often, though, and some kernels (such as the Epanechnikov kernel) are more likely than others to create modes. It is less easy to see, but nonetheless true, that modes can also appear with the Gaussian kernel in dimension two, i.e., with images, and thus in any larger dimension, as shown by an example in \cite{LifshitPizer90a}.

The scale-space theory results were restricted to a single bandwidth $\sigma$, and also the creation of modes does not necessarily imply that a mixture with $N$ components may have more than $N$ modes. The results for the general case of Gaussian mixtures (GMs) are as follows. Again, there is a qualitative difference between 1D and 2D or more. A GM (with possibly different bandwidths $\sigma_n$) can have at most $N$ modes in 1D \cite{CarreirWilliam03b} but more than $N$ modes in 2D or above, even if the components are isotropic \cite{CarreirWilliam03c}. In 2D, if the components have diagonal or full covariance matrices, it is easy to construct examples with more modes than components \cite{Carreir01a,CarreirWilliam03b}. It is far harder to achieve this if all the components are isotropic with equal bandwidth $\sigma$, but still possible. This was first shown by a construction suggested by Duistermaat and studied by Carreira-Perpi{\~n}{\'a}n and Williams \cite{CarreirWilliam03c}, consisting of 3 Gaussians in the vertices of an equilateral triangle. For a narrow interval of scales (lifetime), an extremely shallow fourth mode appears at the triangle barycenter. They generalized this construction to dimension $D$ as a regular simplex with a Gaussian in each vertex, i.e., a GM with $D+1$ components, one at each vertex of the simplex \cite{CarreirWilliam03c}. They showed that the number of modes is either $D+1$ (for small scale, modes near the vertices), $1$ (for large scale, at the simplex barycenter) or $D+2$ (for a narrow range of intermediate scales, modes at the barycenter and near the vertices). Interestingly, the lifetime of the mode that is created in the barycenter peaks at $D=698$ and then slowly decreases towards zero as $D$ increases. Small perturbations (but not vanishingly small) of the construction prevent the creation of the extra mode, which suggests that it may be unlikely for isotropic GMs to have more modes than components. However, apart from a few isolated studies, the geometry of the modes of GMs in high dimensions is poorly understood.

As for the location of the modes of a GM, they are in the interior of the convex hull of the data points if the components are isotropic (with possibly different bandwidths $\sigma_n$), for any dimension \cite{Carreir01a,CarreirWilliam03b}, as seen in fig.~\ref{f:2Dcontours}\textbf{B}. With KDEs, it is easy to see this from the fact that eq.~\eqref{e:MS} is a convex sum if $K' \ge 0$. If the components have diagonal or full covariance matrices, the modes can be outside the convex hull \cite{Carreir01a,CarreirWilliam03a,CarreirWilliam03b}.

Most work on mean shift and scale space theory tacitly assumes that the modes of a GM are finite in number or at least are isolated. Although sometimes this is claimed to be a consequence of Morse theory, no proof seems to exist for an arbitrary dimension
\footnote{In dimension one, this is simple to prove assuming infinite differentiabilty (as for the Gaussian kernel), by setting $f$ to the gradient of the mixture in the following result. Let $f(x)$ be an infinitely differentiable real function of $x\in\bbR$. Then either $f$ is identically zero or its zero crossings are isolated. Indeed, if $f$ is zero in a nonempty interval $(a,b)$, then $\forall x\in(a,b)$ we have $\lim_{h\rightarrow 0}{(f(x+h)-f(x))/h} = 0 = f'(x)$. Repeating this argument for $f'$, $f''$, etc.\ we obtain that all derivatives at any $x \in (a,b)$ are zero, hence by Taylor's theorem $f(x)=0$ $\forall x\in\bbR$.}.
Indeed, kernels such as the uniform or triangular kernel do create continuous ridges of modes even in 1D.

\paragraph{Convergence of MS}

An attractive feature of MS is that it is defined without step sizes, which makes it deterministic given the dataset and bandwidth. However, this is only useful if it converges at all, and whether this is the case depends on the kernel used. Not all kernels $K$ give rise to convergent MS updates, even if they are valid kernels (i.e., they are nonnegative and integrate to one); an example where MS diverges appears in \cite{ComanicMeer02a}. For kernels where MS diverges, it is of course possible to devise optimization algorithms that will converge to the modes, e.g.\ by introducing a line search, but we lose the simplicity of the MS iteration. For kernels where MS converges, the convergence is to a mode for most initial points, but in general to a stationary point (minima and saddle points in addition to modes). Here, we review convergence results for MS, with a focus on the most common kernels (Gaussian and Epanechnikov).

With the Epanechnikov kernel, convergence occurs in a finite number of iterations \cite{ComanicMeer02a}. The intuitive reason for this is that the KDE, which is the sum of $N$ kernels, is piecewise quadratic. There is a finite number of ``pieces'' (each corresponding to a possible subset of the $N$ kernels), and, within one piece, a single Newton iteration would find the maximum (indeed, the MS update coincides with a Newton step for the Epanechnikov kernel \cite{FashinTomasi05a}). 

With the Gaussian kernel, MS also converges, but in an infinite number of iterations, and the convergence rate is generally linear (sublinear or superlinear in limit cases) \cite{Carreir01a,Carreir07a}. One convenient way to see this is by relating MS with the EM algorithm \cite{MclachKrishn08a}, as follows.

In general for mixtures of Gaussians (using arbitrary covariance matrices) or other kernels (and thus in particular for KDEs), Gaussian MS is an expectation-maximization (EM) algorithm and non-gaussian MS is a generalized EM algorithm \cite{Carreir01a,Carreir07a}. This can be seen by defining a certain dataset and a certain probabilistic model with hidden variables (missing data), deriving the corresponding EM algorithm to maximize the log-likelihood for it, and verifying it coincides with the MS algorithm. For the GM case, the model consists of a constrained mixture of Gaussians, where each component has the given weight, mean and covariance matrix (constant, data point and bandwidth for KDEs, respectively), but the whole mixture can be freely (but rigidly) translated. The dataset consists solely of one point located at the origin. Thus, maxima of the likelihood occur whenever the translation vector is such that a mode of the mixture coincides with the origin. The missing data is the index of the mixture component that generated the origin. The resulting E step computes the posterior probabilities $p(n|\x)$ of eq.~\eqref{e:GMS}, and the M step maximizes a lower bound on the log-likelihood in closed form giving the MS update. With non-gaussian kernels, the M step cannot be solved in closed form, and the MS update corresponds to a single iteration to solve this M step. (Whether this iteration actually increases the likelihood, leading to convergence, depends on the kernel.)

Viewing MS as an EM algorithm has several consequences. Convergence for Gaussians MS is assured by the EM convergence theorems \cite{MclachKrishn08a} (which apply because the $Q$ function in the EM algorithm is continuous), and each iterate increases or leaves unchanged the density. General results for EM algorithms also indicate that the convergence order will typically be linear (with a rate dependent on the ratio of information matrices). Finally, MS can be seen as a bound optimization (see also \cite{FashinTomasi05a}).

The convergence order can be studied in detail \cite{Carreir07a} by linearizing the MS update $\x \leftarrow \f(\x)$ of eq.~\eqref{e:GMS} around a mode $\x^*$, so the update can be written as $\x^{(\tau+1)}-\x^* \approx \J(\x^*) (\x^{(\tau)}-\x^*)$, where $\J(\x^*)$ is the $D \times D$ Jacobian of \f. The Jacobian of \f\ and the Hessian of $p$ are related as follows:
\begin{equation}
  \label{e:MS-Jacobian}
  \J(\x^*) = \frac{1}{\sigma^2} \bSigma(\x^*) \qquad \nabla^2 p(\x^*) = \frac{p(\x^*)}{\sigma^2} (\J(\x^*) - \I) \qquad \bSigma(\x^*) = \sum^N_{n=1}{p(n|\x^*) (\x_n-\x^*) (\x_n-\x^*)^T}
\end{equation}
where $\bSigma(\x^*)$ is the local covariance matrix at $\x^*$, since (from $\x^* = \f(\x^*)$ and the definition of \f) the local mean is $\x^*$ itself. The eigenvalues of $\J(\x^*)$ are in $(0,1)$ and the convergence rate is given by the largest one, with the iterates approaching $\x^*$ along the eigenvector associated with the largest eigenvalue (assuming distinct eigenvalues). This shows that the convergence order depends on the bandwidth: it is nearly always linear, approaches superlinear convergence when $\sigma \rightarrow 0$ or $\sigma \rightarrow \infty$, and converges sublinearly at mode merges. The practically useful cases of MS use an intermediate $\sigma$ value, for which the rate $r$ of linear convergence (i.e., the ratio of distances to the mode after and before the update) can be close to $1$, thus convergence will be slow. The MS iterates smoothly approach the mode along the principal component of the local covariance matrix of the data points, from within the convex hull of the data points (see fig.~\ref{f:2Dcontours}\textbf{B}).

\paragraph{Other properties of MS}

We focus on Gaussian MS on a KDE, i.e., with isotropic covariances of bandwidth $\sigma$. As seen in fig.~\ref{f:2Dcontours}\textbf{B}, if initialized at points far from a mode, the first MS steps are often large and make considerable progress towards the mode. This is an advatageous property generally observed in alternating optimization algorithms (such as EM). After that, the steps become small, in agreement with the linear convergence rate.

The path followed by the MS iterates has the following properties (illustrated in fig.~\ref{f:2Dcontours}\textbf{B}). (1) Near convergence, the path follows the direction of the principal eigenvector of the local covariance at the mode \cite{Carreir07a}. (2) Each iterate is a convex linear combination of the data points, so the path lies in the interior of the convex hull of the data points \cite{Carreir00b,Carreir07a}. This is also true of non-gaussian kernels if they satisfy $K'\ge 0$. (3) The path is smooth in the sense that consecutive steps (consecutive mean-shift vectors $\f(\x)-\x$) always make an angle in $(-\frac{\pi}{2},\frac{\pi}{2})$ \cite{ComanicMeer02a}. (4) The mean shift vector $\f(\x) - \x$ is proportional to the gradient of $p$, so the MS iteration is a gradient step, but the step size is not the best one (i.e., does not maximize $p$) along the gradient.

In $K$-means, the centroids partition the space into Voronoi cells, which are convex polytopes. In MS, it is harder to characterize the regions that the modes partition the space into, i.e., their domains of convergence, which can in fact have surprising properties \cite{Carreir07a}. In general, they can be curved, nonconvex and disconnected (for example, one domain can be completely surrounded by another). This allows MS to represent complex-shaped clusters and is an advantage over $K$-means. A less desirable aspect is that the domain boundaries can show fractal behavior, although this seems confined to cluster boundaries, and could be removed if necessary by postprocessing the clusters. This fractal behavior is due to the iterated MS mapping $\f(\x)$, and would not occur if we defined the clusters purely based on flow lines of the gradient.

\paragraph{Convergence of BMS}

Cheng \cite{Cheng95a} proved convergence of blurring mean-shift, as follows. (1) For kernels broad enough to cover the dataset \X\ (e.g.\ infinite-support kernels such as the Gaussian) convergence is to a dataset $\X^{(\infty)}$ with all points coincident ($\smash{\x^{(\infty)}_1 = \cdots = \x^{(\infty)}_N}$), regardless of the value of $\sigma$. This can be seen by noting that the diameter of the data set decreases at least geometrically. (2) For finite-support kernels and small enough $\sigma$, convergence is to several clusters with all points coincident in each of them; the clusters depend on the value of $\sigma$.

Another proof can be obtained from the matrix formulation of BMS \cite{Carreir06b}, since at each iteration the dataset \X\ is multiplied times a stochastic matrix $\PP(\X)$. By the Perron-Frobenius theorem \cite[ch.~8]{HornJohnson86a}, with broad kernels this will have a single eigenvalue equal to 1 and all other eigenvalues with magnitude less than 1. Since a fixed point verifies $\X = \X \PP$ then $\X = \x \1^T$ for some $\x \in \bbR^D$, i.e., all points coincide. For non-broad kernels, the unit eigenvalue is multiple, resulting in multiple clusters where all points coincide.

While, for Gaussian BMS, the dataset converges to a size (therefore variance) zero, its variance need not decrease monotonically, in fact it is easy to construct examples in 1D where it increases at some steps.

Carreira-Perpi{\~n}{\'a}n \cite{Carreir06b} completely characterized the behavior with the Gaussian kernel (Gaussian BMS), assuming that the dataset is infinite with a Gaussian distribution. In this case, one can work with distributions rather than finite samples, and the iteration $\X \leftarrow \X\PP$ can be written as a Gaussian integral and solved in closed form. The result is that the data distribution $p(\x)$ is Gaussian after each iteration, with the same mean, and it shrinks towards its mean independently along each principal axis and converges to it with cubic order. Specifically, the standard deviation $s$ along a given principal axis evolves as $s \leftarrow r(s) s$ with $r(s) = \smash{\frac{1}{1+(\sigma/s)^2}} \in (0,1)$, where $\sigma$ is the BMS bandwidth, and $s$ converges to 0 cubically. The reason for this extremely fast convergence is that, since $\sigma$ is kept constant but the dataset shrinks, effectively $\sigma$ increases. Thus, at each iteration both $s$ and $\smash{\frac{1}{1+(\sigma/s)^2}}$ decrease. Note that the smaller the initial $s$ is, the faster the convergence and so the direction of largest variance (principal component) collapses much more slowly (in relative terms) than all other directions.

This explains the practical behavior shown by Gaussian BMS (see fig.~\ref{f:BMS_cameraman}): (1) clusters collapse extremely fast (in a handful of iterations, for a suitable bandwidth); (2) after a few iterations only the local principal component survives, resulting in temporary linearly-shaped clusters (that quickly straighten). These two behaviors make BMS useful for clustering and denoising, respectively.

The same proof technique applies to the generalized Gaussian BMS algorithms that use an update of the form $\X \leftarrow \X \, \phi(\PP(\X))$. With a Gaussian dataset, each iteration produces a new Gaussian with a standard deviation (separately along each principal axis) $s \leftarrow \abs{\phi(r(s))} s$. This allows a complete characterization of the conditions and order of convergence in terms of the real function $\phi(r)$, $r \in (0,1)$, instead of a matrix function. Convergence occurs if $\abs{\phi(r)} < 1$ for $r \in (0,1)$ and $\phi(1)=1$. Depending on $\phi$, the convergence order can vary from linear to cubic and beyond \cite{Carreir08a}.

\clearpage

\subsection{Relations with other algorithms}
\label{ch4.1sec:relations}

\paragraph{Spectral clustering}

As noted earlier, putting Gaussian BMS in matrix form in terms of a Gaussian affinity matrix \W\ uncovers an intimate relation with spectral clustering. Each BMS iteration is a product $\X \leftarrow \X \PP$ of the data times $\PP = (p(n|\x_m))_{nm}$, the stochastic matrix of the random walk in a graph \cite{Chung97a,MeilaShi01a}, which in BMS represents the posterior probabilities of each point under the kernel density estimate~\eqref{e:kde}. \PP\ is closely related to the matrix $\N = \smash{\D^{-\frac{1}{2}} \W \D^{-\frac{1}{2}}}$ (equivalent to the normalized graph Laplacian) commonly used in spectral clustering, e.g.\ in the normalized cut \cite{ShiMalik00a}. The eigenvalue/eigenvector pairs $(\mu_n,\uu_n)$ and $(\lambda_n,\vv_n)$ of \PP\ and \N\ satisfy $\mu_n = \lambda_n$ and $\uu_n = \smash{\D^{-\frac{1}{2}}} \vv_n$. In spectral clustering, given $\sigma$ and \X\ one computes the eigenvectors associated with the top $K$ eigenvalues of \N\ (if $K$ clusters are desired); in this spectral space the clustering structure of the data is considerably enhanced and so a simple algorithm such as $K$-means can often find the clusters. In BMS, we iterate the product $\X \leftarrow \X \PP$. If \PP\ were kept constant, this would be the power method \cite{GolubLoan96a} and each column of \X\ would converge to the leading left eigenvector of \PP\ (the vector of ones, i.e., a single cluster), with a rate of convergence given by the second eigenvalue $\mu_2 < 1$ (the Fiedler eigenvalue in spectral clustering). However, the dynamics of BMS is more complex because \PP\ also changes after each iteration. In practice \PP\ and \X\ quickly reach a quasistable state where points have collapsed in clusters which slowly approach each other and \PP\ remains almost constant (at which point BMS is stopped). Thus, BMS can be seen as refining the original affinities into a matrix consisting of blocks of (nearly) constant value and then (trivially) extracting piecewise-constant eigenvectors for each cluster with the power method. With the generalized BMS algorithm, one uses instead the matrix $\phi(\PP)$, which has the same eigenvectors $\vv_1,\dots,\vv_N$ as \PP\ but eigenvalues $\phi(\lambda_1),\dots,\phi(\lambda_N)$. However, this manipulation of the spectrum of \PP\ is performed implicitly, without actually having to compute the eigenvectors as in spectral clustering.

While both spectral clustering and BMS rely on the random-walk matrix \PP, they differ in several respects. They do not give the same clustering results. In spectral clustering, the user sets the desired number of clusters $K$ and the bandwidth $\sigma$ (if using Gaussian affinities), while in BMS the user sets only $\sigma$ and $K$ is determined by this. Computationally, spectral clustering solves an eigenproblem (and then runs $K$-means), while BMS (especially if using its accelerated version) performs a small number of matrix products (and then runs connected-components), thus it is considerably faster.

\paragraph{Bilateral filtering and nonlinear diffusion}

Many image processing algorithms (for example, for image denoising) operate on \emph{range variables} (intensity, color) defined on the \emph{space variables} (of the image lattice). This includes bilateral filtering \cite{Paris_08a}, nonlinear diffusion \cite{Weicker98a} and others. Mean-shift is basically an iterated, local averaging that operates jointly on range and space, and bears both similarities and differences with those algorithms, which are described in \cite{BarashComanic04a}. For example, in bilateral filtering the spatial component is fixed during iterations, so only the range variables are updated, and a stopping criterion is necessary to prevent excessive smoothing.

\paragraph{Other mean-shift-like algorithms}

Mean-shift algorithms appear whenever we have expressions having the form of a sum over data points of a function of squared distances of the parameter \x\ (whether the latter is a vector or a matrix or a set thereof). Equating the gradient of this expression to zero we can solve for \x\ and obtain a fixed-point iteration with the form of a weighted average of the data points, where the weights depend on \x. One example mentioned below are Riemannian centers of mass. Another example are Laplacian objective functions, of the form $\sum^N_{n,m=1}{w_{nm} \norm{\x_n-\x_m}^2} = 2 \trace{\X \LL \X^T}$, where \X\ is of $L \times N$ are coordinates of the $N$ data points in a low-dimensional space, $\W = (w_{nm})$ is an affinity matrix and \LL\ is its graph Laplacian. Alternating optimization over each $\x_n$ can be done with a mean-shift algorithm. Laplacian objectives appear in algorithms for (among other problems) dimensionality reduction and clustering, such as Laplacian eigenmaps \cite{BelkinNiyogi03b}, the elastic embedding \cite{Carreir10a} or spectral clustering \cite{ShiMalik00a}. However, as noted in section~\ref{ch4.1sec:theory}, the resulting mean-shift iteration need not converge in general. This is the case, for example, when the weights or the kernels can take negative values, as in the elastic embedding. In this case, one should use a line search or a different optimization algorithm altogether.

\subsection{Extensions of mean-shift for clustering}
\label{ch4.1sec:extensions}

\paragraph{Tracking: mode finding over time}

In some applications, the distribution of the data changes over time, and we want to track clusters and their location, in order to predict their location at a future time (assuming the distribution changes slowly). One example is tracking a nonrigid object over a sequence of images, such as a video of a moving car. A robust way to represent the object is by the color histogram of all the pixels within the object. In its simplest form, one can use a region of fixed shape and size but variable location, initialized by the user in the first image. Then, the most likely location of the object in the next image can be defined as the region (near the current region) having the histogram closest to the current histogram. Comaniciu et al.\ \cite{Comanic_03a} noted that we can use a differentiable KDE instead of a histogram, and that any differentiable similarity function of two KDEs also has the form (up to constant factors) of a weighted KDE to first order (a good approximation if the object moves slowly). Hence, we can use mean-shift iterations to maximize the (linearized) similarity over the location, thus finding the location where pixels look most like in the previous image's region. The resulting algorithm is fast enough to track several objects in real-time in video, and is robust to partial occlusion, clutter, distractors and camera motion.

Hall et al.\ \cite{Hall_06a} define a spatiotemporal KDE $p(\x,t)$ with a product kernel $K_{\x}(\x-\x_n;\sigma_{\x}) K_t(t-t_n;\sigma_t)$. The kernel over the spatial variables \x\ is typically Gaussian and the kernel over the temporal variable $t$ considers only past observations and gives more importance to recent ones. At each new value of $t$, one finds modes over \x\ by starting from the modes of the previous time value (as in scale-space theory, but using time rather than scale).

\paragraph{Manifold data}

The original mean shift algorithm is defined on the Euclidean space $\bbR^D$. Sometimes, the data to be clustered lies on a low-dimensional manifold, so the mean-shift iterates and the modes should also lie on this manifold. In some applications (such as motion segmentation, diffusion tensor imaging and other computer vision problems) this is a known Riemannian manifold, such as rotation matrices, Grassmann manifolds or symmetric positive definite matrices \cite{Berger03a}. Subbarao and Meer \cite{SubbarMeer09a} extended mean-shift to Riemannian manifolds by defining the squared distances appropriately with respect to the manifold (although this does not result in a proper KDE on the manifold). However, most times the manifold is not known a priori. Shamir et al.\ \cite{Shamir_06a} extended mean-shift to 3D point clouds in computer graphics applications, by constraining the mean shift steps to lie on the surfaces of a triangulated mesh of the data. The Laplacian $K$-modes described later uses a different approach to find modes that are valid patterns with data lying on nonconvex manifolds.

\paragraph{Graph data}

The mean-shift algorithm operates on data where each data point $\x_n$ is defined by a feature vector in $\bbR^D$. In some applications, the data is best represented as a weighted graph, where each vertex is a data point $\x_n$, and an edge $(\x_n,\x_m)$ represents a neighborhood relation between a pair of data points, with a real-valued weight $\delta(\x_n,\x_m)$ representing distance. This allows one to work with data that need not live in a Euclidean space, and can be used to represent distances along a manifold (for example, by approximating geodesics through shortest paths over the graph).

Some algorithms have been proposed for graph data that are based on a KDE. They work by assigning to each data point a ``parent'' in the graph, which is another data point in their neighborhood having a higher density (roughly speaking). This results in a forest of directed trees that spans the graph, whose roots are ``modes'' in the sense of not having any neighboring point with higher density. Each tree corresponds to a cluster. The parent of a data point is defined as the data point that optimizes a criterion based on the KDE and the distances between data points. The crucial aspect is that this optimization is defined only over the $N$ data points rather than over the space $\bbR^D$ (as happens with $k$-medoid algorithms).

One of the earliest approaches is the algorithm of \cite{Koontz_76a}. This uses as criterion $(p(\x) - p(\x_n))/\delta(\x,\x_n)$, which is a numerical approximation to the directional gradient of the KDE. This is maximized over the data points $\x\in\{\x_1,\dots,\x_N\}$ that are in a neighborhood of $\x_n$ (e.g.\ the $k$ nearest neighbors or the points within a distance $\epsilon$). The output of this algorithm can be improved with cluster-merging based on topological persistence \cite{Chazal_13a}.

The medoidshift algorithm of \cite{Sheikh_07a} is more directly related to mean-shift. It uses as criterion the Riemannian center of mass (sometimes called Fr{\'e}chet mean \cite{Afsari11a,Pennec_06a}) $\f(\x) = \arg\min_{\x \in \{\x_1,\dots,\x_N\}}{\smash{\sum^N_{n=1}{ w_n \delta(\x,\x_n) }}}$, where $w_n = \smash{p(\x^{(\tau)},\x_n)} \propto K\bigl(\smash{\norm{(\x^{(\tau)}-\x_n)/\sigma}}^2\bigr)$ is evaluated at the current iterate $\smash{\x^{(\tau)}}$. The algorithm then alternates computing the Riemannian center of mass (over the data points) with updating the weights. We recover mean-shift by using the squared Euclidean distance for $\delta$ and optimizing over $\bbR^D$. Unlike mean-shift (which finds maxima of the KDE), this does not seem to maximize a global objective function, but rather maximizes a local objective at each iteration and data point. The algorithm gives better clusterings and is faster in its ``blurring'' version, where at every iteration the entire dataset is updated. In matrix form (using the random-walk matrix \PP\ and the matrix $\bDelta = (\delta(\x_n,\x_m))_{nm}$ of pairwise distances, and updating all $N$ points synchronously), this takes the form $\X \leftarrow \arg\min_{\x_1,\dots,\x_N}{(\bDelta \PP)}$, where the minimization is columnwise. Hence, medoidshift can be seen as a discrete filter, while the BMS iteration $\X \leftarrow \X \PP$ is a continuous filter. As in the accelerated BMS, each iteration contracts the graph, reducing the number of points. It is possible to define other variations by using different types of distances $\delta$, such as the local Tukey median in the medianshift algorithm of \cite{Shapir_09a}.

\paragraph{Other extensions}

The mean-shift update as originally proposed \cite{FukunagHostet75a} results from estimating the gradient of a density in two steps: first, one estimates the density with a KDE, and then one differentiates this. Sasaki et al.\ \cite{Sasaki_14a} directly estimate the gradient of the log-density by using score matching \cite{Hyvaer05a} and derive a mean-shift algorithm based on this (by equating the gradient to zero and obtaining a fixed-point iteration). In score matching, one does a least-squares fit to the true log-density gradient using a model. If the model is a linear combination of basis functions (one per data point), one obtains an update that is identical to the mean-shift update of eq.~\ref{e:GMS} but with weighted kernels (although, since these weights can be negative, this fixed-point iteration is not guaranteed to converge). A faster, parametric method is obtained by using fewer basis functions than data points. They observe better results than the original mean-shift with high-dimensional data. Note that, as seen in eq.~\eqref{e:GMS-adaptive}, using an adaptive KDE (e.g.\ with bandwidths obtained using the entropic affinities) also gives a weighted mean-shift update, but the weights obey a different criterion.

Ranking data consists of permutations of a given set of items, and are discrete. Meil\u{a} and Bao \cite{MeilaBao10a} extended blurring mean-shift to ranking data by using the Kendall distance rather than the Euclidean distance, and by rounding the continuous average of eq.~\eqref{e:GMS} to the nearest permutation after each iteration (so the algorithm stops in a finite number of steps).

Functional data, such as a collection of curves or surfaces, live in an infinite-dimensional space. Mean-shift clustering can be extended to functional data, where it corresponds to a form of adaptive gradient ascent on an estimated surrogate density \cite{Ciollar_14a}.

\subsection{Extensions of mean-shift beyond clustering: manifold denoising}
\label{ch4.1sec:denoising}

\begin{figure*}[t]
  \begin{center}
    \begin{tabular}[c]{@{}c@{\hspace{0.005\linewidth}}c@{}c@{}c@{}c@{}c@{}c@{}c@{}c@{}}
      & $\tau=0$ & $\tau=1$ & $\tau=2$ & $\tau=3$ & $\tau=5$ & $\tau=10$ & $\tau=20$ & $\tau=60$ \\
      \raisebox{0.062\linewidth}{\rotatebox{90}{\makebox[0pt]{MBMS}}} &
      \includegraphics[width=0.1228\linewidth]{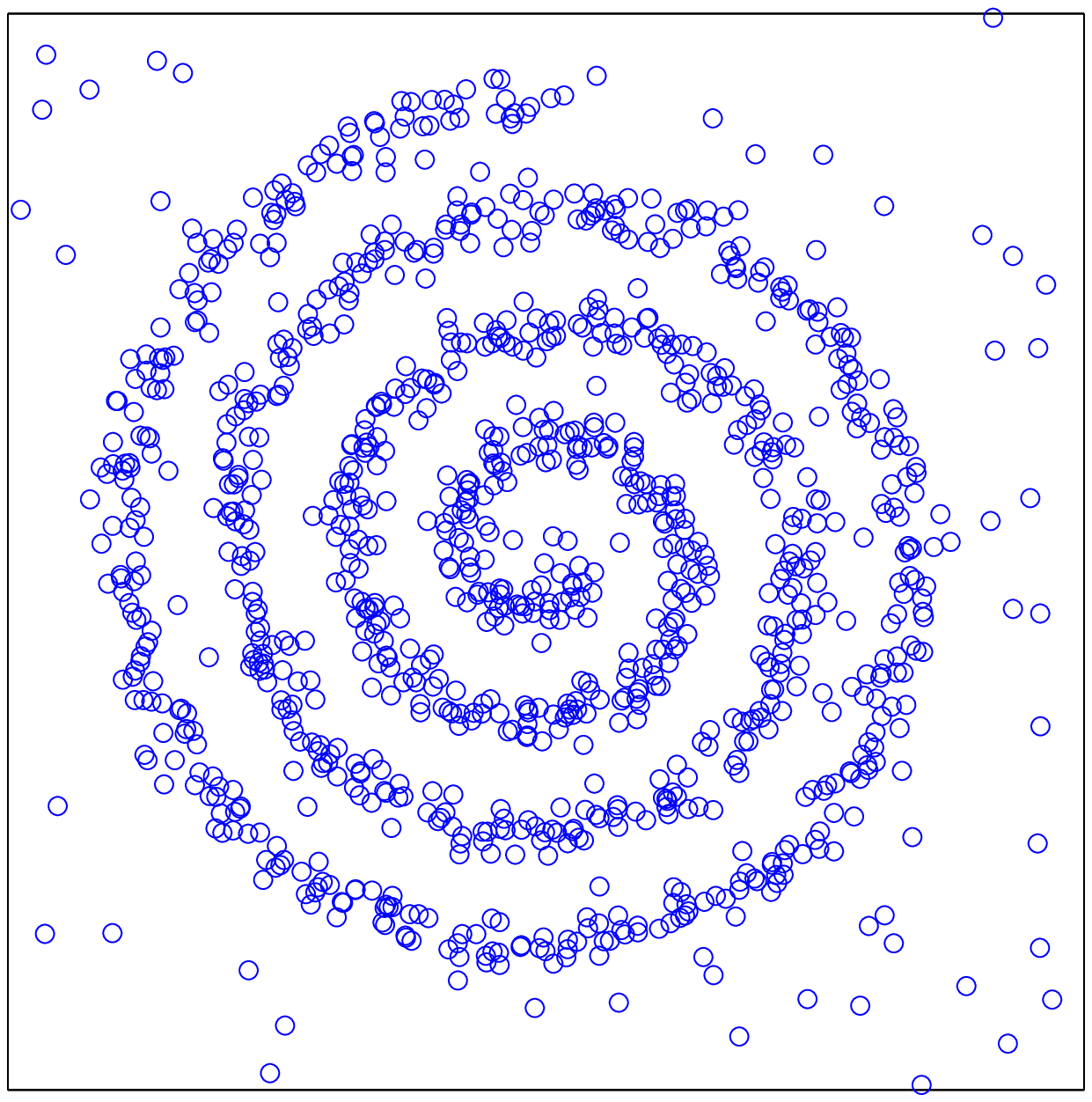} &
      \includegraphics[width=0.1228\linewidth]{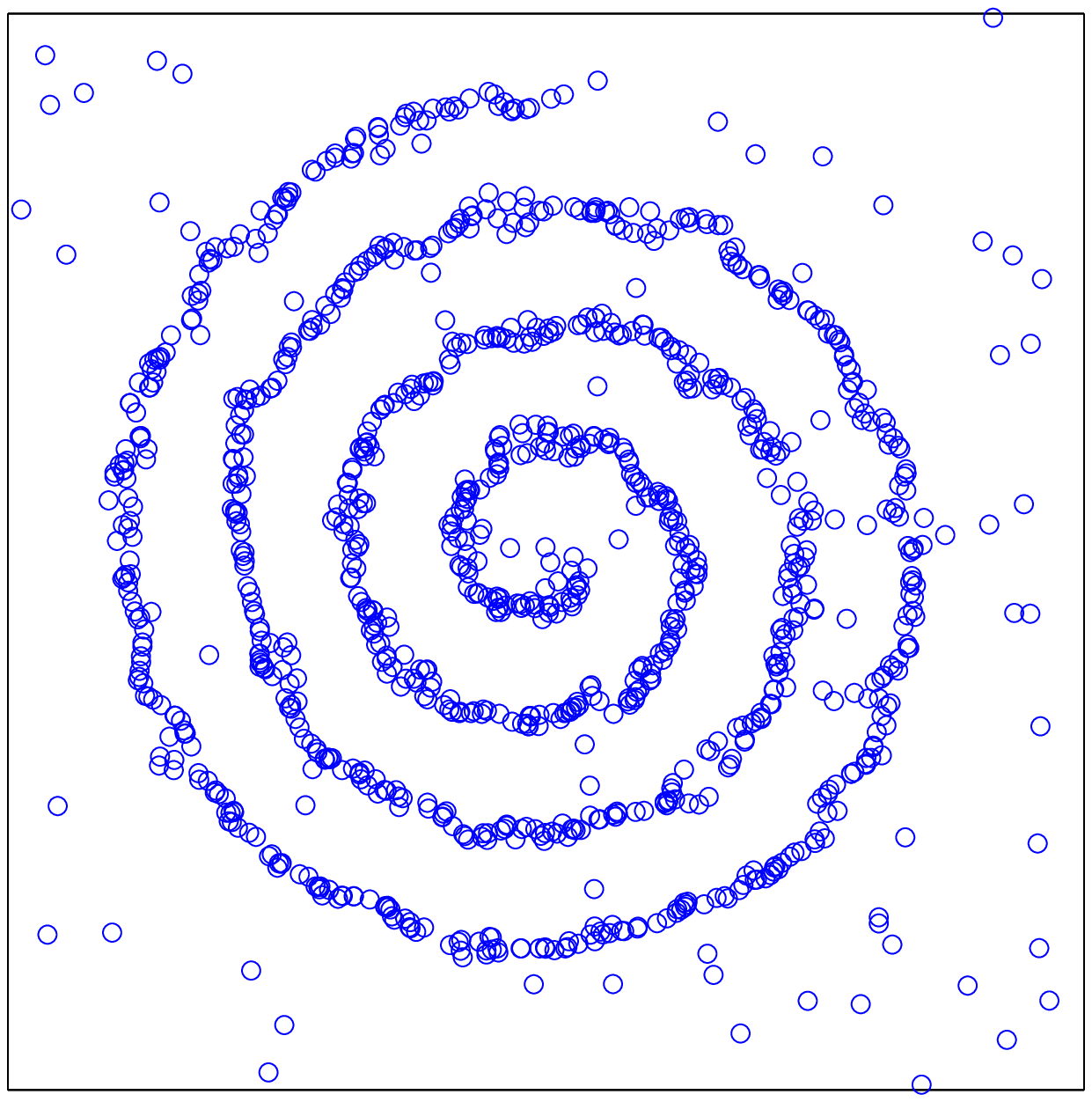} &
      \includegraphics[width=0.1228\linewidth]{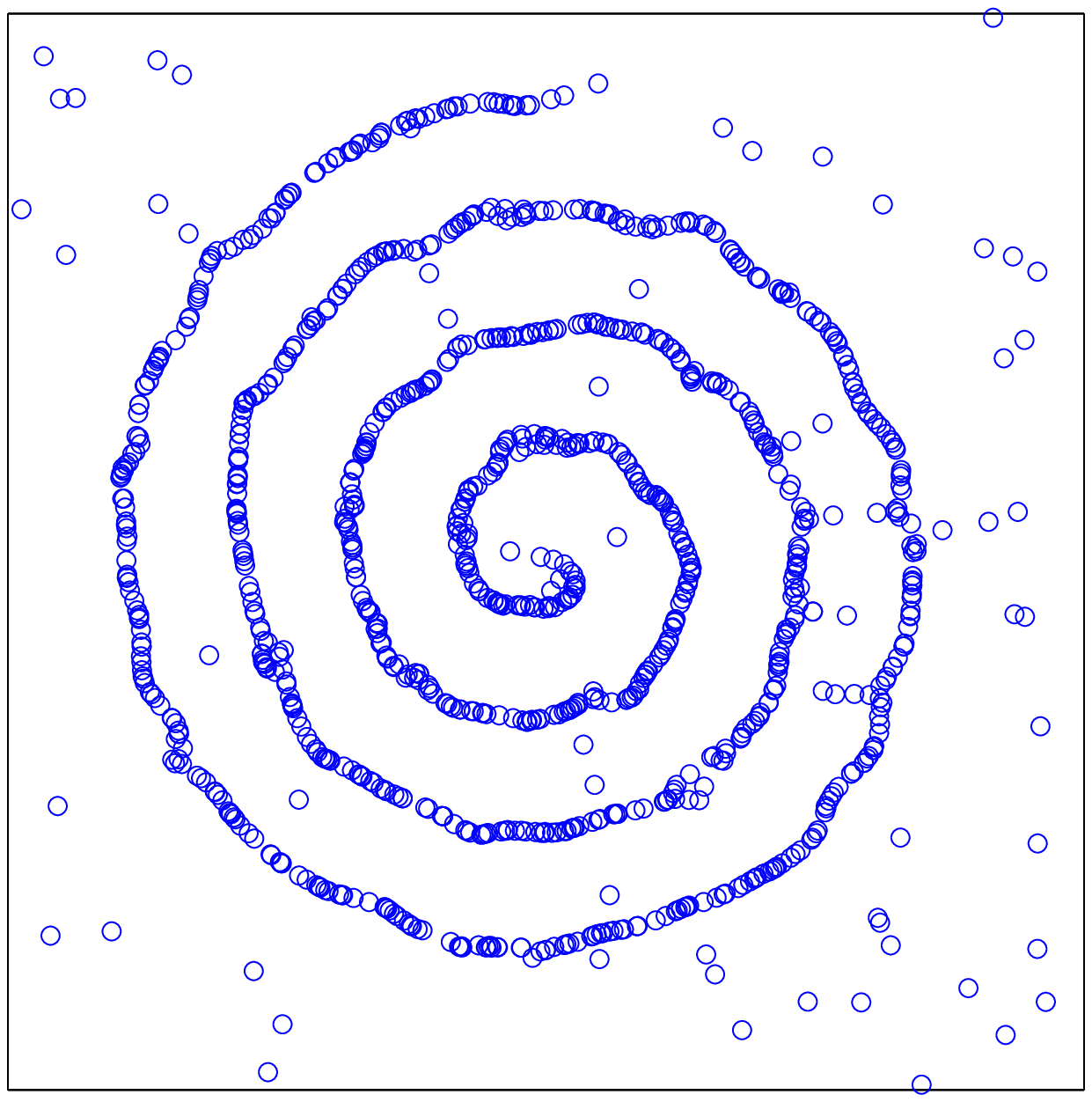} &
      \includegraphics[width=0.1228\linewidth]{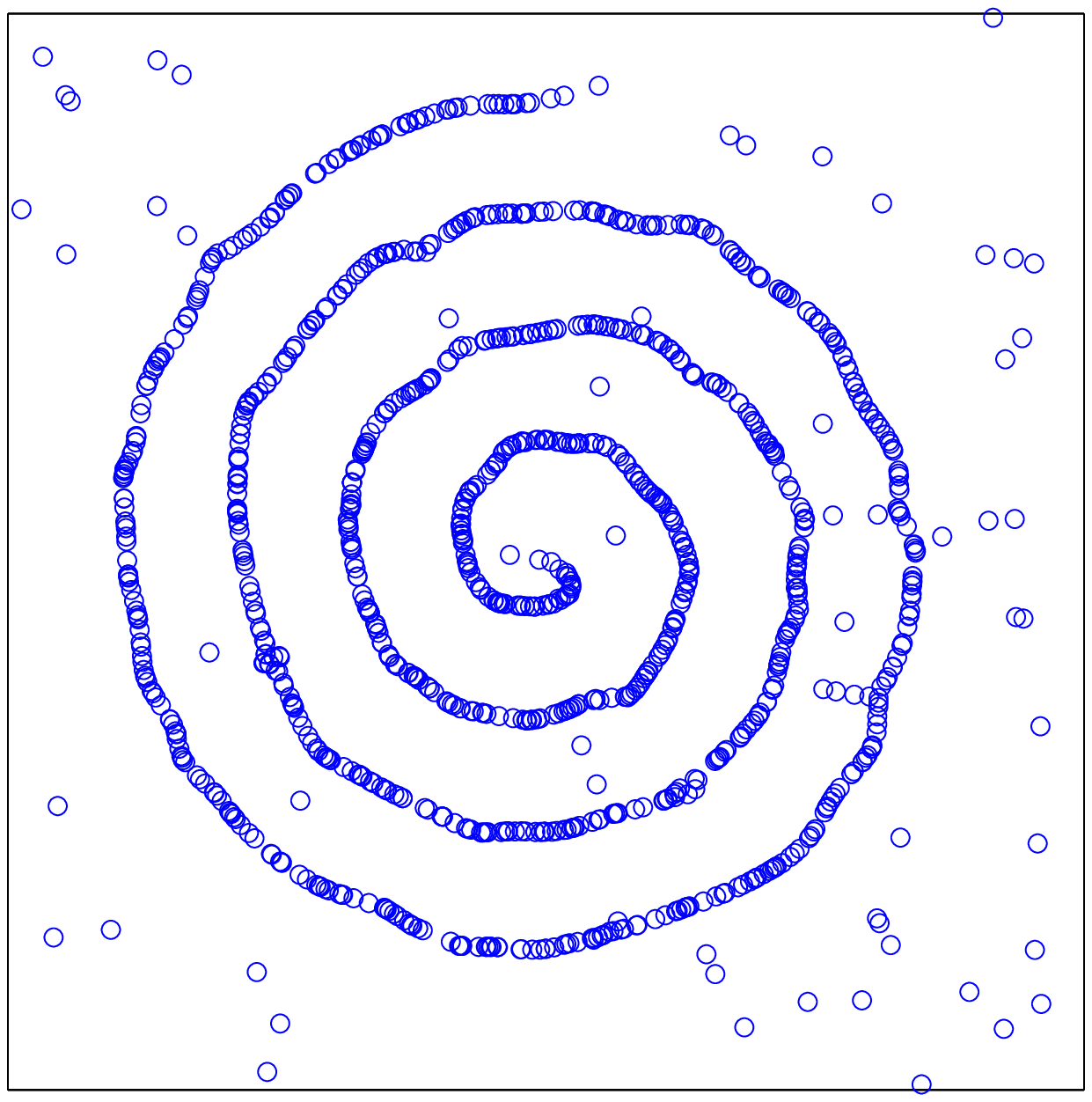} &
      \includegraphics[width=0.1228\linewidth]{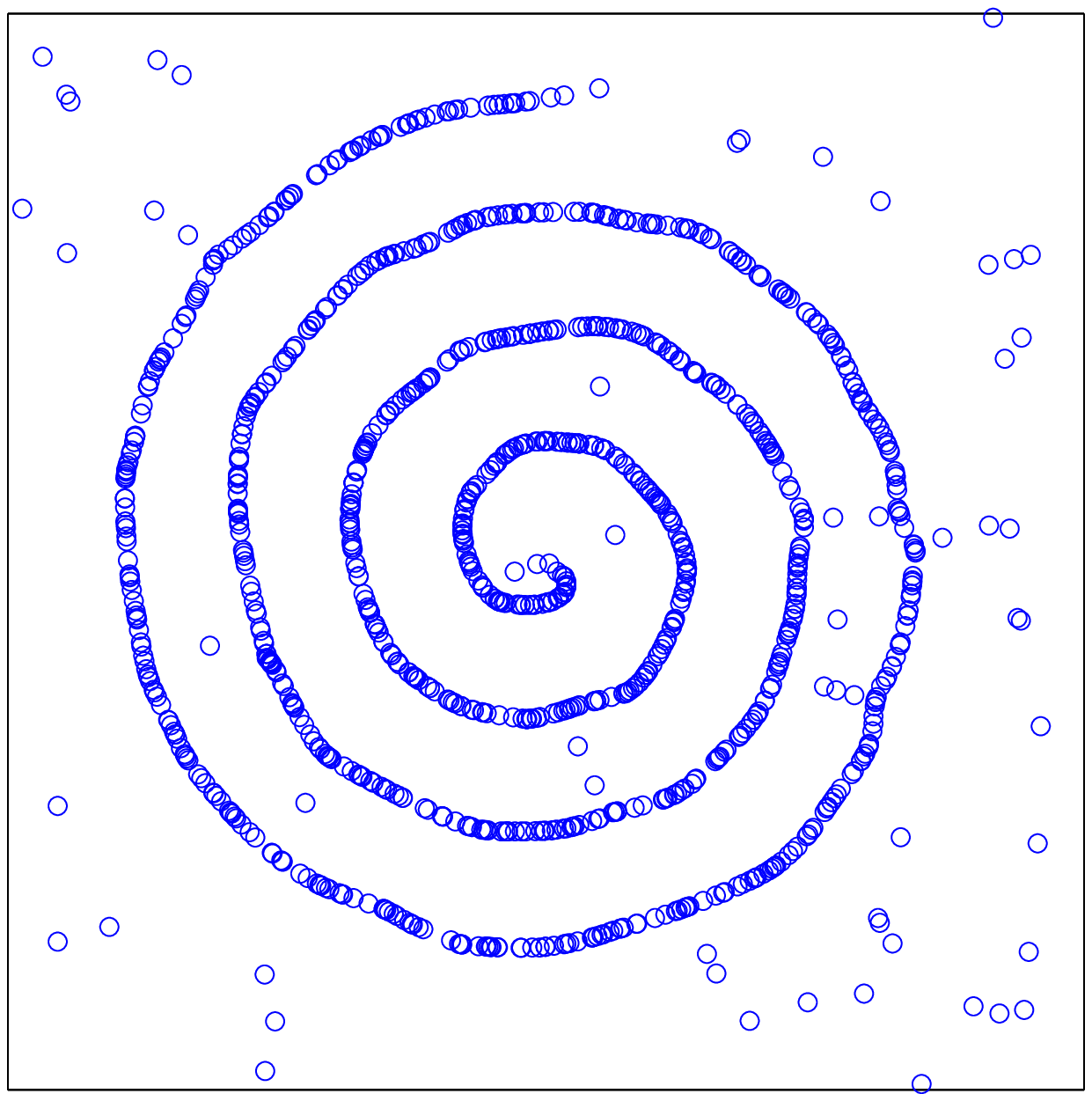} &
      \includegraphics[width=0.1228\linewidth]{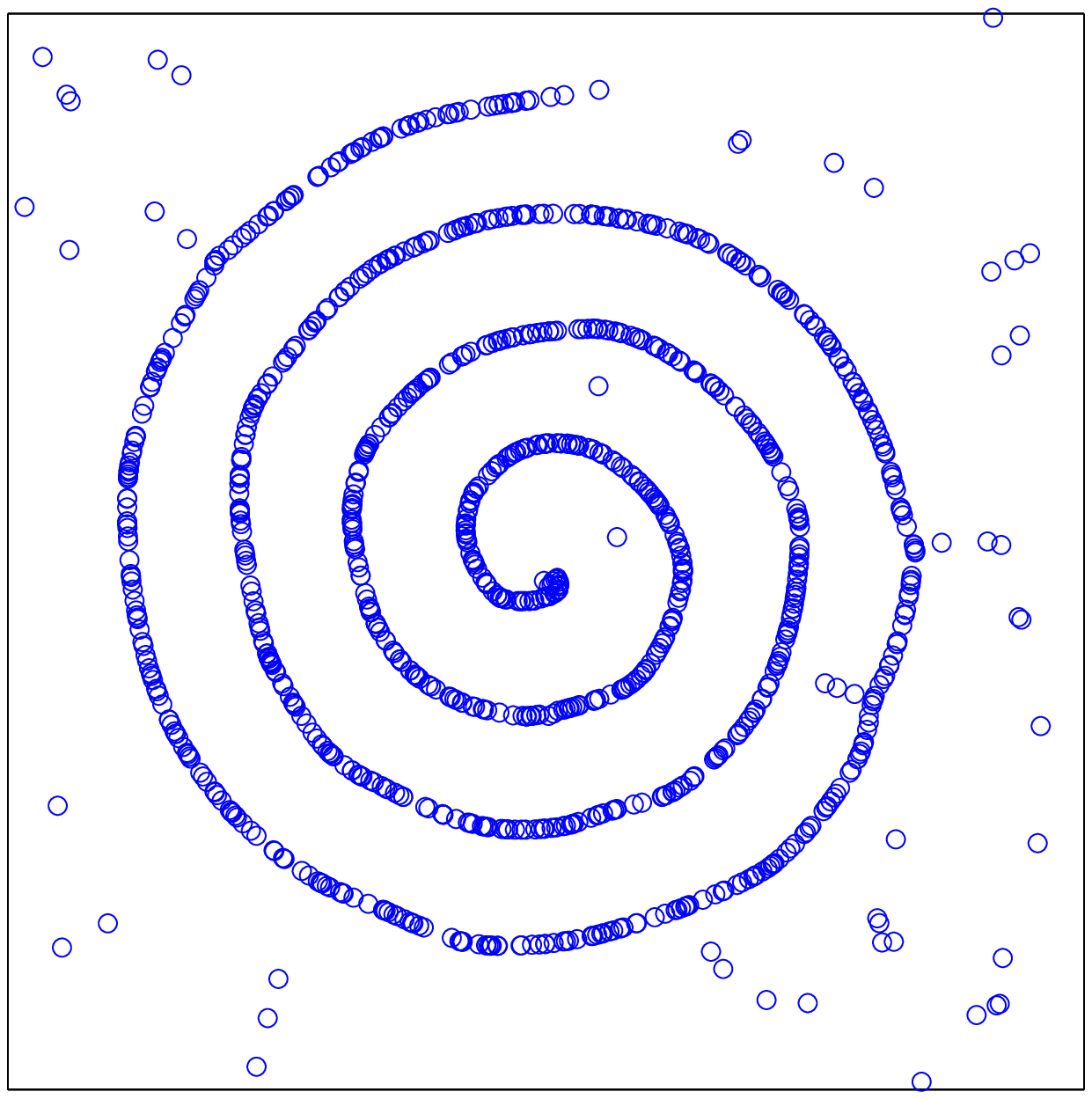} &
      \includegraphics[width=0.1228\linewidth]{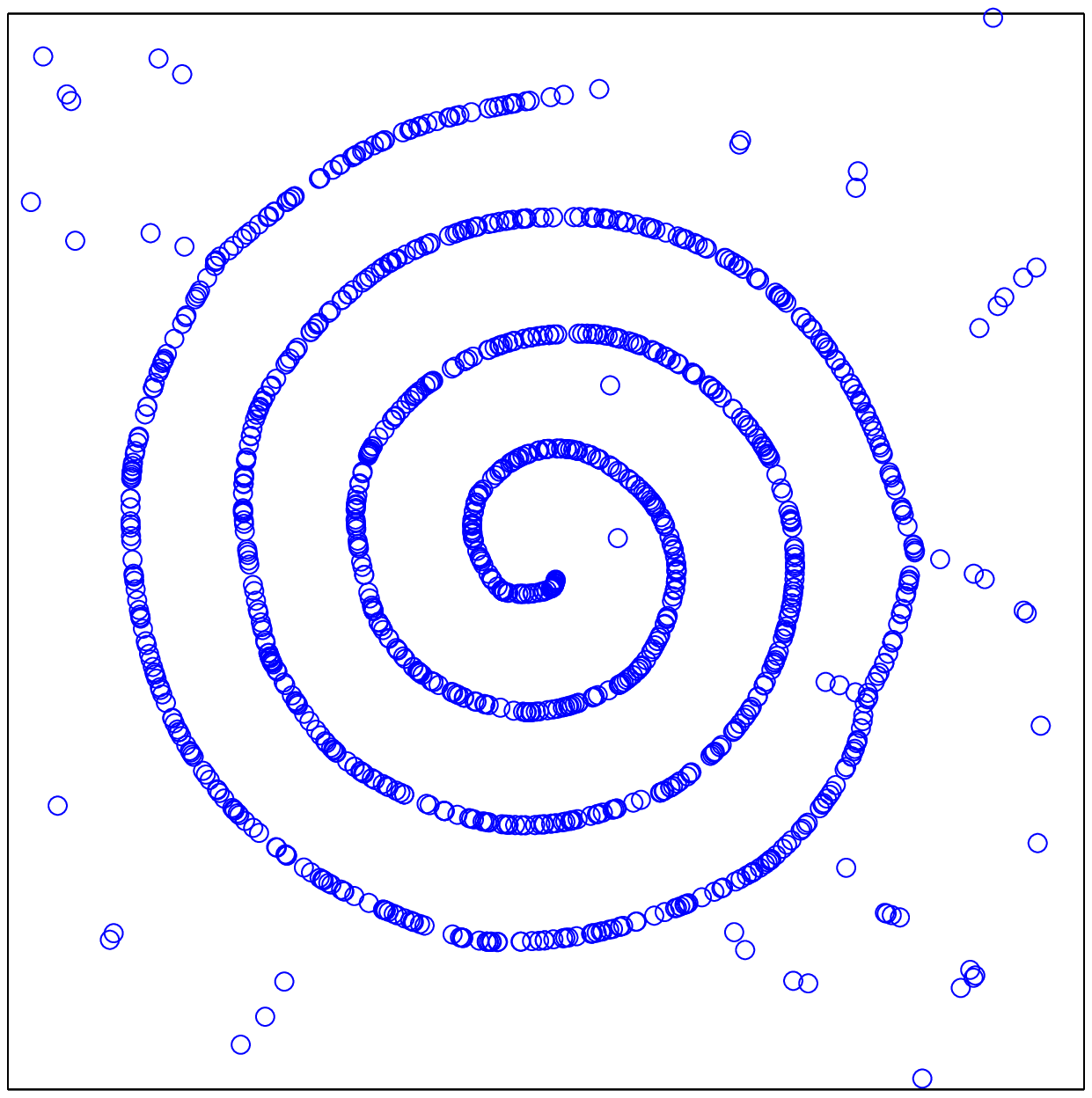} &
      \includegraphics[width=0.1228\linewidth]{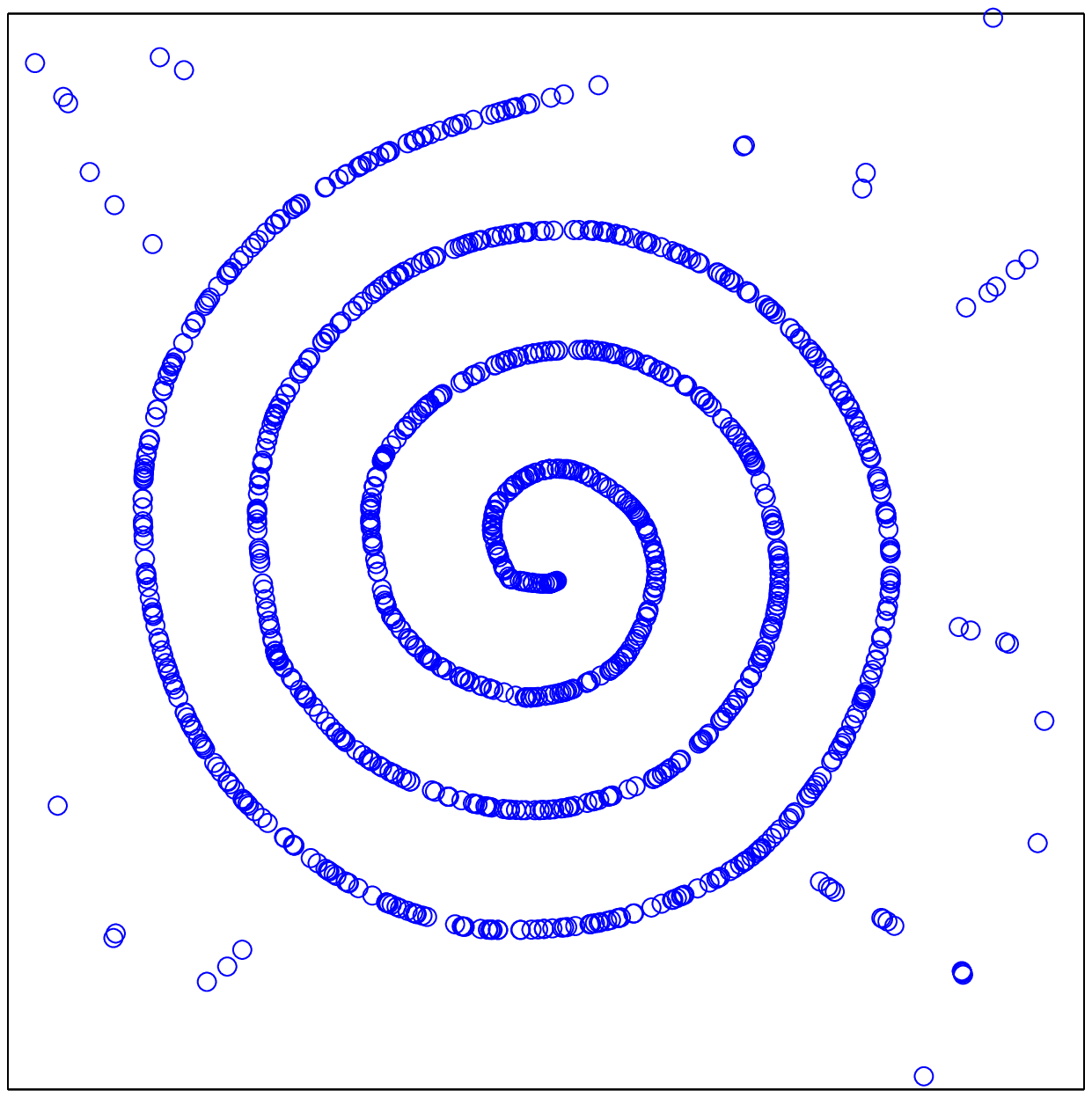} \\
      \raisebox{0.062\linewidth}{\rotatebox{90}{\makebox[0pt]{BMS}}} &
      \includegraphics[width=0.1228\linewidth]{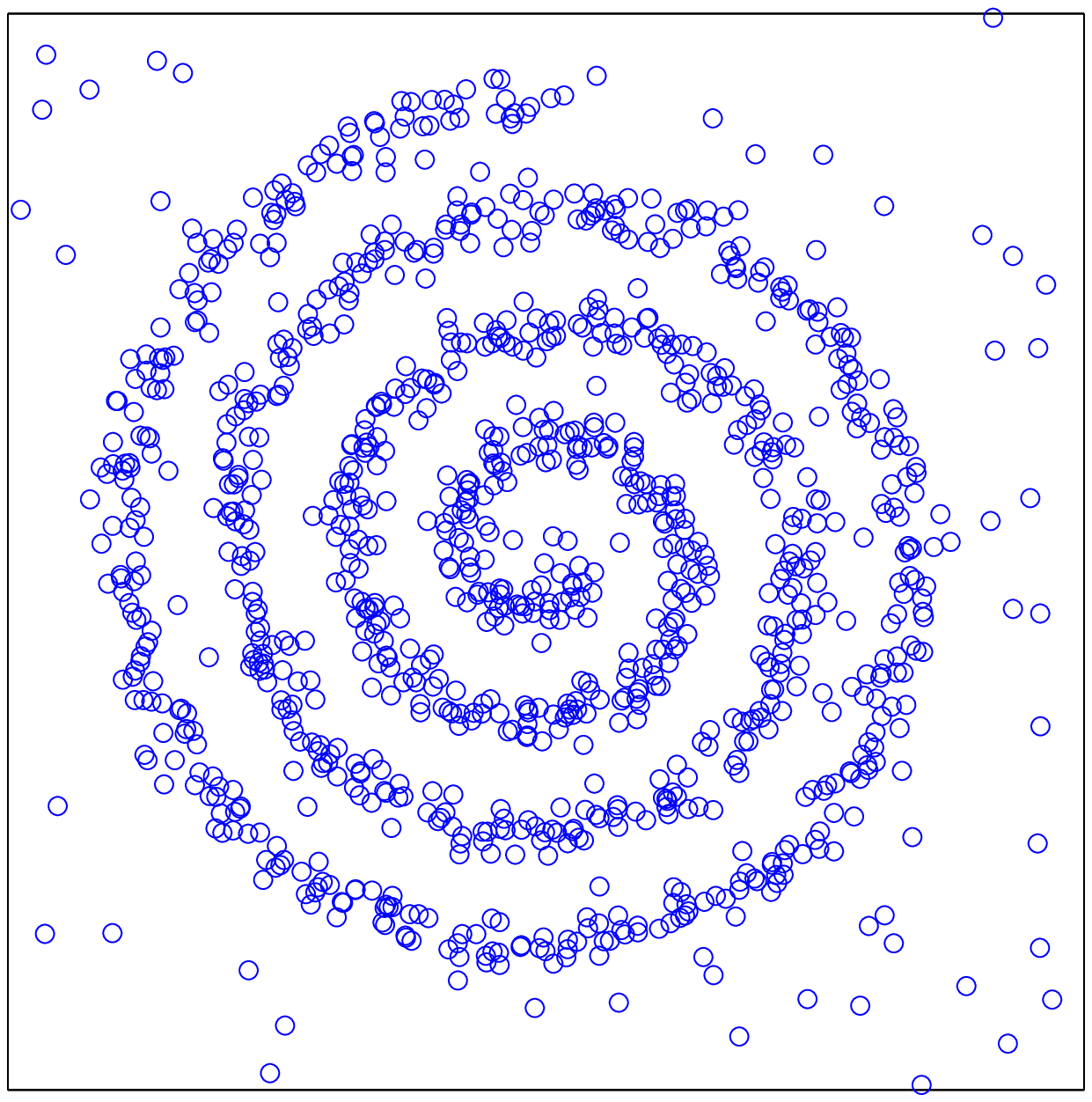} &
      \includegraphics[width=0.1228\linewidth]{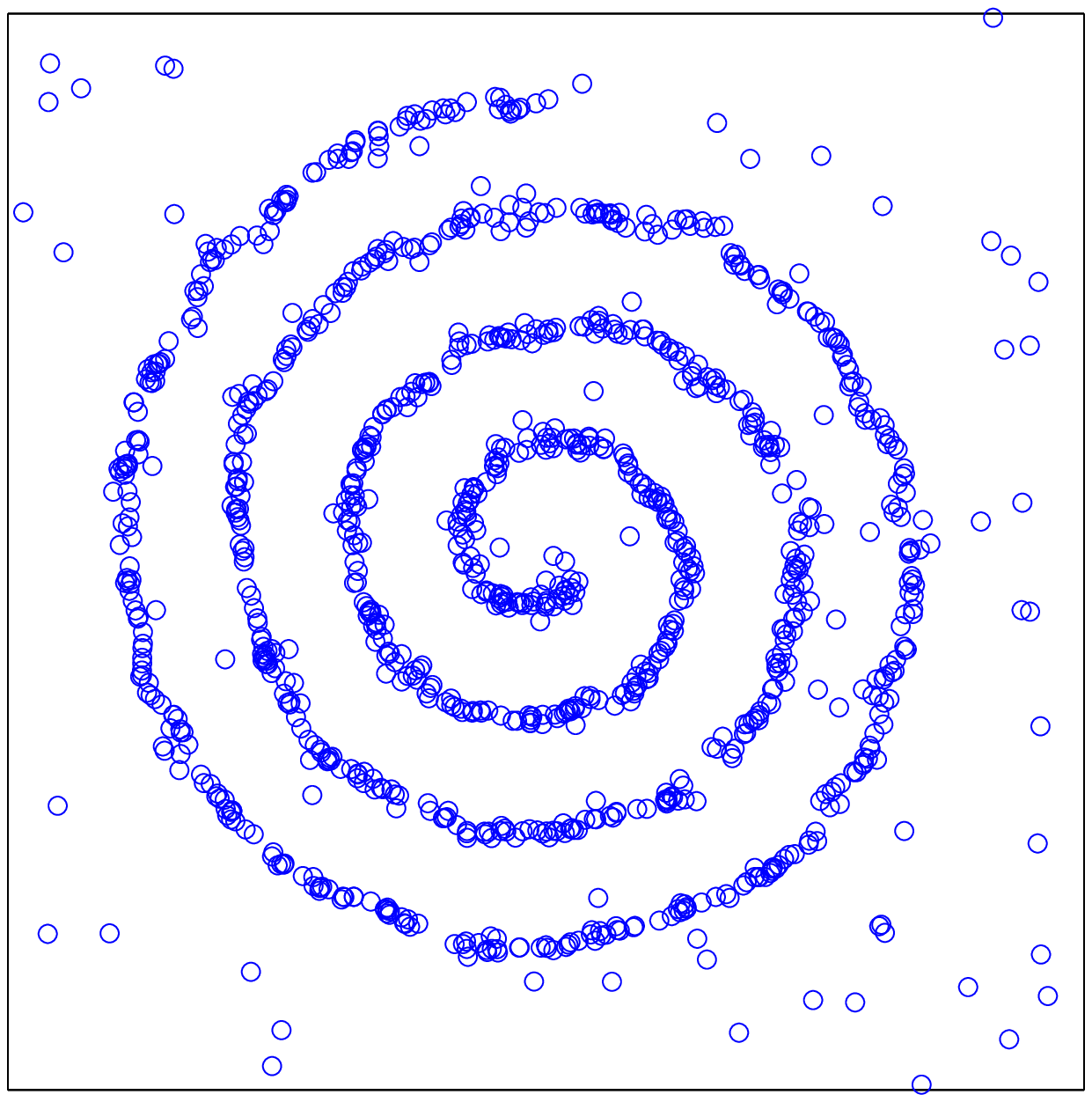} &
      \includegraphics[width=0.1228\linewidth]{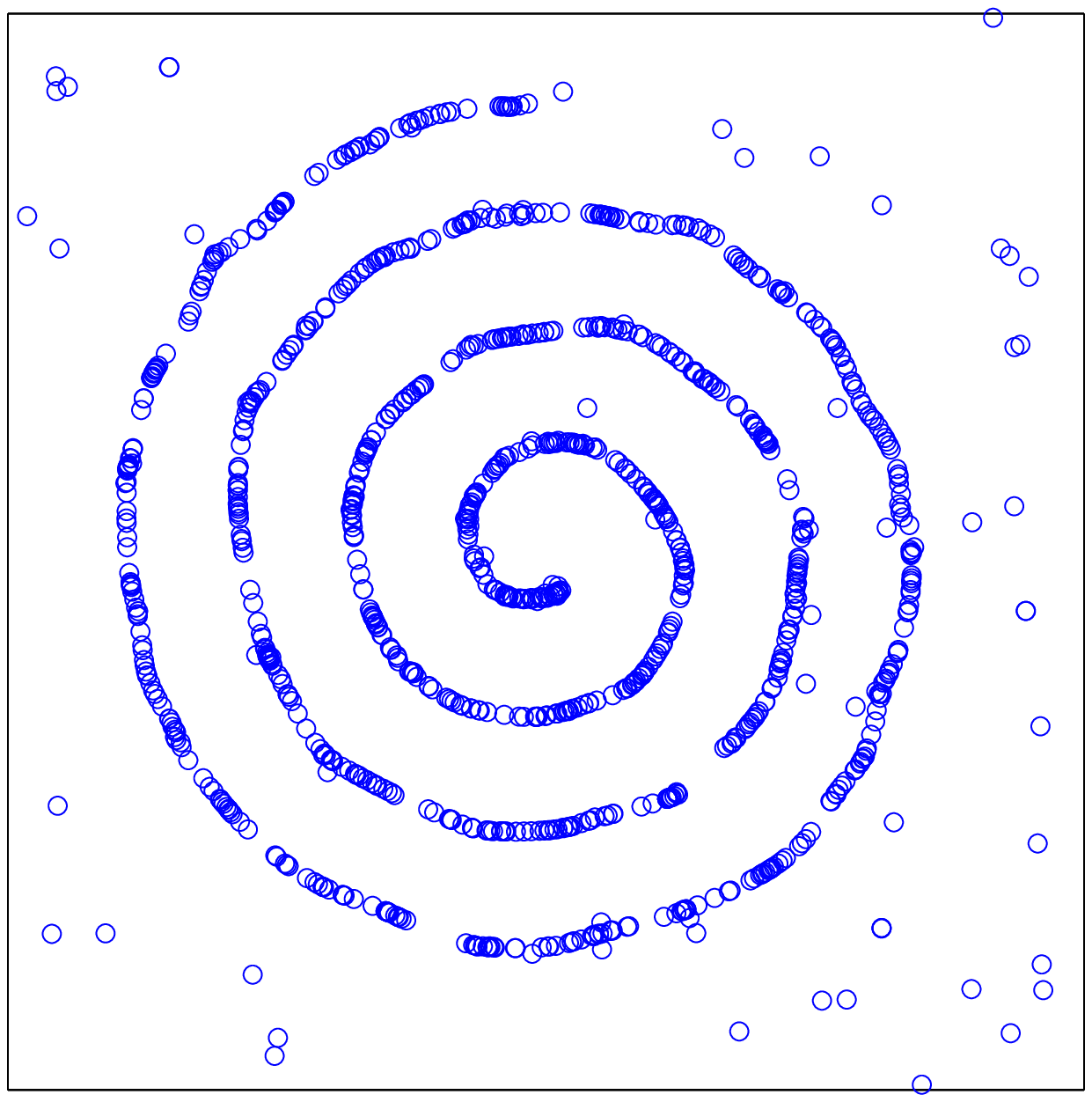} &
      \includegraphics[width=0.1228\linewidth]{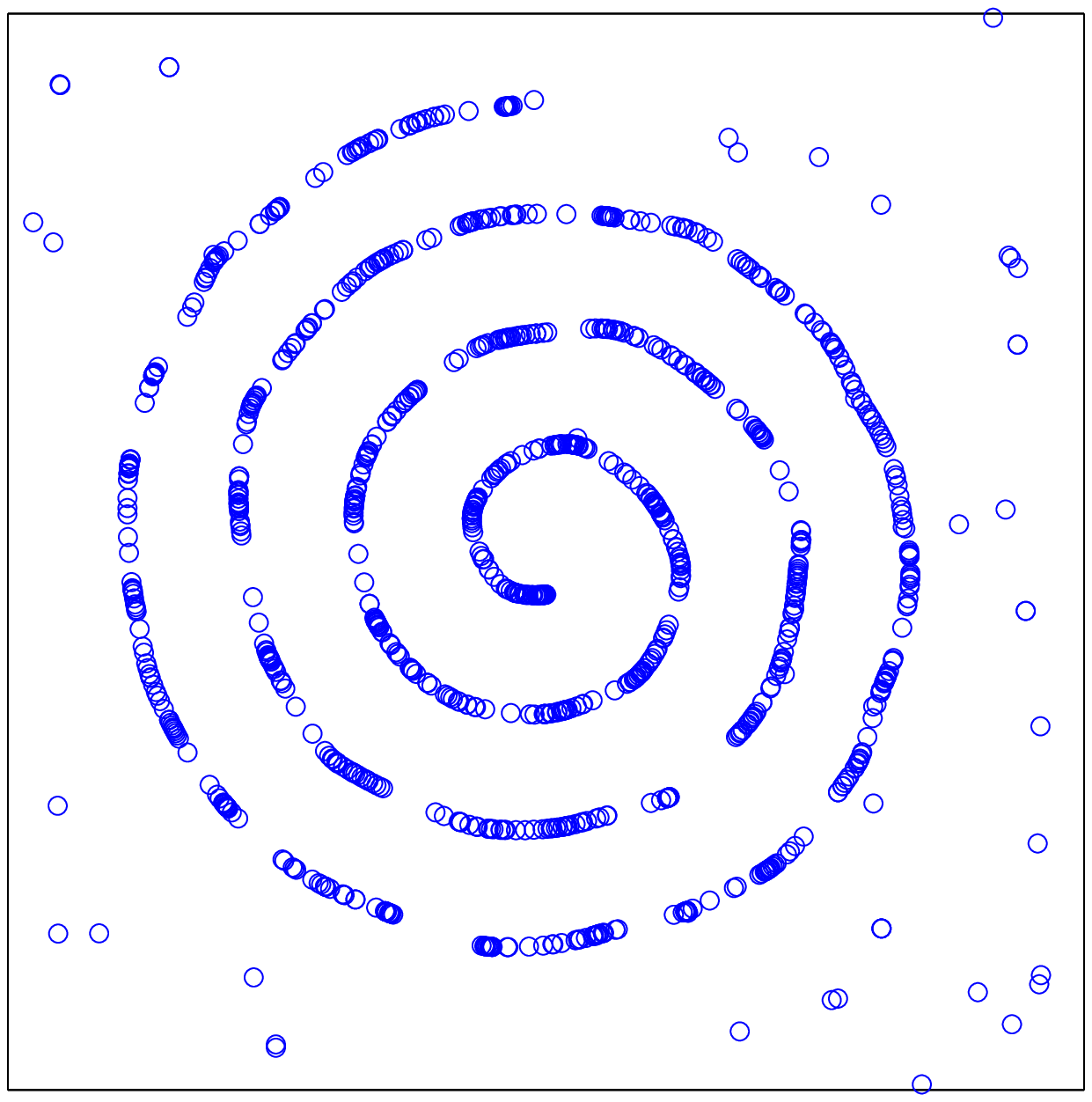} &
      \includegraphics[width=0.1228\linewidth]{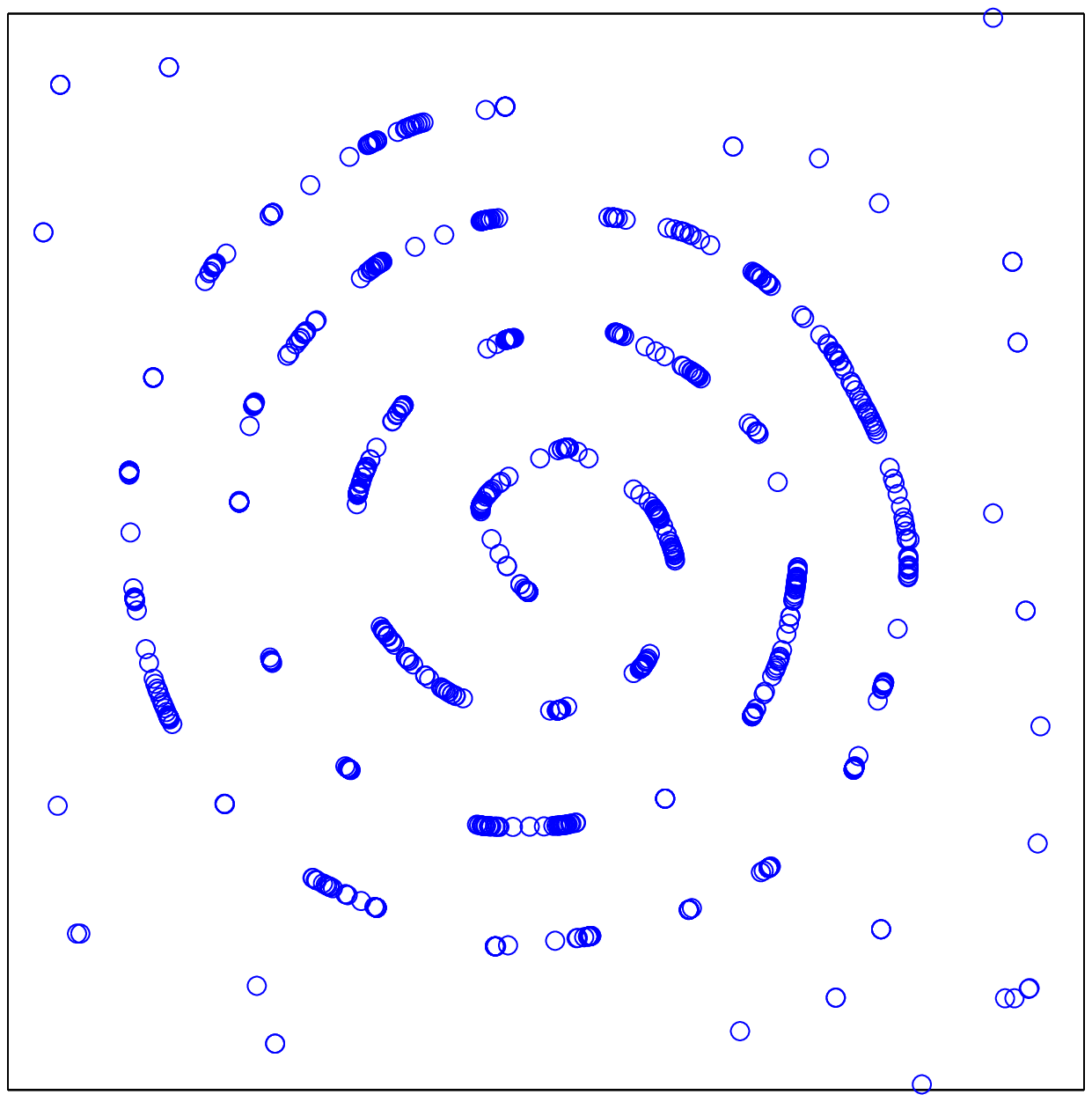} &
      \includegraphics[width=0.1228\linewidth]{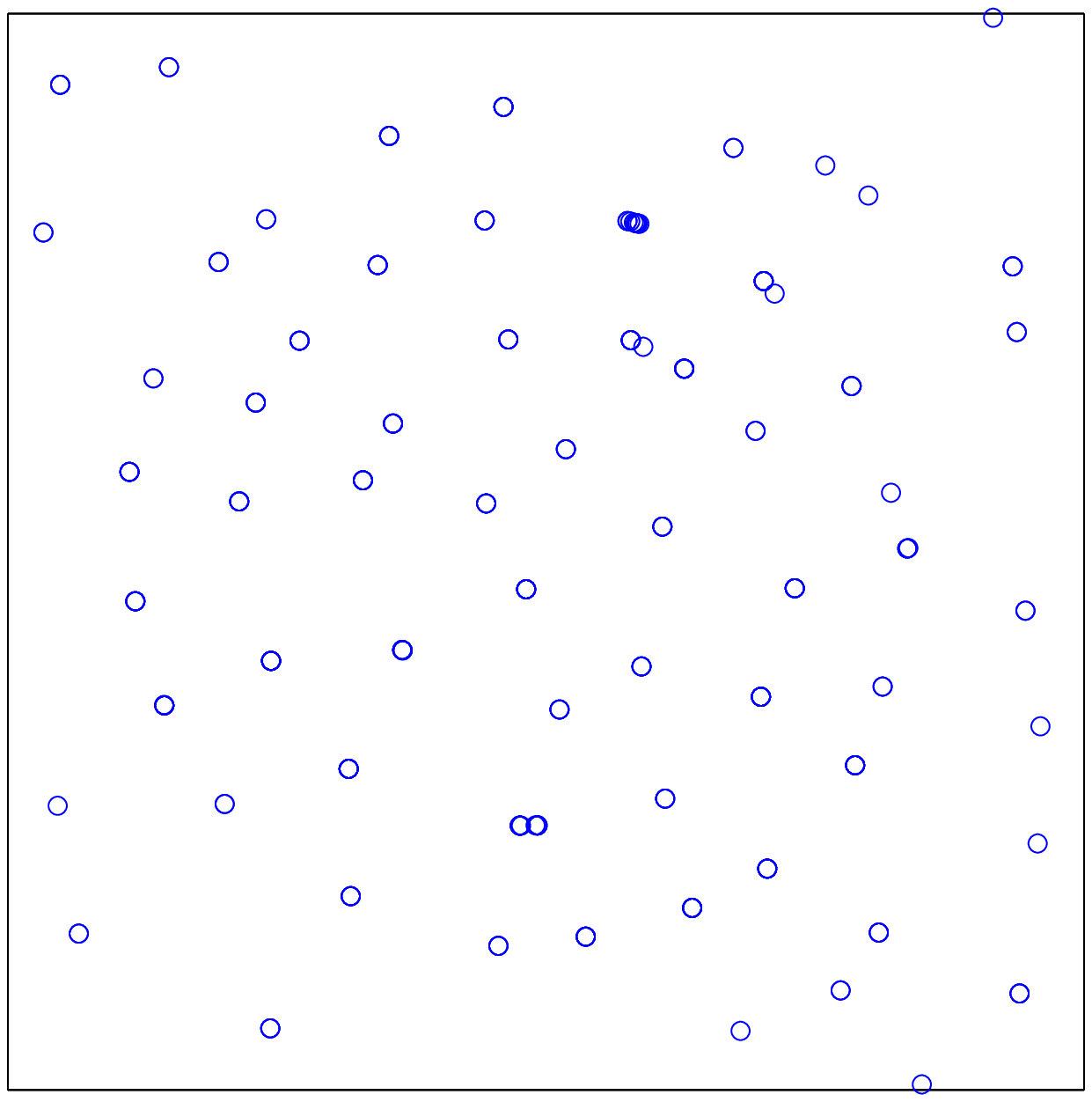} &
      \includegraphics[width=0.1228\linewidth]{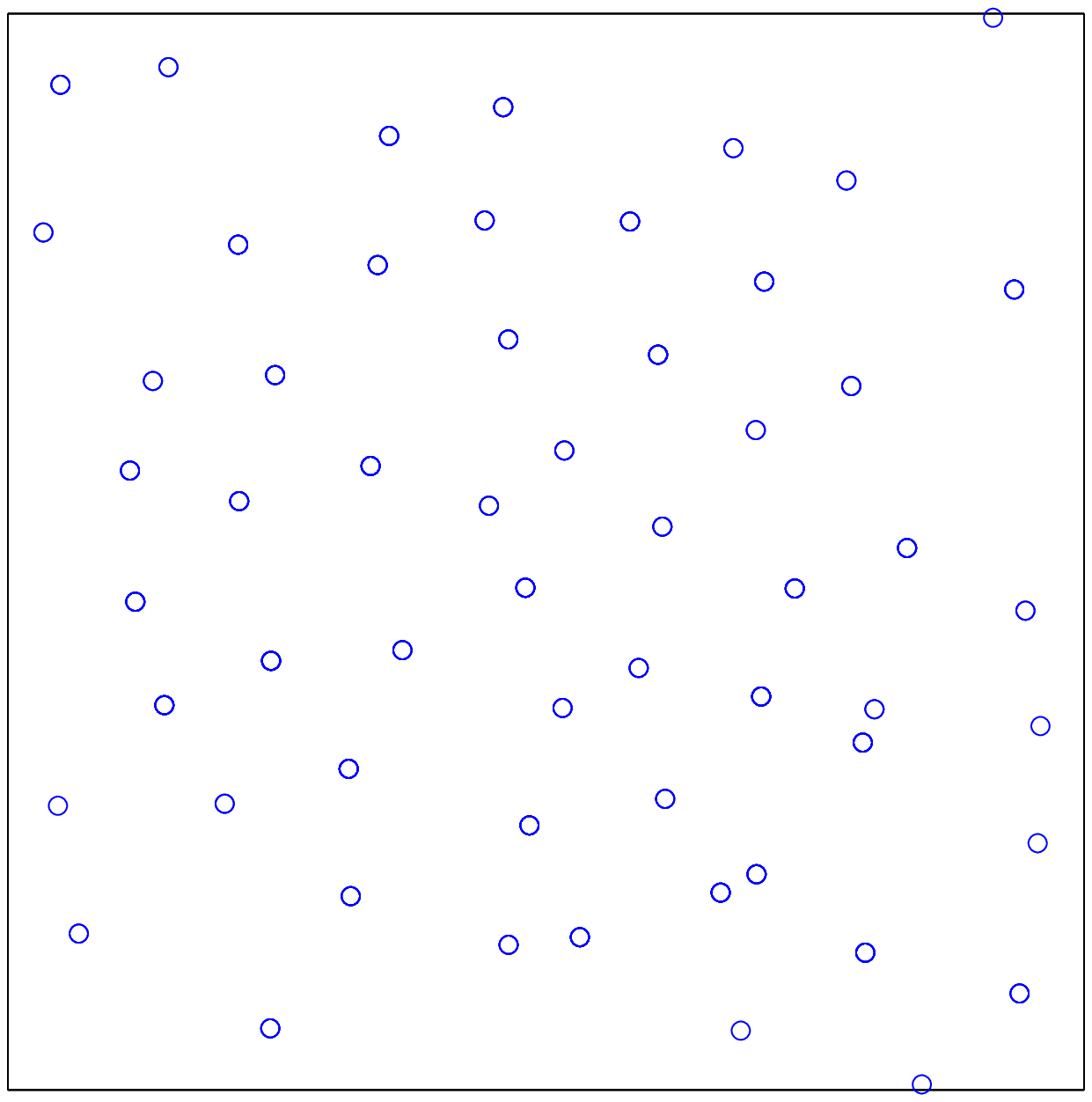} &
      \includegraphics[width=0.1228\linewidth]{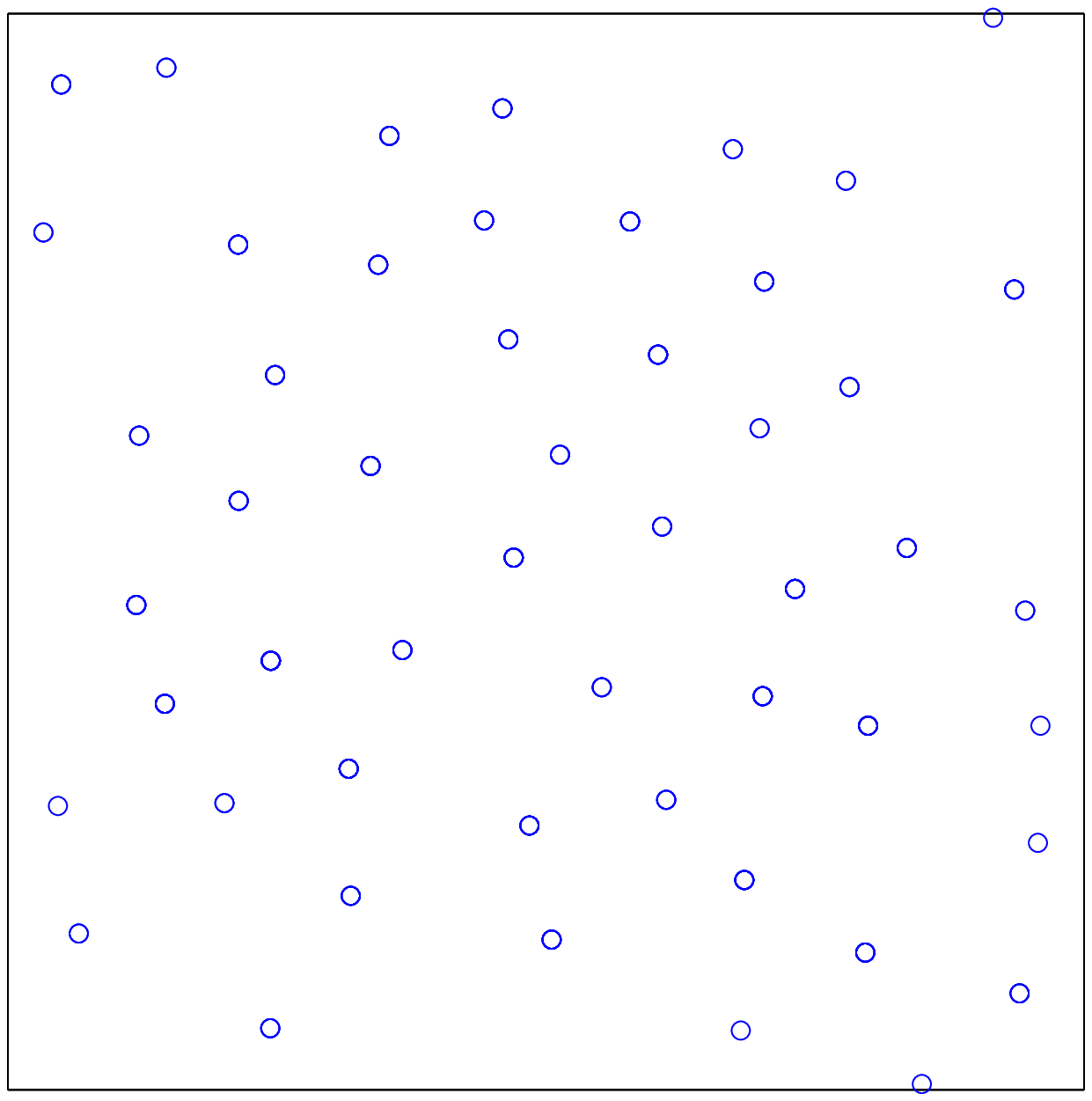}
    \end{tabular}
    \caption{Denoising a spiral with outliers over iterations ($\tau=0$ is the original dataset). Each box is the square $[-30,30]^2$, where $100$ outliers were uniformly added to an existing $1\,000$-point noisy spiral. Algorithms: MBMS and BMS, both with the same bandwidth $\sigma$ for the Gaussian kernel. MBMS denoises while preserving the spiral structure and ignoring the outliers. BMS locally collapses points onto clusters, destroying the spiral.}
    \label{f:spiral}
  \end{center}
\end{figure*}

The mean-shift iteration~\eqref{e:MS} or~\eqref{e:GMS} is essentially a \emph{local smoothing}, which provides a smooth or denoised version of a data point as a weighted average of its nearby data points. Clustering is just one specific problem that can make use of this smoothing. Here we describe work based on mean-shift for manifold denoising. This was already pointed out in \cite{FukunagHostet75a}. Consider data lying on a low-dimensional manifold in $\bbR^D$, but with noise added (as in fig.~\ref{f:spiral}). If we run one mean-shift iteration for each data point in parallel, and then replace each point with its denoised version, we obtain a denoised dataset, where points have moved towards the manifold. Repeating this eventually compresses the dataset into clusters, and this is the basis of the BMS algorithm, although of course it destroys the manifold structure. However, if we stop BMS very early, usually after one or two iterations, we obtain an algorithm that can remove noise from a dataset with manifold structure.

The denoising ability of local smoothing was noted independently in the computer graphics literature for the problem of surface fairing or smoothing. Earlier, the Laplacian had also been used to smooth finite element meshes \cite{Ho-Le88a}. In surface smoothing, 3D point clouds representing the surface of an object can be recorded using LIDAR, but usually contain noise. This noise can be eliminated by Laplacian smoothing, which replaces the 3D location of each point with the average of its neighbors \cite{Taubin95a,Taubin00a}. The neighbors of each point are obtained from a triangulated graph of the cloud, which is usually available from the scanning pattern of LIDAR. Typically, the Laplacian is constructed without the notion of a KDE or a random-walk matrix, and both the values of its entries and the connectivity pattern are kept constant during iterations.

One important problem of BMS (or Laplacian smoothing) is that points lying near the boundary of a manifold or surface will move both orthogonally to and tangentially along the manifold, away from the boundary and thus shrinking the manifold. While this is good for clustering, it is not for surface smoothing because it distorts the object shape, or for manifold learning because it distorts the manifold structure. For example, in the MNIST dataset of handwritten digit images \cite{Lecun_98a}, motion along the manifold changes the style of a digit (e.g.\ slant, thickness). Various approaches, such as rescaling to preserve the object volume, have been proposed in computer graphics \cite{Desbrun_99a}. In machine learning, the \emph{manifold blurring mean-shift (MBMS)} algorithm \cite{WangCarreir10a} is an extension of BMS to preserve the local manifold structure. Rather than applying to each point the mean-shift vector directly, this vector is corrected to eliminate tangential motion. The mean-shift vector is first projected onto the local tangent space to the manifold at that point (estimated using local PCA on the nearest neighbors of the point), and this projection is removed from the motion. Hence, the motion is constrained to be locally orthogonal to the manifold. Comparing the local PCA eigenvalues between the orthogonal and tangent spaces gives a criterion to stop iterating.

\begin{figure*}[t]
  \begin{center}
  \begin{tabular}[c]{@{}c@{}c@{\hspace{0.0057\linewidth}}c@{}c@{\hspace{0.0057\linewidth}}c@{}c@{\hspace{0.0057\linewidth}}c@{}c@{\hspace{0.0057\linewidth}}c@{}c@{\hspace{0.0057\linewidth}}c@{}c@{\hspace{0.0057\linewidth}}c@{}c@{\hspace{0.0057\linewidth}}c@{}c@{}}

\includegraphics[width=0.06\linewidth]{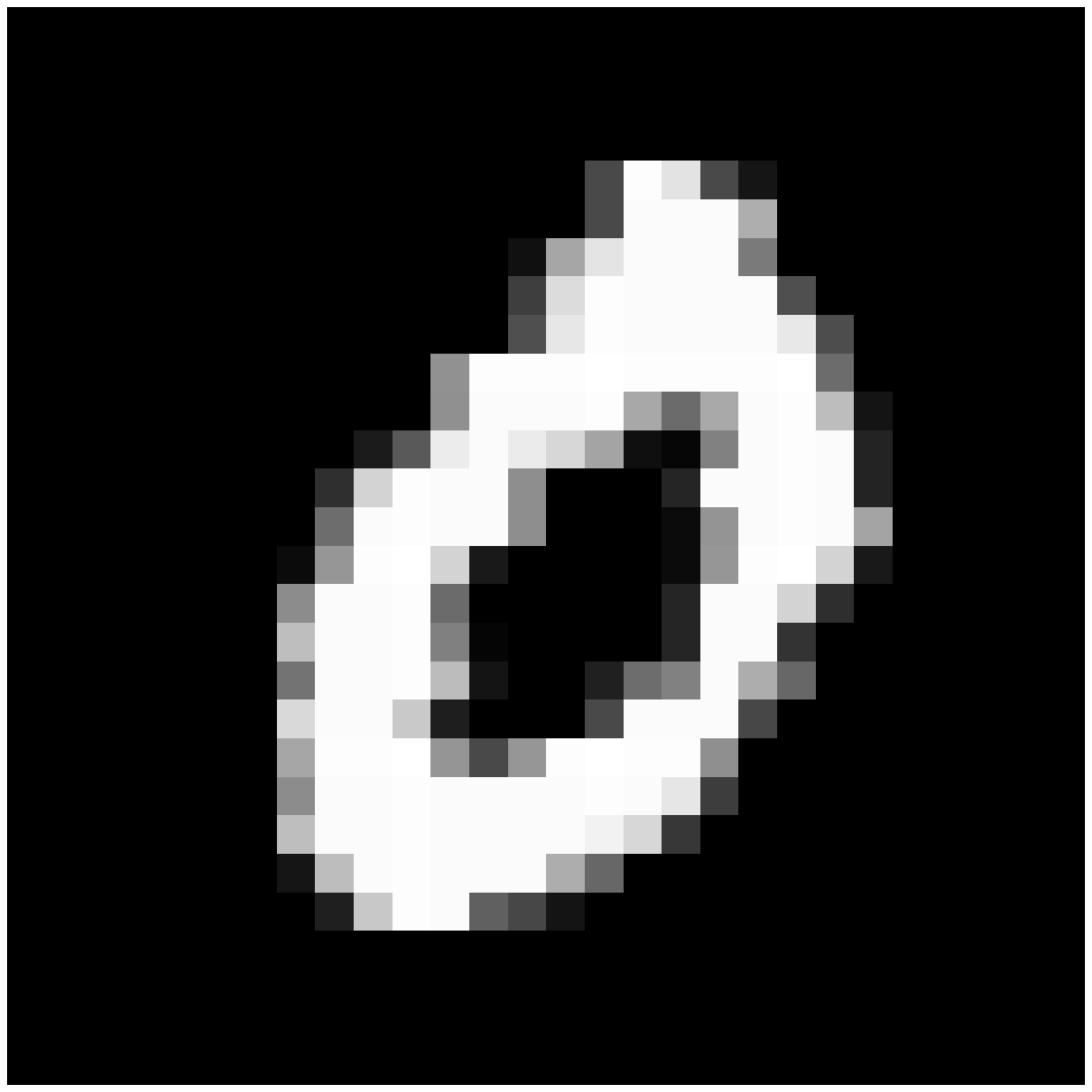}&\includegraphics[width=0.06\linewidth]{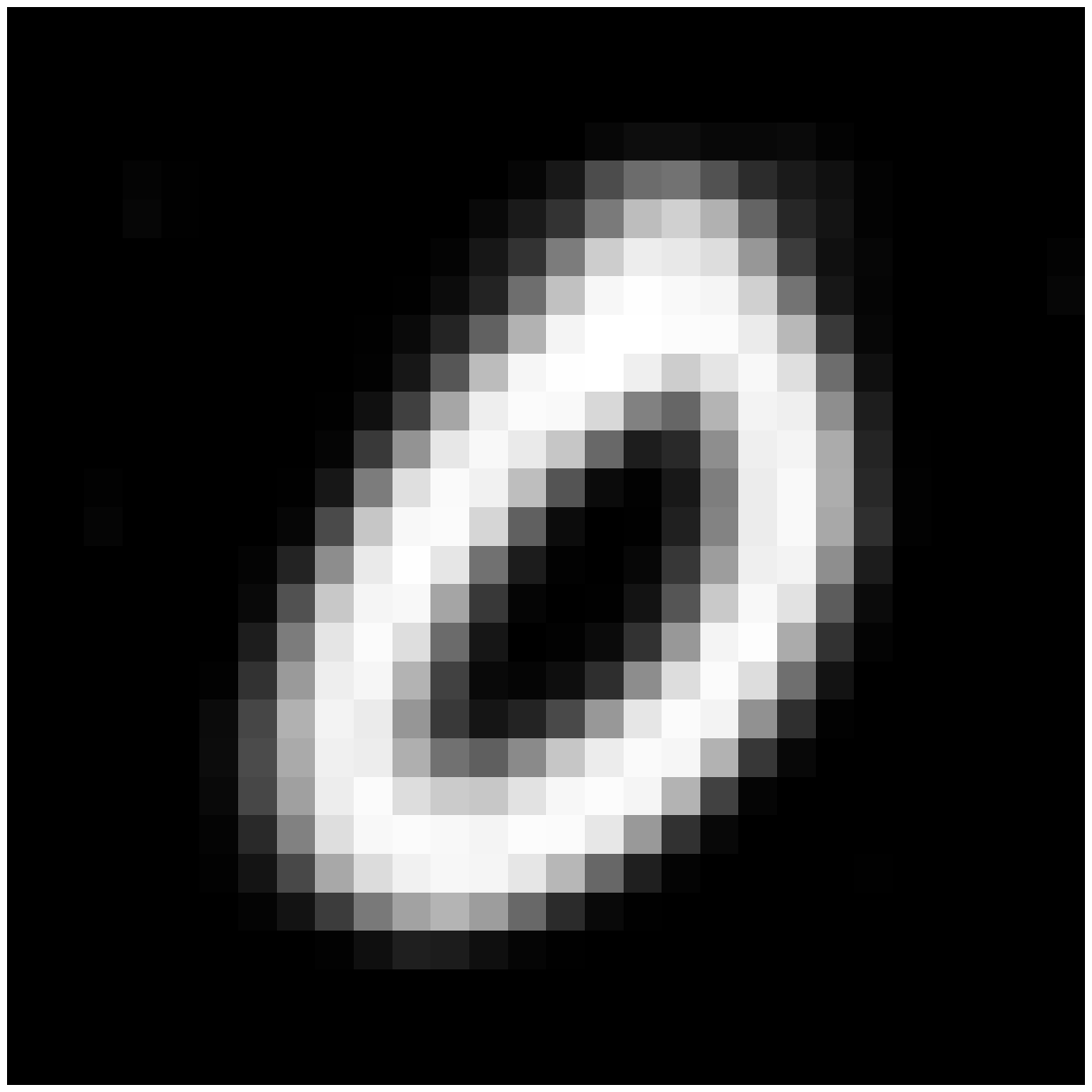} &
\includegraphics[width=0.06\linewidth]{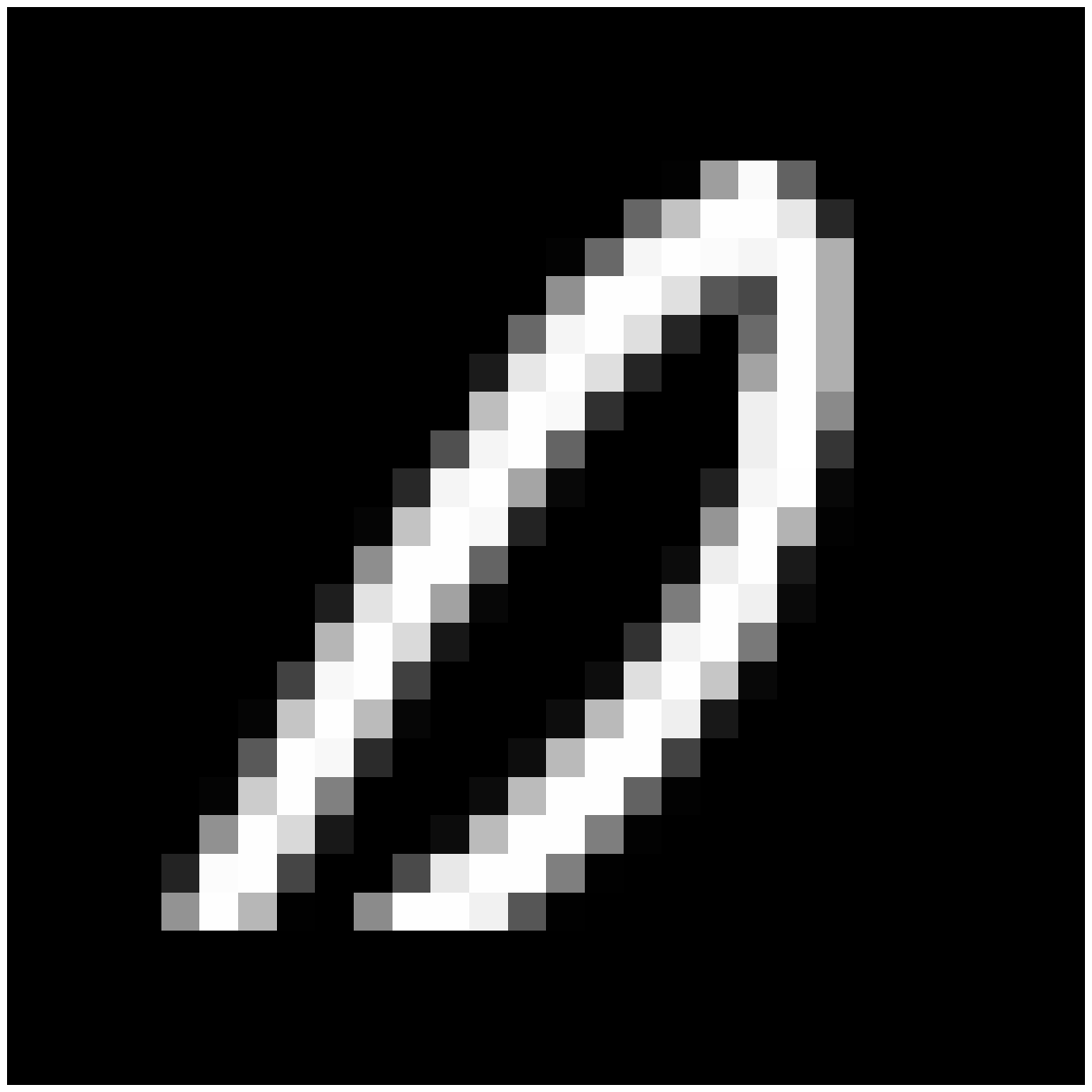}&\includegraphics[width=0.06\linewidth]{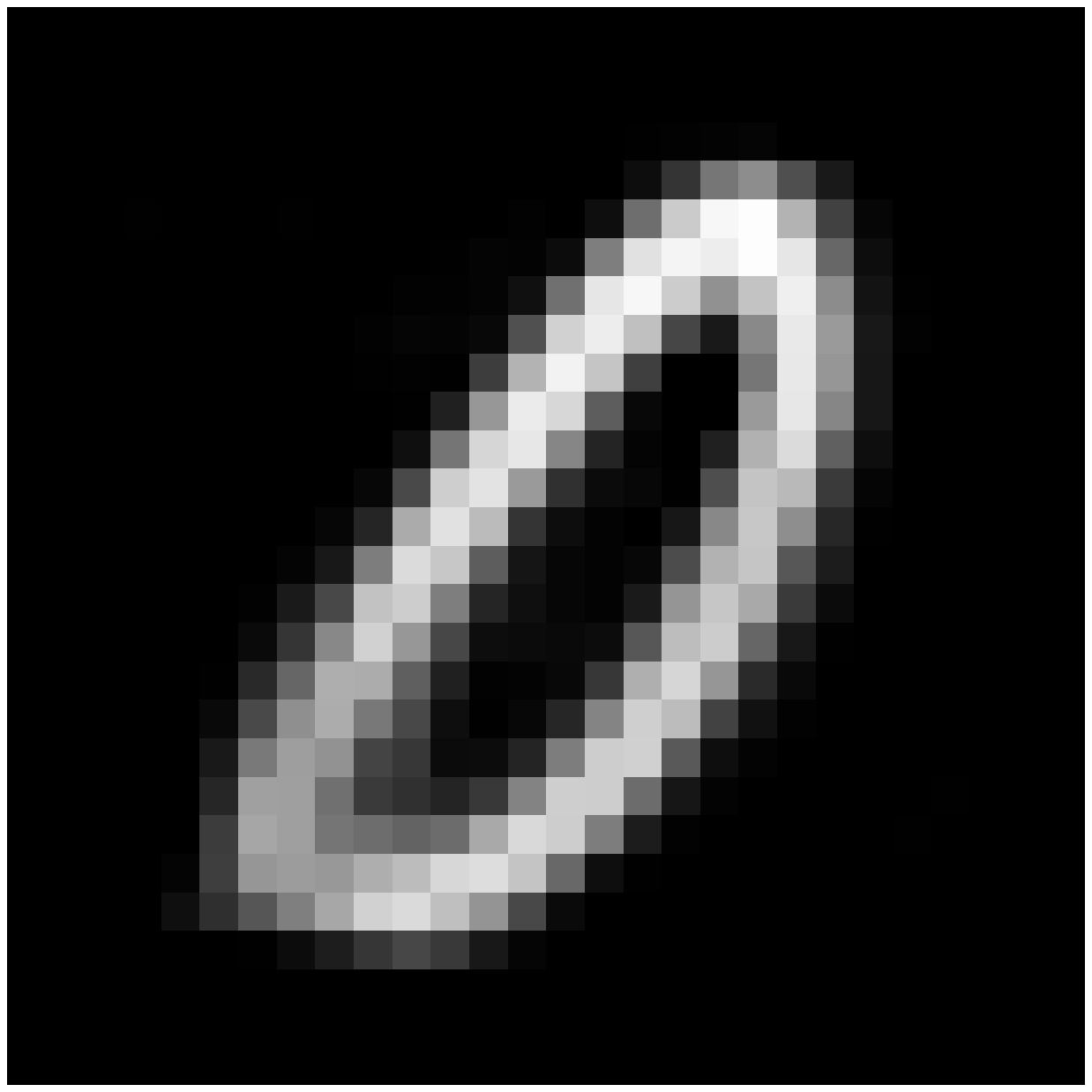}&
\includegraphics[width=0.06\linewidth]{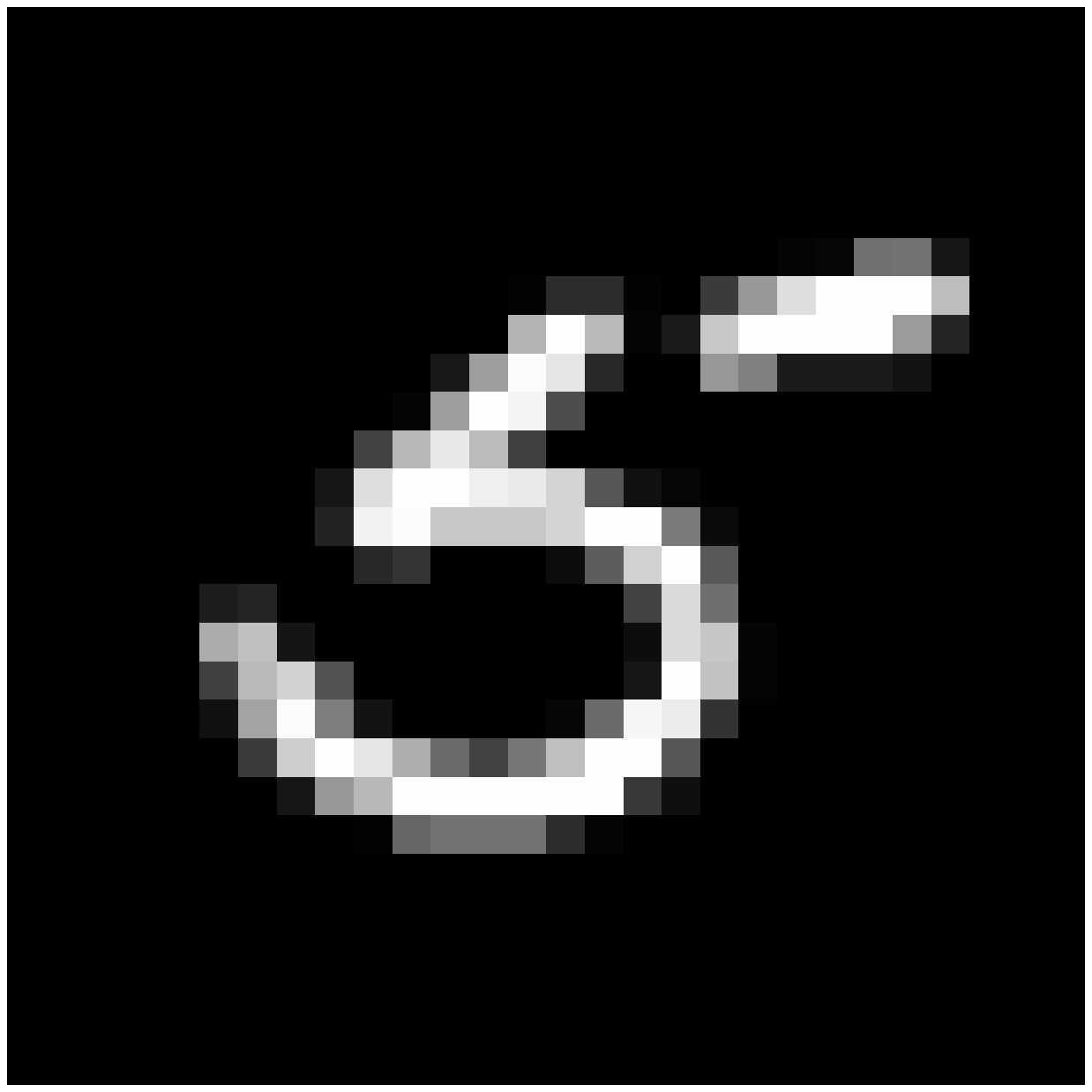}&\includegraphics[width=0.06\linewidth]{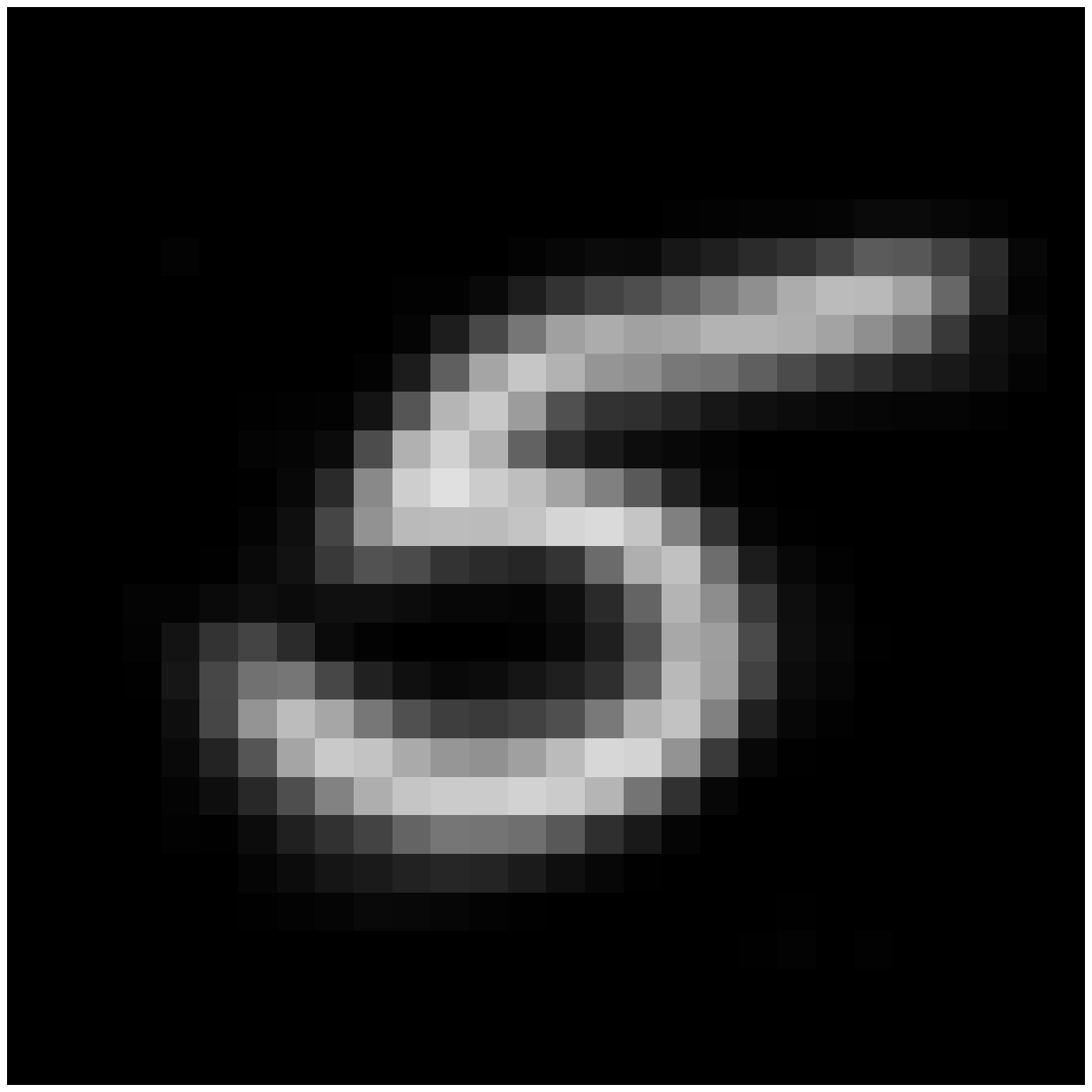}&
\includegraphics[width=0.06\linewidth]{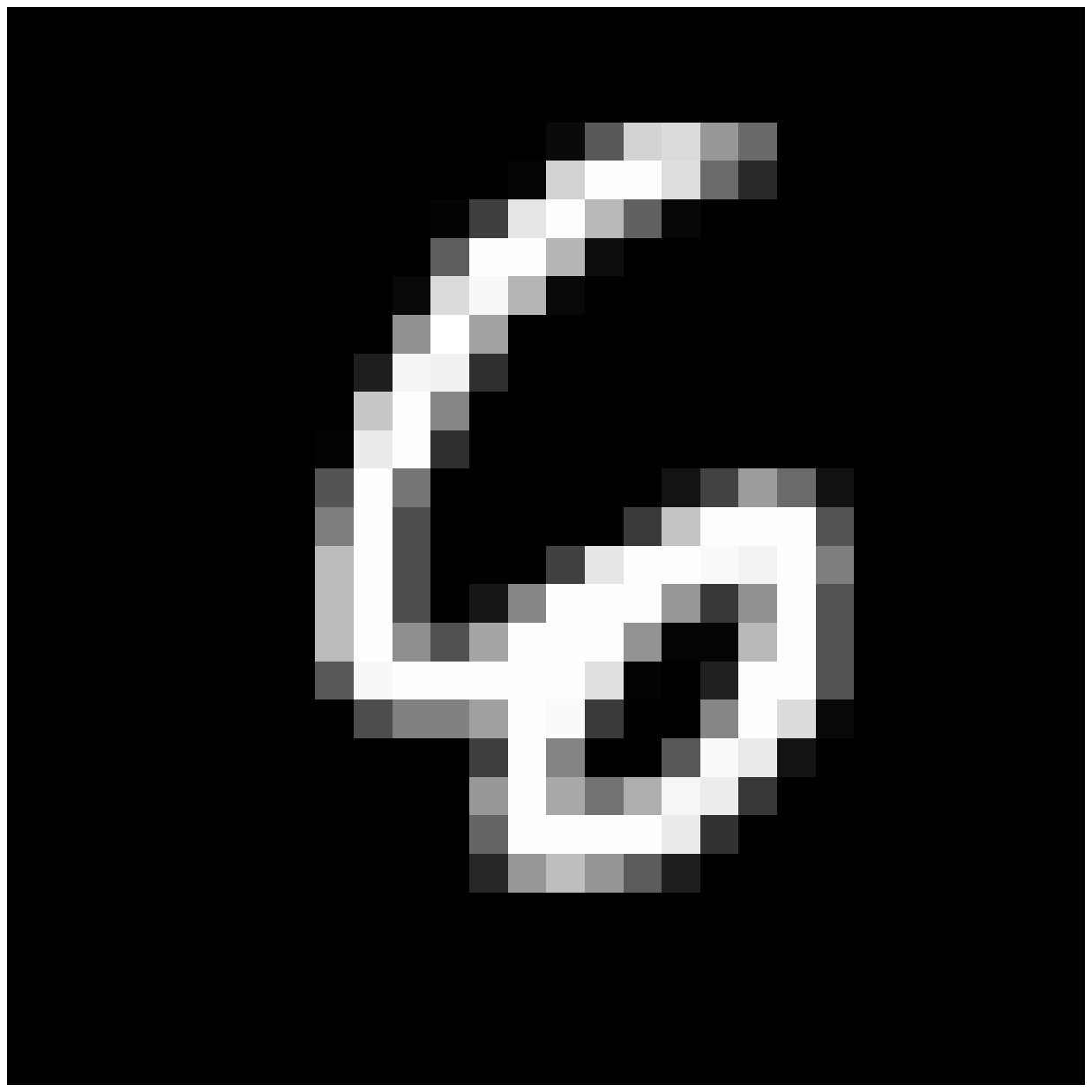}&\includegraphics[width=0.06\linewidth]{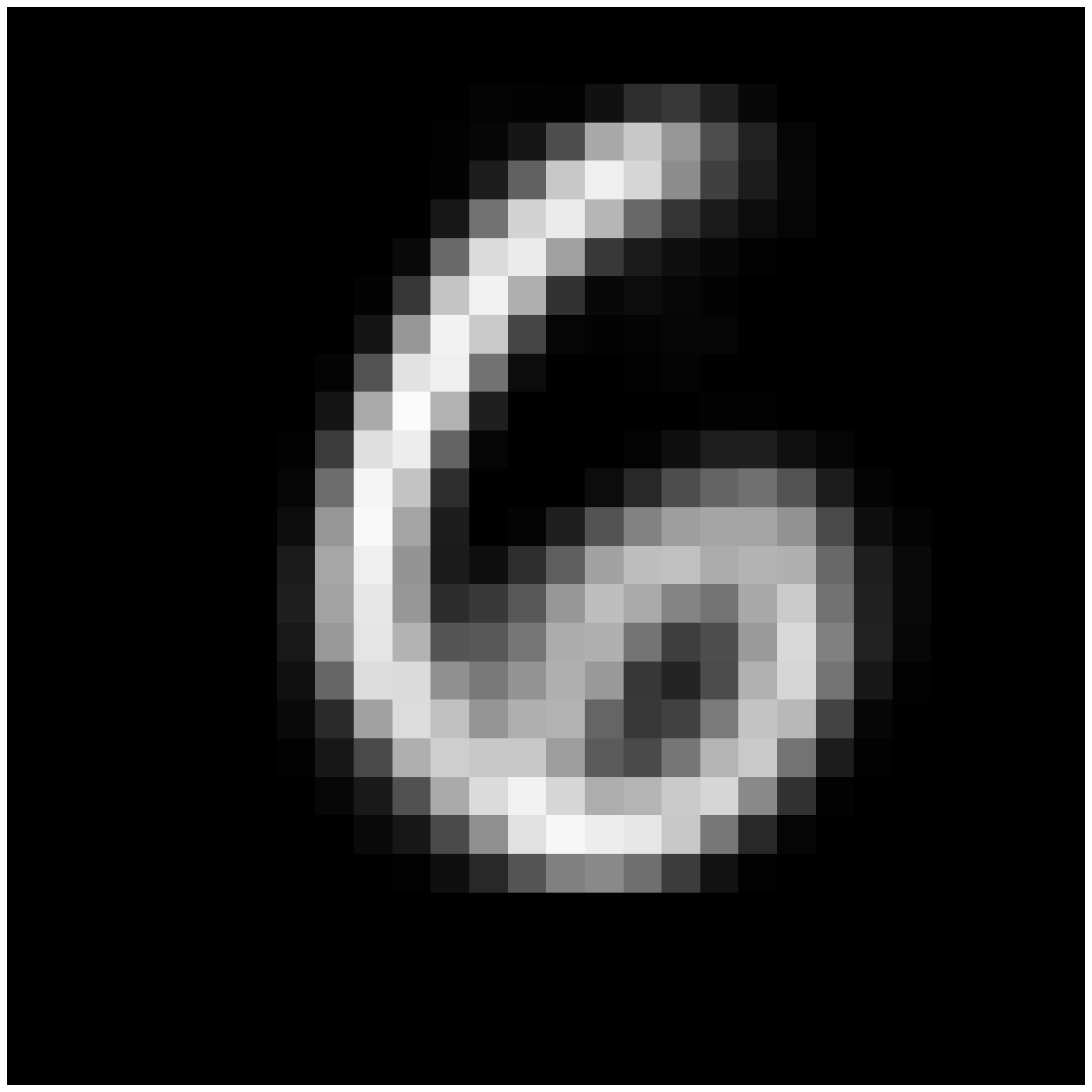}&
\includegraphics[width=0.06\linewidth]{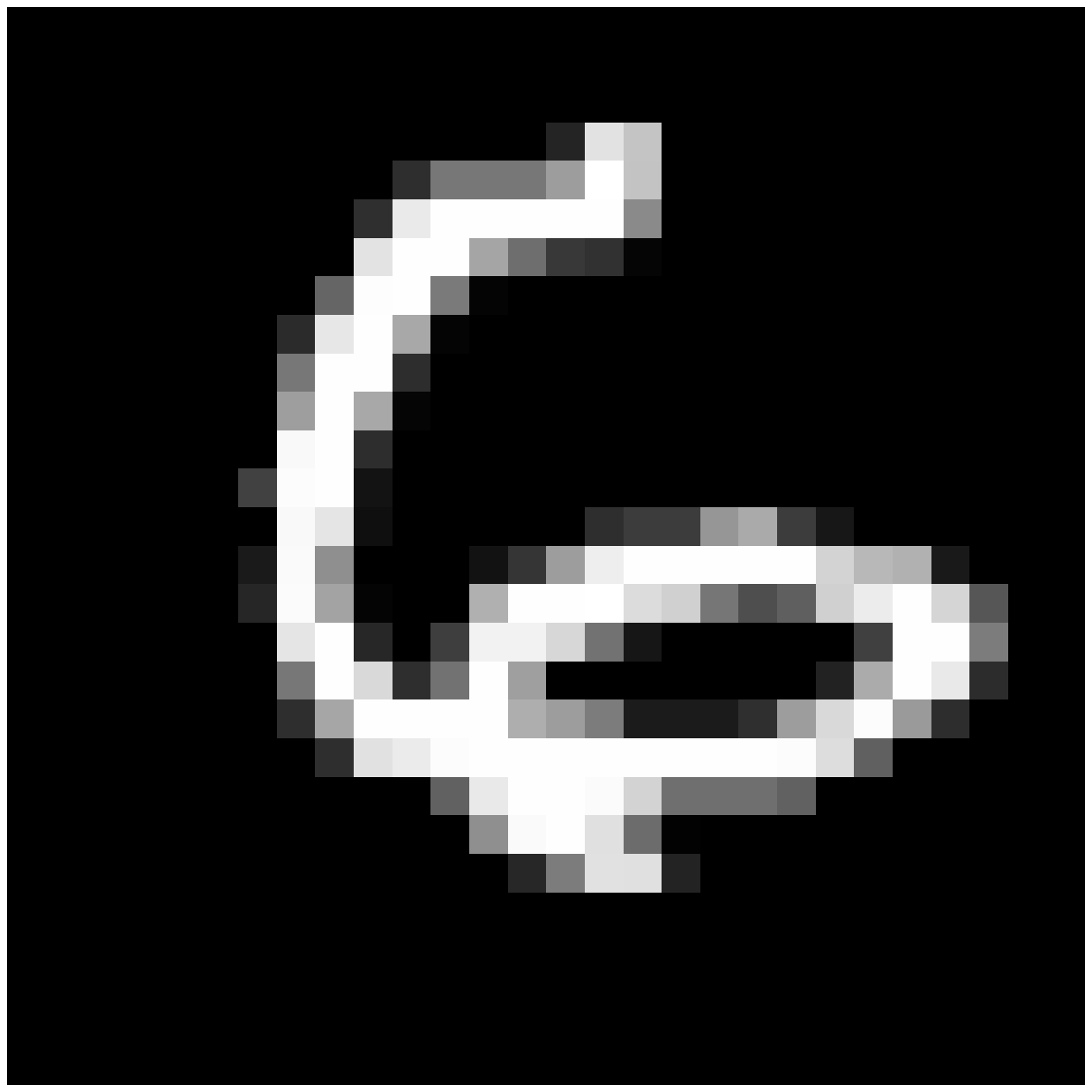}&\includegraphics[width=0.06\linewidth]{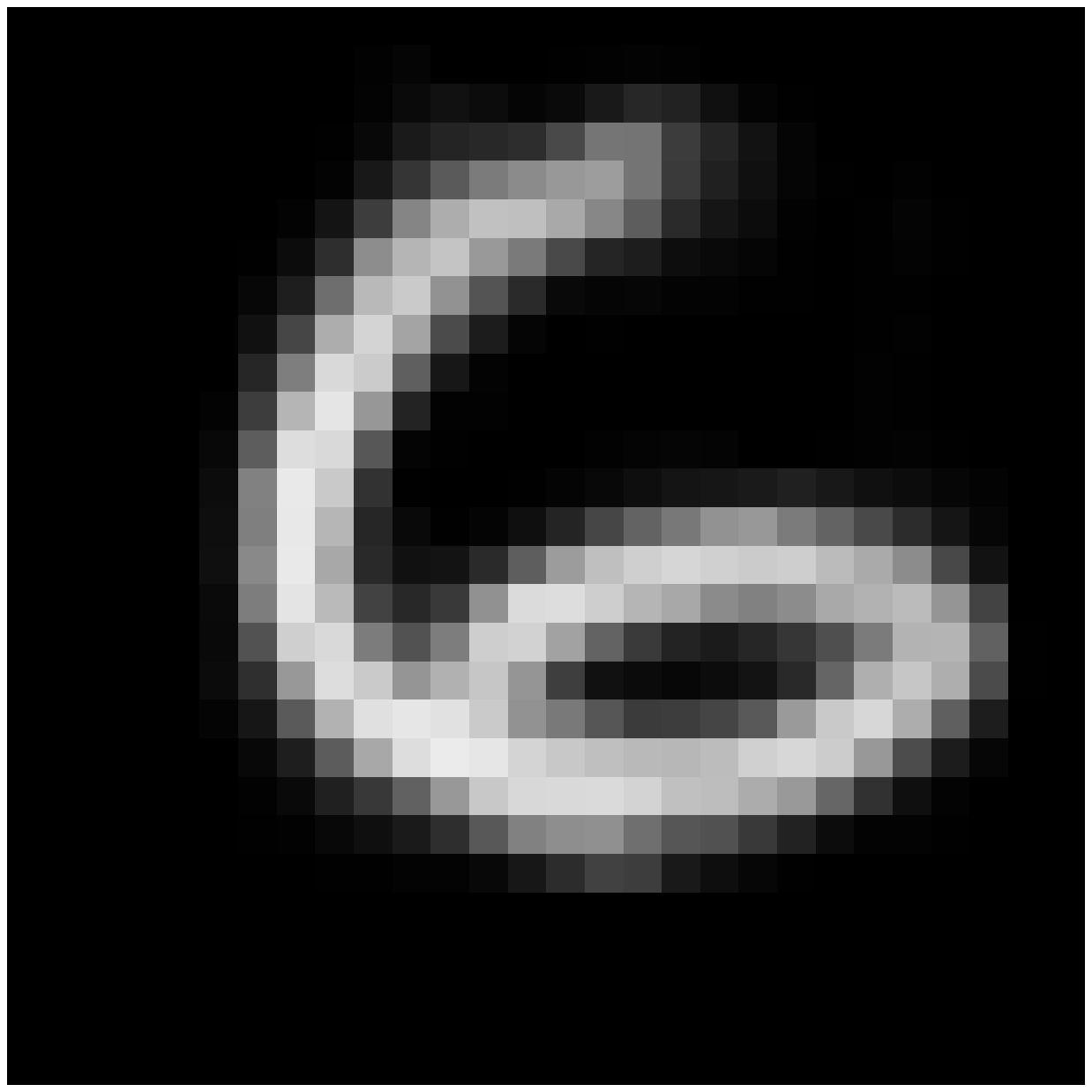}&
\includegraphics[width=0.06\linewidth]{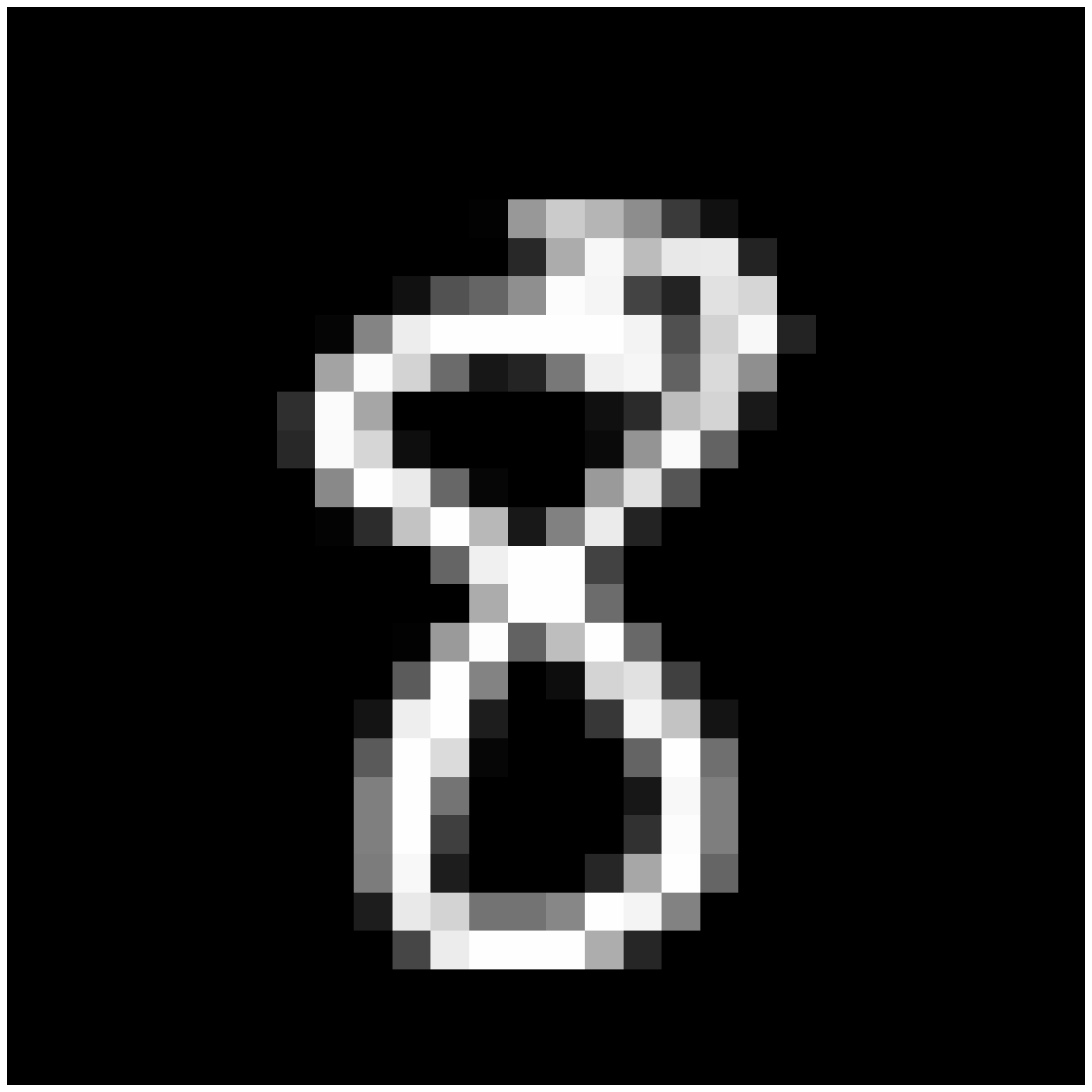}&\includegraphics[width=0.06\linewidth]{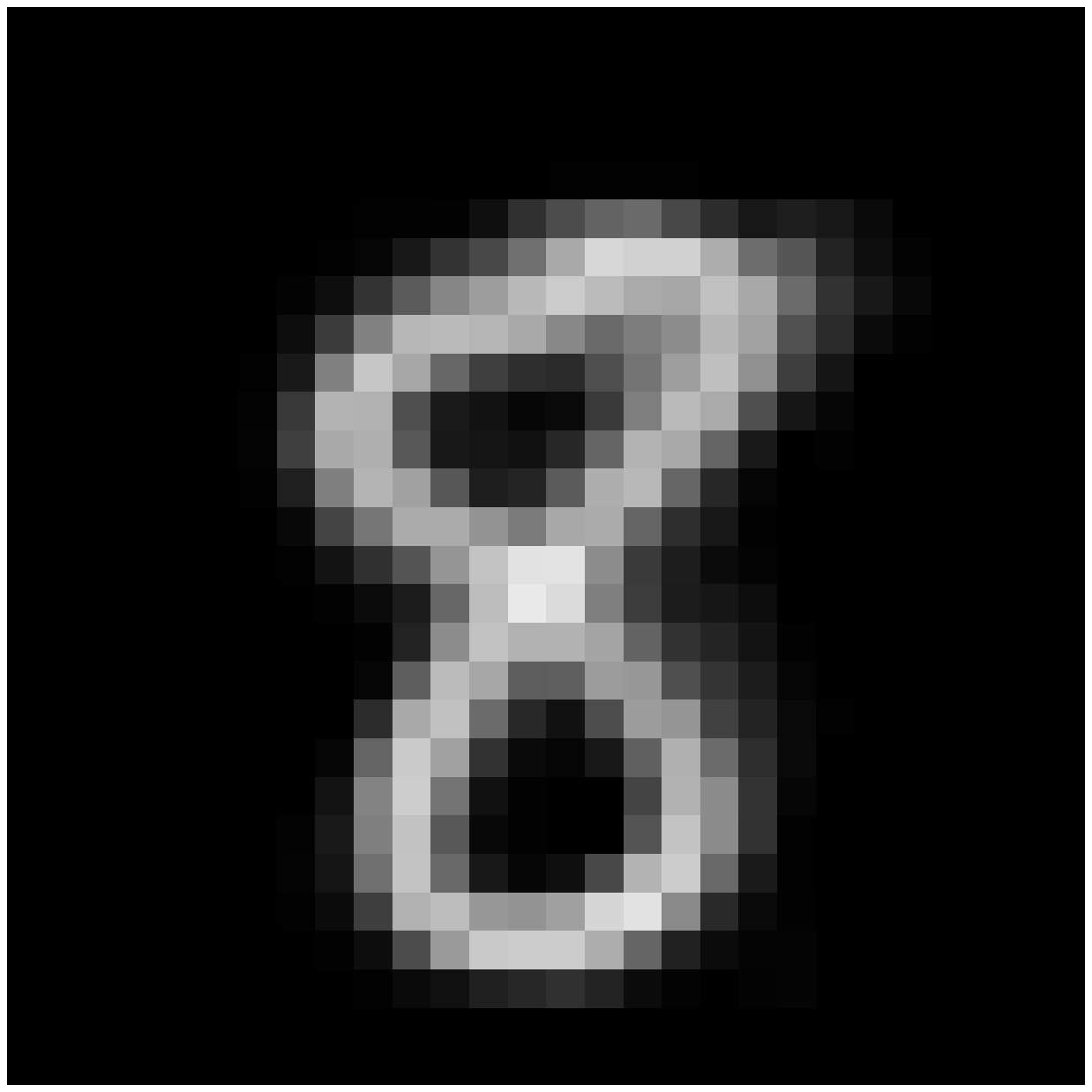}&
\includegraphics[width=0.06\linewidth]{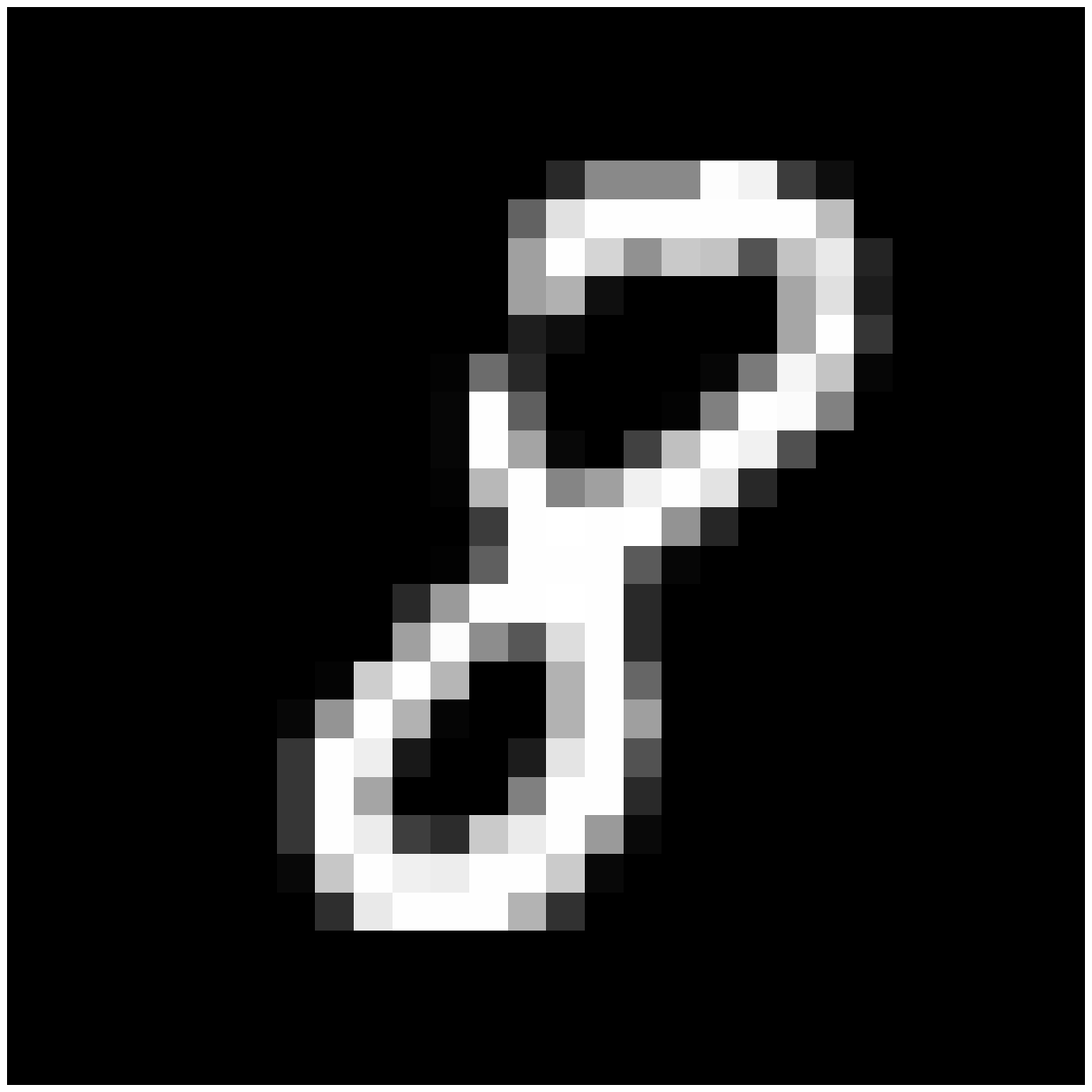}&\includegraphics[width=0.06\linewidth]{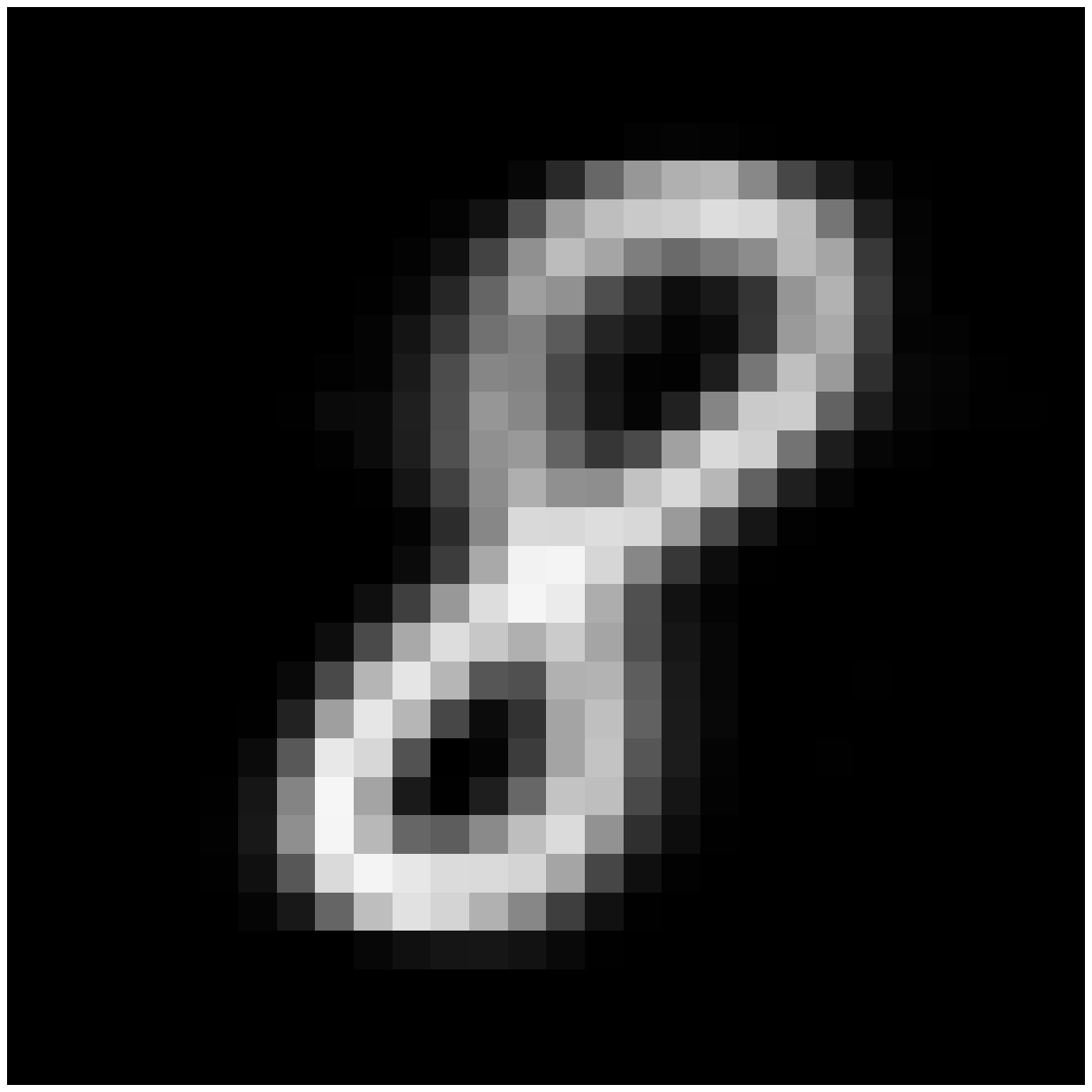}&
\includegraphics[width=0.06\linewidth]{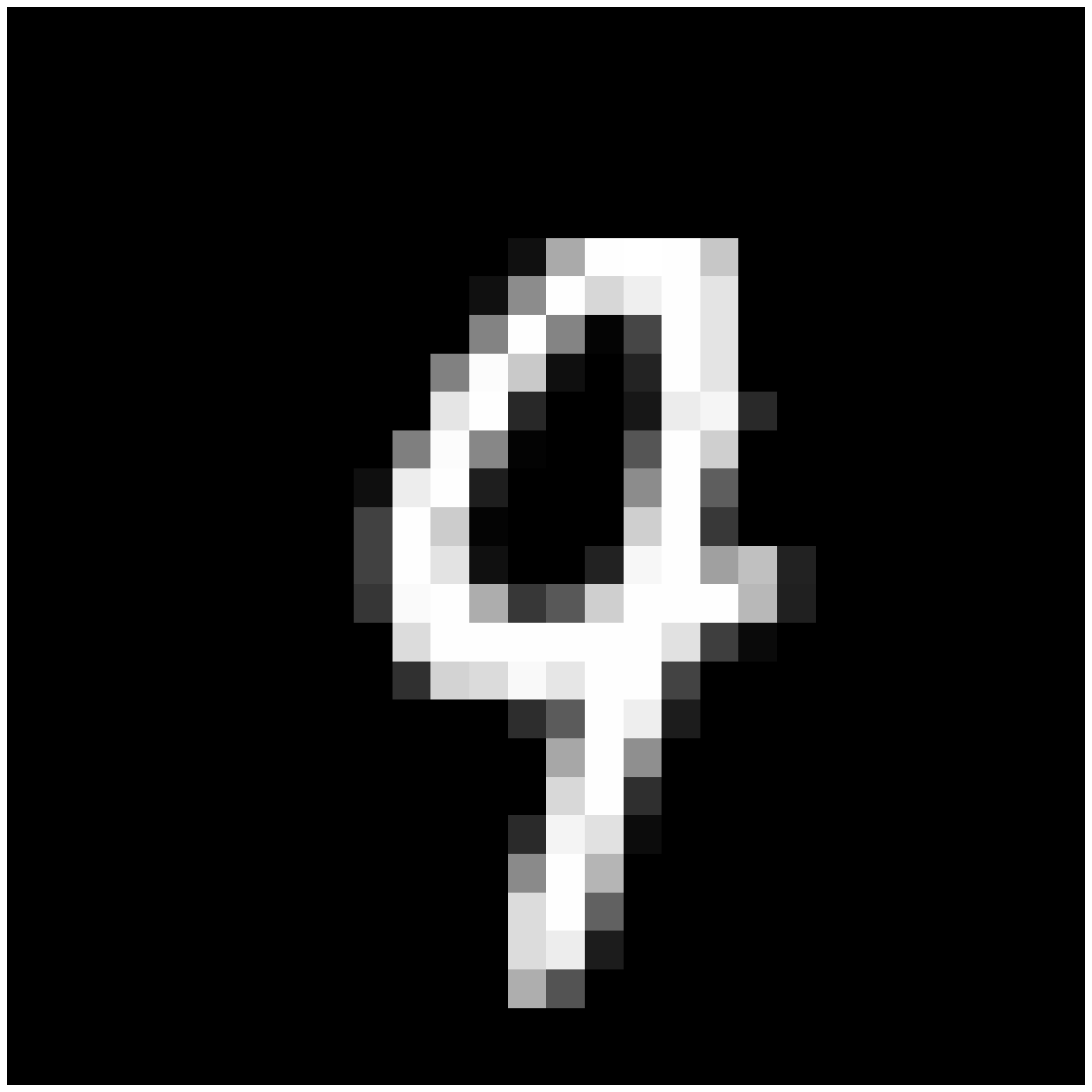}&\includegraphics[width=0.06\linewidth]{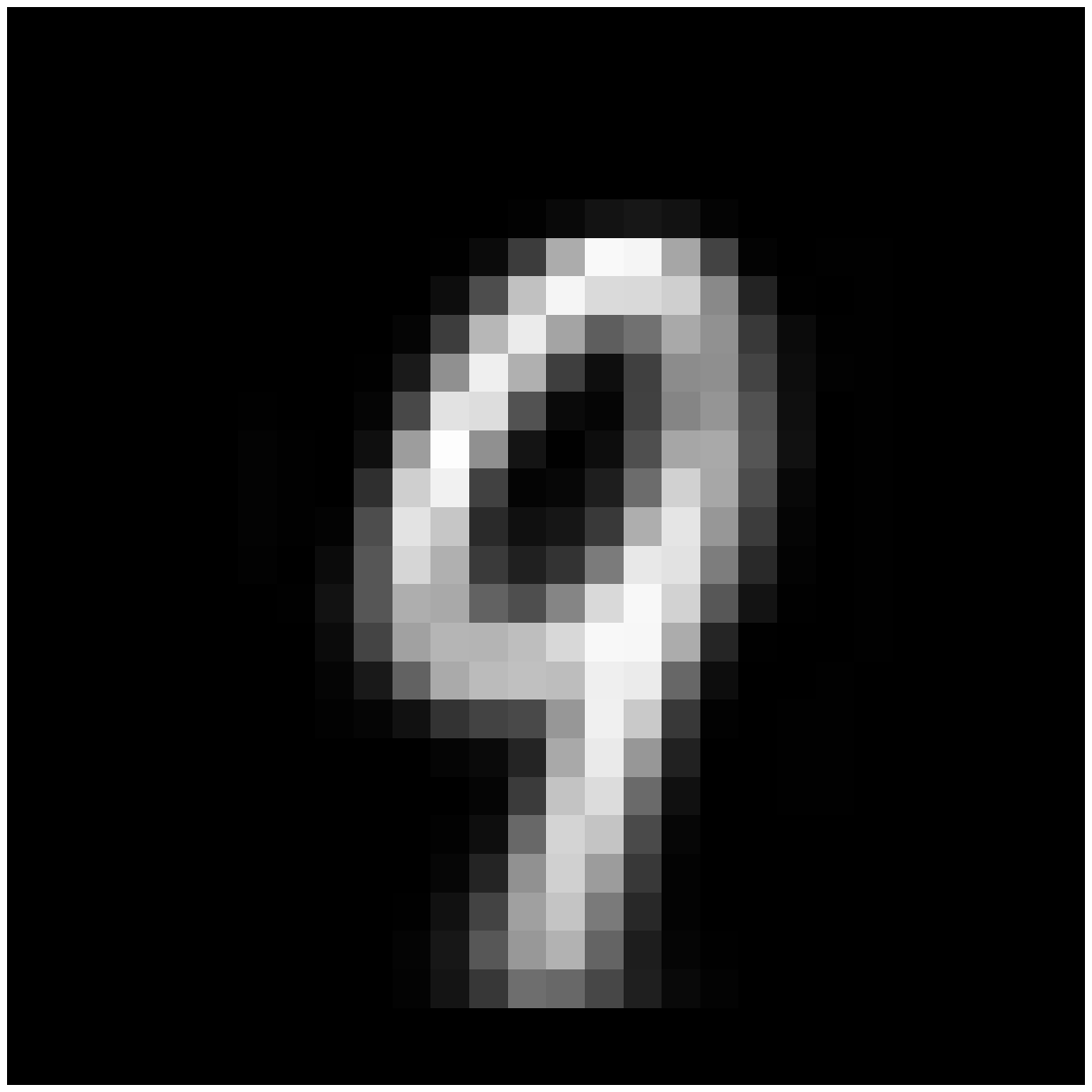}
    \end{tabular}
    \caption{Sample pairs of (original,MBMS-denoised) images from the MNIST dataset.}
    \label{f:MNIST-denoised}
  \end{center}
\end{figure*}

Fig.~\ref{f:MNIST-denoised} shows MNIST images before and after denoising with MBMS. The digits look smoother (as if they had been anti-aliased to reduce pixelation) and easier to read (compare the original \raisebox{-0.5ex}{\includegraphics[width=10em]{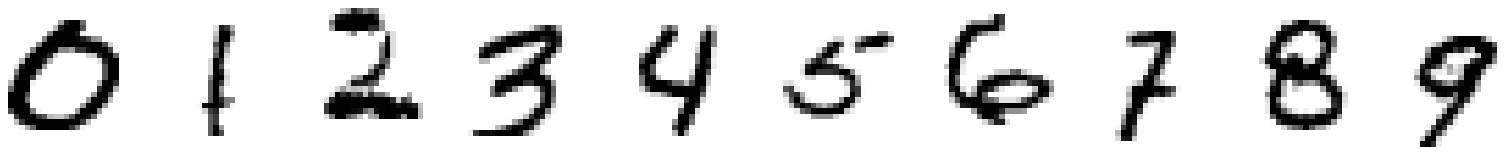}} vs the denoised \raisebox{-0.5ex}{\includegraphics[width=10em]{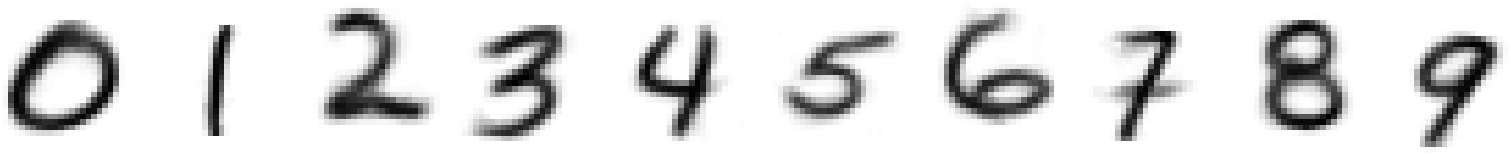}}), and indeed a classifier trained on the denoised data performs better \cite{WangCarreir10a}. While this smoothing homogenizes the digits somewhat, it preserves distinctive style aspects of each digit. MBMS performs a sophisticated denoising (very different from simple averaging or filtering) by intelligently closing loops, removing or shortening spurious strokes, enlarging holes, removing speckle noise and, in general, subtly reshaping the digits while respecting their orientation, slant and thickness. MBMS has also been applied to the identification of the structure of tectonic faults from seismic data \cite{Grillen13a}.

(Manifold) denoising can be used as a preprocessing stage for several tasks, leading to more robust models, such as dimensionality reduction \cite{WangCarreir10a}, classification \cite{HeinMaier07a,WangCarreir10a}, density estimation \cite{HallMinnot02a} or matrix completion \cite{Wang_11b}.

In some algorithms, one can establish a continuum between clustering and dimensionality reduction, with clustering being the extreme case of dimensionality reduction where the manifold dimensionality is zero. This is the case with MBMS, where a tangent space of dimension zero recovers BMS. Another case is spectral clustering and spectral dimensionality reduction, e.g.\ normalized cut \cite{ShiMalik00a} vs Laplacian eigenmaps \cite{BelkinNiyogi03b}: both operate by extracting eigenvectors from the graph Laplacian, but in spectral clustering the number of eigenvectors is the number of clusters, while in spectral dimensionality reduction one cluster is assumed and the eigenvectors correspond to degrees of freedom along the manifold defined by the cluster.

\subsection{Accelerated mean-shift algorithms}
\label{ch4.1sec:accel}

Computationally, mean-shift algorithms are slow, because their complexity is quadratic on the number of points (which would limit their applicability to small data sets), and many iterations may be required to converge (for MS). Significant work has focused on faster, approximate mean-shift algorithms, using a variety of techniques such as subsampling, discretization, search data structures \cite{Samet06a}, numerical optimization \cite{NocedalWright06a}, and others, which we review below. Some of them are specialized to image segmentation while others apply to any data. A practical implementation of a fast mean-shift algorithm could combine several of these techniques. In addition, although it does not seem to have been studied in detail, mean-shift algorithms benefit from embarrasing parallelism, since the iterations for each point proceed independently from the rest of the points.

\paragraph{Accelerating mean-shift (MS)}

With the Epanechnikov kernel, mean-shift converges in a finite number of iterations. If the neighborhood structure of the data is known a priori, as is the case in image segmentation \cite{ComanicMeer02a}, MS can be very efficient without any approximations, since the iteration for each point involves only its neighbors.

With the Gaussian kernel and with arbitrary data, for which the neighborhood structure is not known and varies over iterations, the runtime of MS can be large, particularly with large datasets, and much work has tried to accelerate Gaussian MS. Computationally, MS has two bottlenecks \cite{Carreir06a}: (1) Accurately converging to a mode can require many iterations, since its convergence is typically linear and can approach sublinearity in regions where two modes are close. (2) Each iteration is linear on the number of points $N$, because it is an average of the data. Thus, running MS for the entire dataset is $\calO(I D N^2)$, where $I$ is the average number of iterations per point and $D$ the dimension. For example, in image segmentation, typical values are 10 to 100 for $I$ (depending on the accuracy sought and the bandwidth $\sigma$), $D \le 5$ (higher if using texture features) and thousands to millions for $N$ (the number of pixels). Acceleration algorithms should attack either or both bottlenecks, while keeping a low approximation error, i.e., producing a clustering (assignment of points to modes found) that is close to that of the naive MS---otherwise one would really be running a different clustering algorithm.

As for accelerating the convergence, the obvious answer is using Newton's method (with a modified Hessian to ensure ascent), because it has quadratic convergence, computing the Hessian of a Gaussian mixture is simple \cite{Carreir00b} and Newton iterations are typically not much costlier than those of MS. The reason is that, for the low-dimensional problems for which MS is best suited, computing the Newton direction (which involves solving a $D \times D$ linear system) is not costly compared to computing the gradient or Hessian themselves. However, Newton's method does have other problems with MS: (1) it introduces user parameters such as step sizes or damping coefficients; (2) it need not be effective far from a mode, where the Hessian is undefined; (3) since Newton's method is very different from MS, it can make a point converge to a different mode than under MS. A solution to this is to use a MS-Newton method \cite{Carreir06a} that starts with MS iterations (which often suffice to move the iterate relatively close to a mode) and then switches to Newton's method.

As for reducing the cost of one iteration below $\calO(N)$, the most obvious strategy is to approximate the average in each iteration with a small subset of the data, namely the points closest to the current iterate. This can be done by simply ignoring faraway points in the average, which introduces an approximation error in the modes, or by updating faraway points infrequently as in the sparse EM algorithm of \cite{Carreir06a}, which guarantees convergence to a true mode. Unfortunately, either way this requires finding nearest neighbors, which is itself $\calO(N)$ unless we know the neighborhood structure a priori. And, if the bandwidth is relatively large (which is the case if we want a few clusters), then a significant portion of the data have non-negligible weights in the average. Another approach is to use $N$-body methods such as the (improved) fast Gauss transform \cite{GreengStrain91a,Yang_03a} or various tree structures \cite{Wang_07a}. These can reduce the $\calO(N^2)$ cost of one iteration for every data point to $\calO(N)$ or $\calO(N \log{N})$, respectively. However, this implies we run synchronous iterations (all points move at once), and does not scale to high dimensions (because the data structures involved grow exponentially with the dimension), although it is generally appropriate for the low dimensions used in image segmentation.

A final approach consists of eliminating many of the iterations for many of the points altogether, in effect predicting what mode a given point will converge to. This cannot be simply done by running MS for a small subset of the points and then assigning the remaining points to the mode of their closest point in the subset, because mean-shift clusters have complex shapes that significantly differ from a nearest-neighbor clustering, and the clustering error would be large. A very effective approach for image segmentation is the spatial discretization approximation of \cite{Carreir06a}. This is based on the fact that the trajectories followed by different points as they converge towards a given mode collect close together (see section~\ref{ch4.1sec:theory} and fig.~\ref{f:2Dcontours}\textbf{B}). Thus, we can run the naive MS for a subset of the points, keeping track of the trajectories they followed. For a new point, as soon as we detect that its iterate is close to an existing trajectory, we can stop and assign it to that trajectory's mode. With image segmentation (even if the number of features, and thus dimensions, is large), the trajectories can be coded by discretizing the image plane into subpixel cells and marking each cell the first time an iterate visits it with the mode it converges to. Since nearly all these cells end up empty, the memory required is small. This reduces the average number of iterations per point $I$ to nearly 1 with only a very small clustering error \cite{Carreir06a}. This error can be controlled through the subpixel size.

Yuan et al.\ \cite{Yuan_10a} show that, if $\y = \f(\x)$ is the result of applying a mean-shift step to a point \x, then all the points \z\ within a distance $\norm{\y-\x}$ from \y\ have a KDE value $p(\z) \ge p(\x)$. Not all these points \z\ actually converge to the same mode as \x, however many do. A fast mean-shift algorithm results from not running mean-shift iterations for all those points and assigning them to the same mode as \x. Hence, as in the discretization approach above, only a few data points actually run mean-shift iterations. The approximation error was not controlled in \cite{Yuan_10a}, although one could use instead a distance $r \norm{\y-\x}$ and use $r \in [0,1]$ to control it.

Paris and Durand \cite{ParisDurand07a} combine several techniques to accelerate mean-shift image segmentation, including spatial discretization and 1D convolutions (classifying most pixels without iterating, as above), and construct a hierarchy of clusterings based on the notion of topological persistence from computational geometry \cite{Edelsb_02a}. The algorithm is fast enough to segment large images and video, although it is not clear how good an approximation it is to the true mean-shift.

As the dimension of the data points increases, searching for nearest neighbors (to compute the mean shift update in eq.~\eqref{e:MS} or~\eqref{e:GMS} with a truncated kernel) becomes a bottleneck. One can then use approximate nearest-neighbor algorithms. For example, Locality-Sensitive Hashing (LSH) \cite{AndoniIndyk08a} is used in \cite{Georges_03a}.

\paragraph{Accelerating blurring mean-shift (BMS)}

A simple and very effective acceleration that has essentially zero approximation error was given in \cite{Carreir06b}. It essentially consists of interleaving connected-components and blurring steps. The fact that BMS collapses clusters to a single point suggests that as soon as one cluster collapses we could replace it with a single point with a weight porportional to the cluster's number of points. This will be particularly effective if clusters collapse at different speeds, which happens if they have different sizes, as predicted in section~\ref{ch4.1sec:theory}; e.g.\ see fig.~\ref{f:BMS_cameraman}. The total number of iterations remains the same as for the original BMS but each iteration uses a dataset with fewer points and is thus faster. Specifically, the Gaussian kernel density estimate is now $p(\x) = \sum^N_{n=1}{\pi_n p(\x|n)}$ where $p(\x|n) = \calN(\x;\x_n,\sigma^2\I)$ and the posterior probability is $p(n|\x_m) = p(\x_m|n) \pi_n / p(\x_m)$ as in \cite{Carreir00b}. At the beginning $\pi_n = \frac{1}{N}$ $\forall n$ and when clusters $m$ and $n$ merge then the combined weight is $\pi_m + \pi_n$. Using the matrix notation of section~\ref{ch4.1sec:MSalg} we have $w_{nm} \propto p(\x_m|n)$, $\bPi = \diag{\pi_n}$, $d_m = \smash{\sum^N_{n=1}{w_{nm} \pi_n}} = p(\x_m)$ and $(\bPi \W \D^{-1})_{nm} = p(n|\x_m)$. This can be proven to be equivalent to the original BMS.

The reduction step where coincident points are replaced with a single point can be approximated by a connected-components step where points closer than $\epsilon$ are considered coincident, where $\epsilon > 0$ takes the same value as in the final connected-components step of BMS (fig.~\ref{f:code}\textbf{B}). Thus, $\epsilon$ is the resolution of the method (below which points are indistinguishable), and while BMS applies it only after having stopped iterating, the accelerated version applies it at each iteration. Hence, the accelerated BMS algorithm alternates a connected-components graph contraction with a BMS step.

Experimentally with image segmentation, a remarkable result arises: the runtime of this accelerated BMS is almost constant over bandwidth values, unlike for both MS and BMS, corresponding to about 4 to 5 BMS iterations, and is dominated by the first few iterations, where almost no merging occurs. This means a speedup factor of 2 to 4 over BMS and 5 to 60 over MS \cite{Carreir06b}.

\section{$K$-modes and Laplacian $K$-modes algorithms}
\label{ch4.1sec:kmodes}

As discussed earlier, equating modes with clusters as in mean-shift implies we lose direct control on the number of clusters and do not differentiate between meaningful and non-meaningful modes. Recently, a \emph{$K$-modes algorithm} \cite{CarreirWang13a} has been proposed that addresses these problems%
\footnote{There exists another algorithm called ``$K$-modes'' \cite{Huang98a,Chatur_01a}, but defined for categorical data, rather than continuous data.}.
The (Gaussian) $K$-modes algorithm takes \emph{both} $K$ and $\sigma$ as user parameters and maximizes the objective function
\begin{equation}
  \label{e:kmodes-objfcn}
  \max_{\Z,\C} \quad \frac{1}{N} \sum^K_{k=1}{\sum^N_{n=1}{z_{nk} G\bigg( \norm{\frac{\x_n-\c_k}{\sigma}}^2\bigg)}} \quad \text{s.t.} \quad \Z \1_K = \1_N,\ \Z \in \{0,1\}^{NK}
\end{equation}
over the cluster centroids $\c_k \in \bbR^D$ and the assignments of points to clusters $z_{nk} \in \{0,1\}$, where $G$ is the Gaussian kernel. For a given assignment \Z, this can be seen as the sum of a KDE as in eq.~\eqref{e:kde} but separately for each cluster. Thus, a good clustering must move centroids to local modes, but also define $K$ separate KDEs. This naturally combines the idea of clustering through binary assignment variables (as in $K$-means or $K$-medoids) with the idea that, for suitable bandwidth values, high-density points are representative of a cluster (as in mean-shift). As a function of the bandwidth $\sigma$, the $K$-modes objective function becomes $K$-means for $\sigma\rightarrow\infty$ and a version of $K$-medoids for $\sigma\rightarrow 0$. The training algorithm alternates an assignment step as in $K$-means, where each data point is assigned to its closest centroid, with a mode-finding step as in mean-shift, but only for each centroid $\c_k$ in the KDE defined by its current cluster (rather than for each data point). A more robust, homotopy-based algorithm results from starting with $\sigma = \infty$ (i.e., $K$-means) and gradually decreasing $\sigma$ while optimizing the objective for each $\sigma$. The computational complexity of this algorithm is the same as for $K$-means ($\calO(DNK)$ per iteration), although it requires more iterations, and is thus much faster than mean-shift ($\calO(DN^2)$ per iteration).

$K$-modes obtains exactly $K$ modes even if the KDE of the data has more or fewer than $K$ modes, because it splits the data into $K$ KDEs. Using the homotopy-based algorithm, it tends to track a major mode in each cluster and avoid outliers. Thus, $K$-modes can work well even in high dimensions, unlike mean-shift.

The fundamental disadvantage of $K$-modes is that the clusters it defines are convex, as with $K$-means, since they are the Voronoi cells defined by the $K$ modes. This is solved in the \emph{Laplacian $K$-modes algorithm} \cite{WangCarreir14c}, which minimizes the objective function
\begin{equation}
  \label{e:Lap-kmodes-objfcn}
  \min_{\Z,\C} \quad \frac{\lambda}{2} \sum^N_{n,m=1}{ w_{nm} \norm{\z_n-\z_m}^2 } - \frac{1}{N} \sum^K_{k=1}{\sum^N_{n=1}{z_{nk} G\bigg( \norm{\frac{\x_n-\c_k}{\sigma}}^2\bigg)}} \quad \text{ s.t.} \quad \Z \1_K = \1_N,\ \Z \ge \0.
\end{equation}
This relaxes the hard assignments of $K$-modes to soft assignments (so each point may belong to different clusters in different proportions) and adds a term to the objective with a weight $\lambda \ge 0$ that encourages neighboring points to have similar soft assignments. This term can be equivalently written as $\lambda \trace{\Z^T \LL \Z}$ in terms of the graph Laplacian $\LL = \D - \W$ constructed from an affinity matrix \W, where $\D = \diag{\smash{\sum^N_{m=1}{w_{nm}}}}$ is the degree matrix. The training algorithm is as with $K$-modes, but the assignment step becomes a convex quadratic program on \Z, which can be solved efficiently with a variety of solvers.

Laplacian $K$-modes becomes the $K$-modes algorithm with $\lambda = 0$, and a \emph{Laplacian $K$-means} algorithm with $\sigma = \infty$. The introduction of the Laplacian term allows the clusters to be nonconvex, as happens with other Laplacian-based algorithms such as spectral clustering. However, here we solve a quadratic program rather than a spectral problem, and obtain a KDE, centroids (= modes) and soft assignments of points to clusters (unlike spectral clustering, which only returns hard assignments).

An out-of-sample mapping to predict soft assignments $\z(\x)$ for a test point \x\ not in the original training set can be obtained by augmenting~\eqref{e:Lap-kmodes-objfcn} with \x\ and solving but keeping the centroids and the soft assignments for the training points fixed. The result equals the projection on the simplex of the average of two terms: the average of the assignments of \x's neighbors, and a term dependent on the distances of \x\ to the centroids \cite{WangCarreir14c}. The assignment $z_k(\x)$ can be readily interpreted as a posterior probability $p(k|\x)$ of \x\ belonging to cluster $k$. Hence, Laplacian $K$-modes can be seen as incorporating \emph{nonparametric posterior probabilities} into mean-shift (or spectral clustering). The usual way of obtaining posterior probabilities in clustering is by using a mixture model, for which an EM algorithm that maximizes the likelihood is derived. In contrast to this parametric approach, in Laplacian $K$-modes the assignments optimize an objective function that is designed specifically for clustering, unlike the likelihood; the clusters are not obliged to follow a model (such as Gaussian); and, consequently, the optimization algorithm does not depend on the particular type of clusters (unlike the EM algorithm, which depends on the cluster model). These soft assignments or posterior probabilities are helpful to estimate the uncertainty in the clustering or for other uses. For example, in image segmentation they can be used as a smooth pixel mask for image matting, conditional random fields or saliency maps.

\paragraph{Cluster representatives: means, modes and medoids}

In centroid-based clustering algorithms, each cluster is associated with a centroid: the cluster mean in $K$-means, an exemplar (data point) in $K$-medoids, and the KDE mode in mean-shift, $K$-modes and Laplacian $K$-modes. A desirable property of centroids, which makes them interpretable, is that they should be valid patterns and be representative of their cluster (``look'' like a typical pattern in the cluster). A well-known disadvantage of $K$-means is that the mean of a nonconvex cluster need not be a valid pattern, since it may lie in a low-density area. This is often the case with clusters with manifold structure. For example, given a sequence of rotated digit-1 images \raisebox{-0.5ex}{\mbox{\includegraphics[width=1em,bb=275 367 327 419,clip]{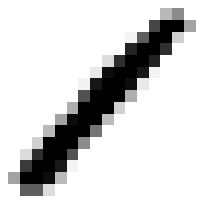}\includegraphics[width=1em,bb=275 367 327 419,clip]{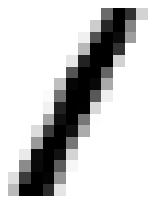}\includegraphics[width=1em,bb=275 367 327 419,clip]{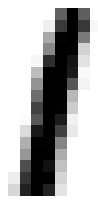}\includegraphics[width=1em,bb=275 367 327 419,clip]{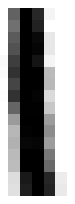}\includegraphics[width=1em,bb=275 367 327 419,clip]{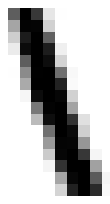}\includegraphics[width=1em,bb=275 367 327 419,clip]{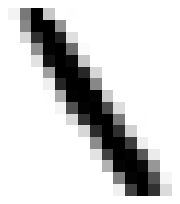}\includegraphics[width=1em,bb=275 367 327 419,clip]{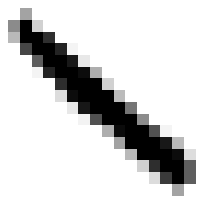}}}, its mean \raisebox{-0.5ex}{\includegraphics[width=1em,bb=275 367 327 419,clip]{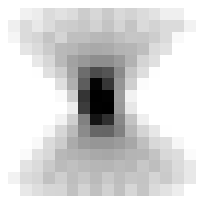}} is not a valid digit-1 image itself. In $K$-medoids, exemplars (data points) are valid patterns by definition, although when the data is noisy, individual exemplars may look somewhat atypical (see the manifold denoising discussion in section~\ref{ch4.1sec:denoising}). Modes can strike an optimal tradeoff, in being typical, valid and yet denoised representatives, as follows. First, modes are by definition on high-density areas, so in this sense they are representative of the KDE and cluster. Second, a mode can be characterized as a scale-dependent, local average of the data. Indeed, a mode $\x^*$ is a maximum of the density $p$, so its gradient at $\x^*$ is zero, and this implies that $\x^*$ equals the weighted average of the data in the sense of eq.~\eqref{e:MS} and~\eqref{e:GMS} (where $\x^* = \f(\x^*)$), as seen in section~\ref{ch4.1sec:existing}. As a function of the scale, a mode spans a continuum between equaling the regular mean of the whole data ($\sigma = \infty$) and equaling any individual data point ($\sigma = 0$). With an intermediate bandwidth in the KDE, they are local averages of the data and so can remove noise that affects each individual exemplar, thus being even more prototypical than actual data points. In this sense, (Laplacian) $K$-modes achieves a form of intelligent denoising similar to that of MBMS (section~\ref{ch4.1sec:denoising}).

Table~\ref{t:clustercomp} compares several clustering algorithms in terms of whether their centroids are valid patterns, whether they can model nonconvex clusters, whether they estimate a density, and whether the cluster assignments are hard or soft.

\begin{table}[t]
  \centering
  \caption{Comparison of properties of different clustering algorithms.}
  \label{t:clustercomp}
  \small
  \begin{tabular}{@{}l|ccccccc@{}}
    \hline
    & $K$-means & $K$-medoids & Mean-shift & \caja{c}{c}{Spectral\\ clustering} &  $K$-modes & \caja{c}{c}{Laplacian\\ $K$-modes} & \caja{c}{c}{Gaussian \\ mixture, EM} \\
    \hline
    Centroids     & likely invalid   & ``valid'' & ``valid'' & N/A & valid & valid & likely invalid \\
    \caja{c}{c}{Nonconvex \\ clusters}  & no        & depends & yes & yes & no & yes & to some extent \\
    Density       & no        & no & yes & no & yes & yes & yes \\
    Assignment    & hard      & hard & hard & hard & hard & soft, nonparam. & soft, param. \\
    \hline
  \end{tabular}
\end{table}

\section{Examples, applications, software}
\label{ch4.1sec:apps}

We illustrate MS and BMS in the clustering application where they have been most effective (image segmentation), and point out clustering applications where they are not so effective (high-dimensional data or data with manifold structure), as well as applications beyond clustering where MS and BMS have been used. Matlab code for most of the algorithms described may be obtained from the author.

\paragraph{Image segmentation}

\begin{figure}[t!]
  \begin{center}
    \begin{tabular}[c]{@{}c@{\hspace{0.01\textwidth}}c@{\hspace{0.01\textwidth}}c@{\hspace{0.01\textwidth}}c@{}}
      \textbf{A}: original image & \textbf{B}: MS segmentation & \textbf{C}: datapoints \& iterates & \textbf{D}: data/iter.\ (2D proj.) \\
      \includegraphics[width=0.24\textwidth]{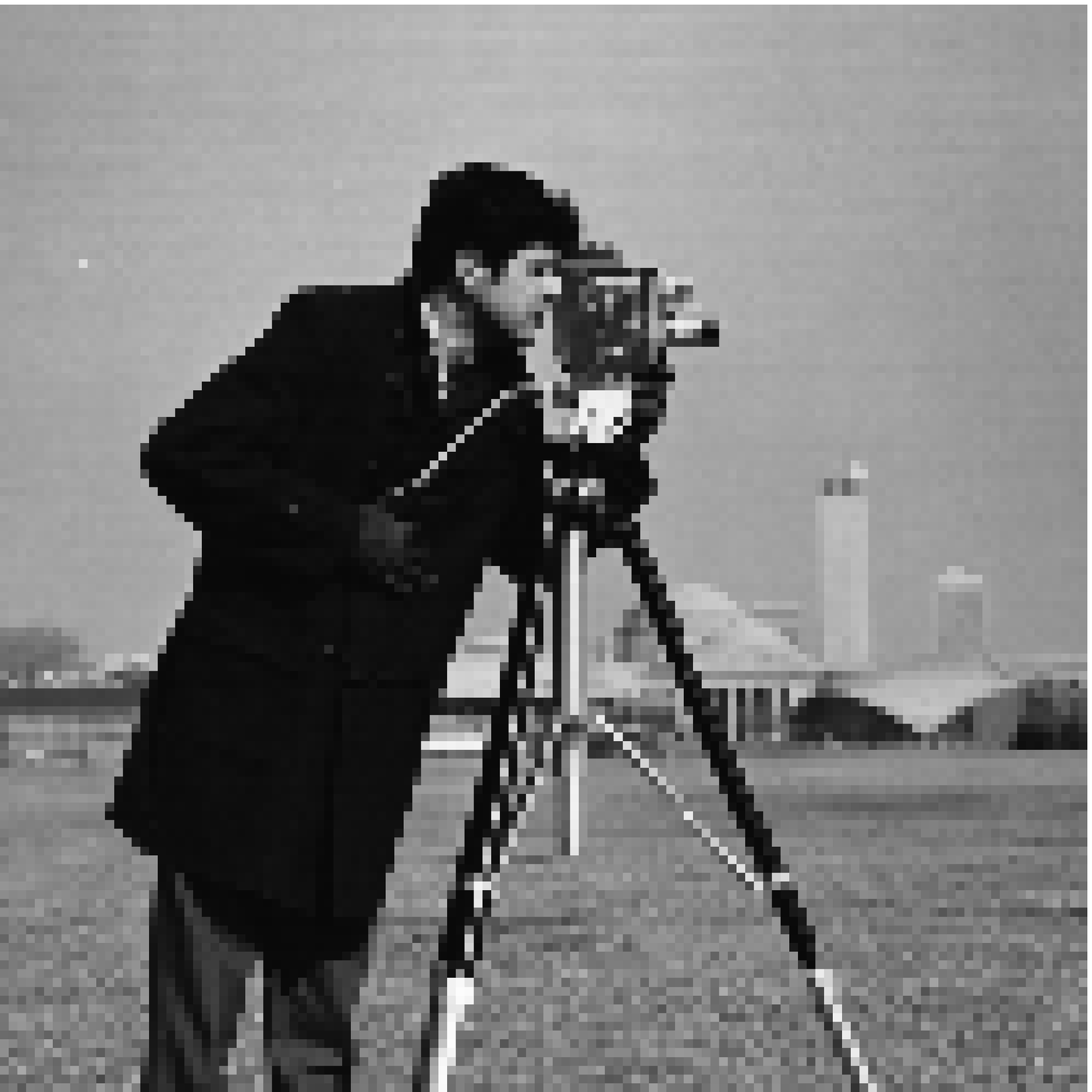} &
      \includegraphics[width=0.24\textwidth,bb=18 18 131 131,clip]{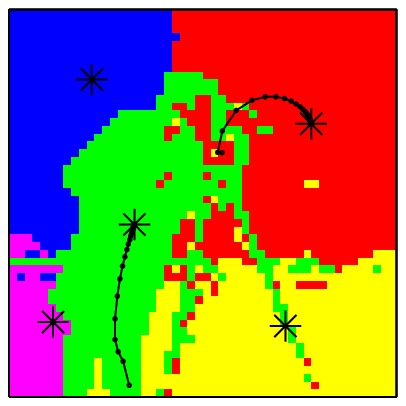} &
      \psfrag{I}[][][1][90]{$I$}
      \psfrag{H}[][]{$i$}
      \psfrag{V}[][b]{$j$}
      \includegraphics[width=0.24\textwidth]{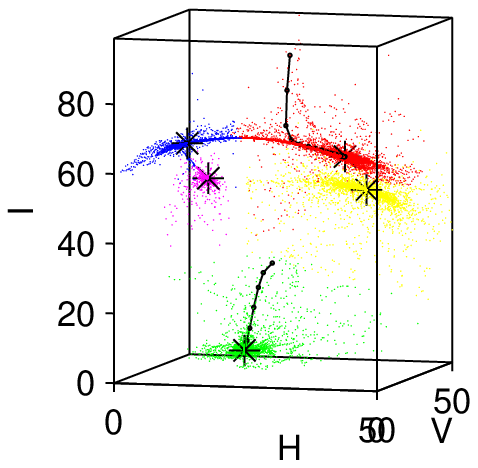} &
      \includegraphics[width=0.24\textwidth,bb=18 18 131 131,clip]{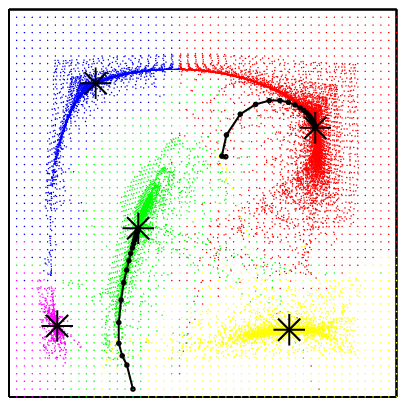}
    \end{tabular}
    \caption{\textbf{A}: test image (\texttt{cameraman} $50 \times 50$ grayscale). \textbf{B}: Gaussian MS segmentation with bandwidth $\sigma = 8$ pixels, resulting in 5 clusters (shown in different colors), with the respective modes marked $\ast$. \textbf{C}: the dataset (pixels) and all the MS iterates for all starting points in $(i,j,I)$ space, colored by cluster. \textbf{D}: projection of the middle plot on $(i,j)$ space. The paths for two starting pixels are shown in plots \textbf{B}--\textbf{D}.}
    \label{f:MS_cameraman}
  \end{center}
\end{figure}

MS and BMS are most effective with low-dimensional data, and its most successful application is in image segmentation \cite{ComanicMeer02a}. They are applied as follows. Given an image, we consider each pixel as a data point in a space consisting of two types of features: \emph{spatial features} (the $i$ and $j$ location of a pixel) and \emph{range features} (the intensity $I$ or grayscale value, the color values, texture features, etc.\ of the pixel). The range features are scaled to span approximately the range of the spatial features. This way, all features and the bandwidth have pixel units. For example, for the image of fig.~\ref{f:MS_cameraman}, we rescale the original intensity values to the range $[0,100]$, so a feature vector $(4,13,80)$ would correspond to the pixel located at coordinates $(4,13)$, which has an intensity equal to $80$\% of the maximum intensity (white). The precise scaling will affect the clustering and should be done carefully. Using spatial features is beneficial because they introduce spatial coherence (nearby pixels tend to belong to the same cluster), although sometimes only the range features are used. One should use a perceptually uniform color space such as LAB rather than the RGB space, so that Euclidean distances approximately match perceptual differences in color \cite{ForsytPonce03a}.

\begin{figure*}[t!]
  \begin{center}
    \begin{tabular}[c]{@{}c@{\hspace{0.005\textwidth}}c@{\hspace{0.005\textwidth}}c@{\hspace{0.005\textwidth}}c@{\hspace{0.005\textwidth}}c@{\hspace{0.005\textwidth}}c@{}}
 $\tau = 0$ & $\tau = 1$ & $\tau = 2$ & $\tau = 3$ & $\tau = 4$ & $\tau = 5$ \\
      \includegraphics[width=0.16\textwidth,bb=138 226 494 581,clip]{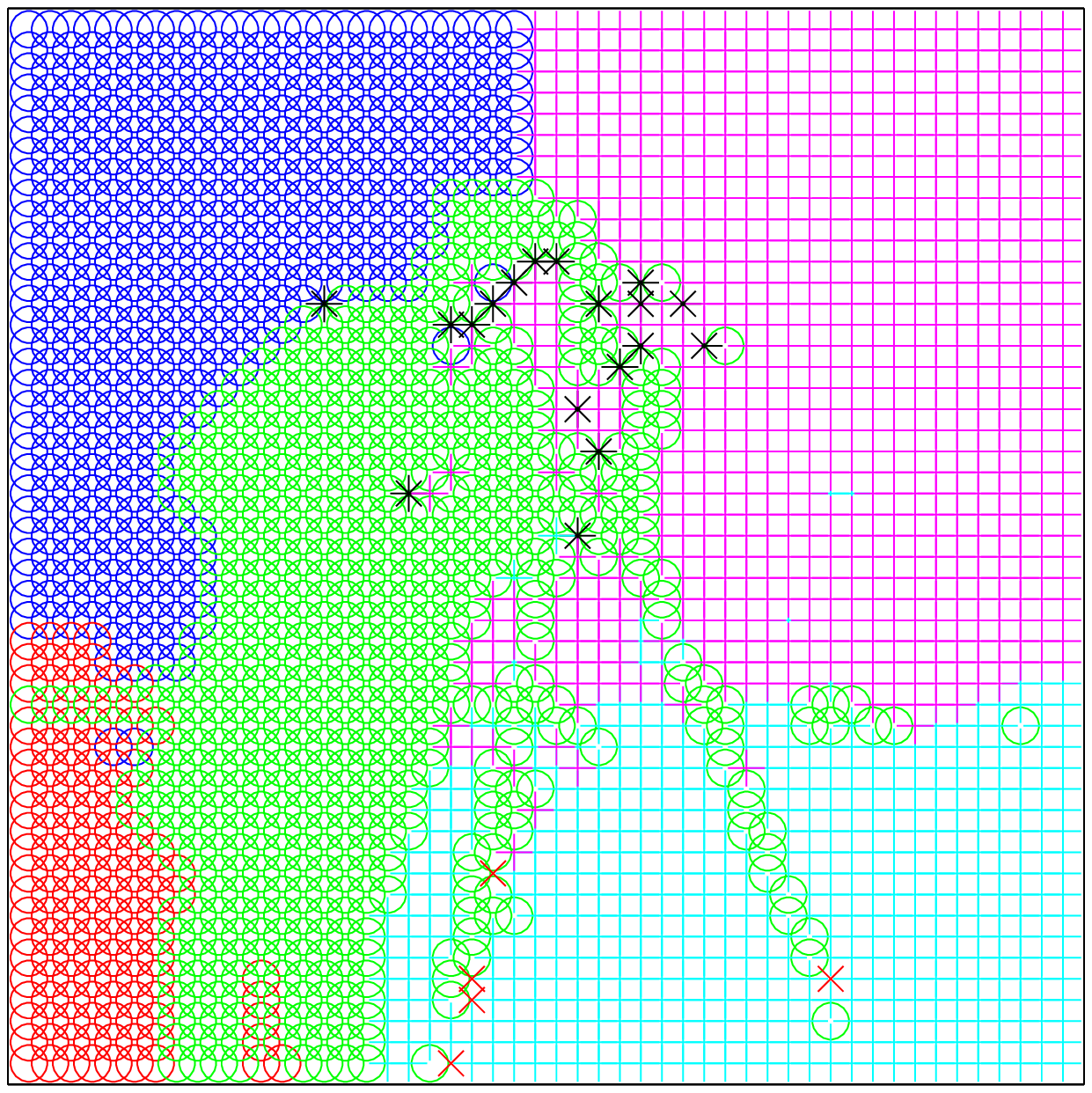} &
      \includegraphics[width=0.16\textwidth,bb=138 226 494 581,clip]{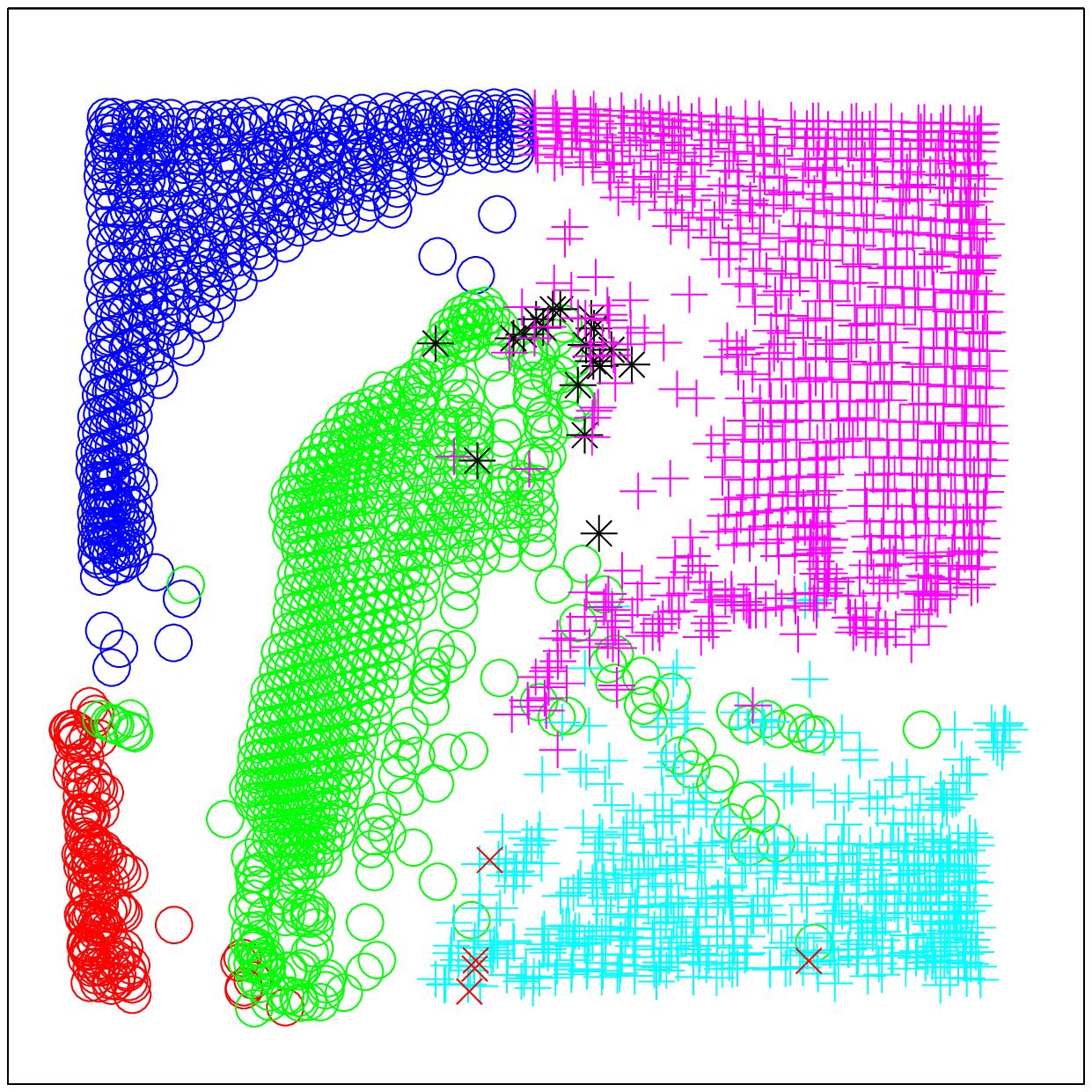} &
      \includegraphics[width=0.16\textwidth,bb=138 226 494 581,clip]{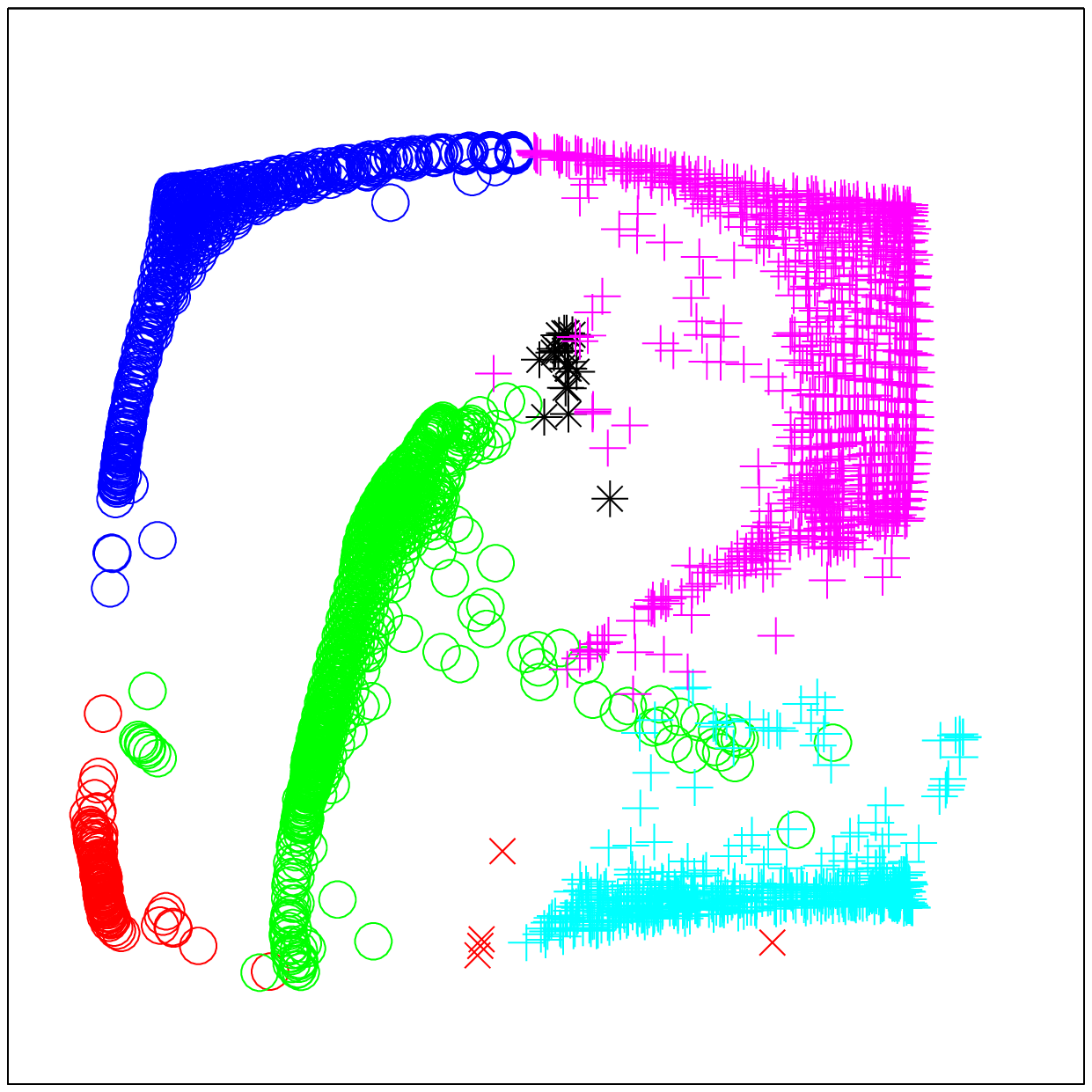} &
      \includegraphics[width=0.16\textwidth,bb=138 226 494 581,clip]{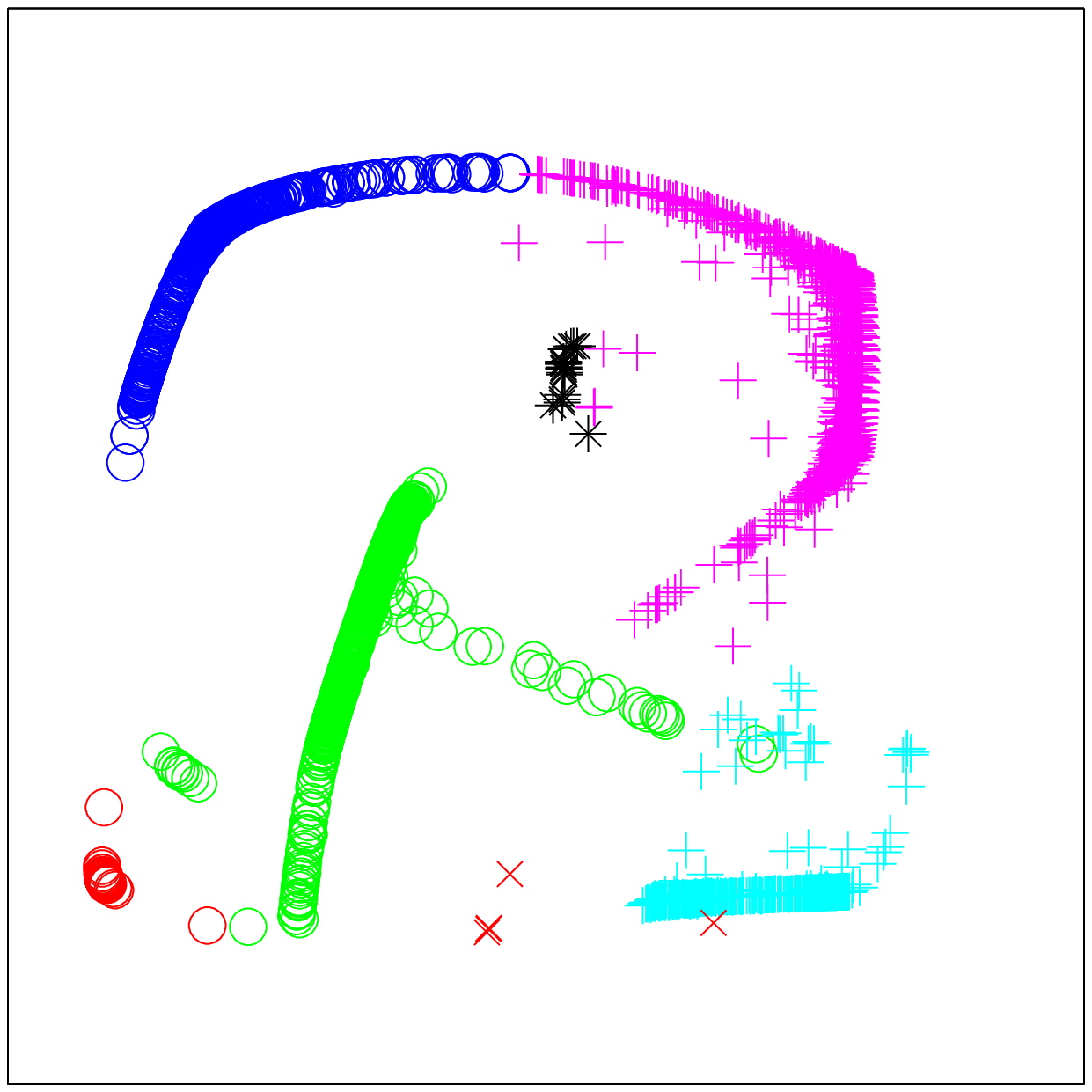} &
      \includegraphics[width=0.16\textwidth,bb=138 226 494 581,clip]{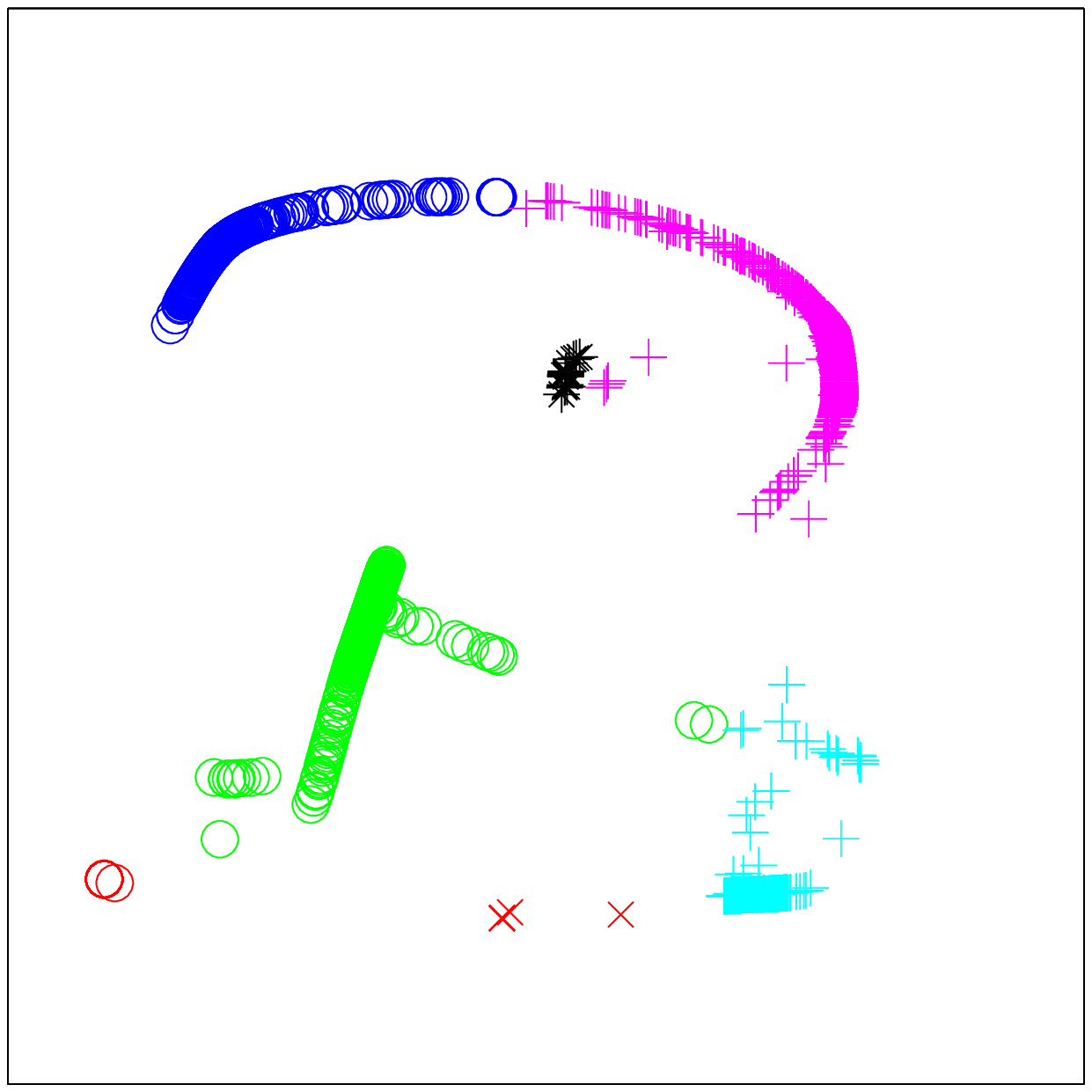} &
      \includegraphics[width=0.16\textwidth,bb=138 226 494 581,clip]{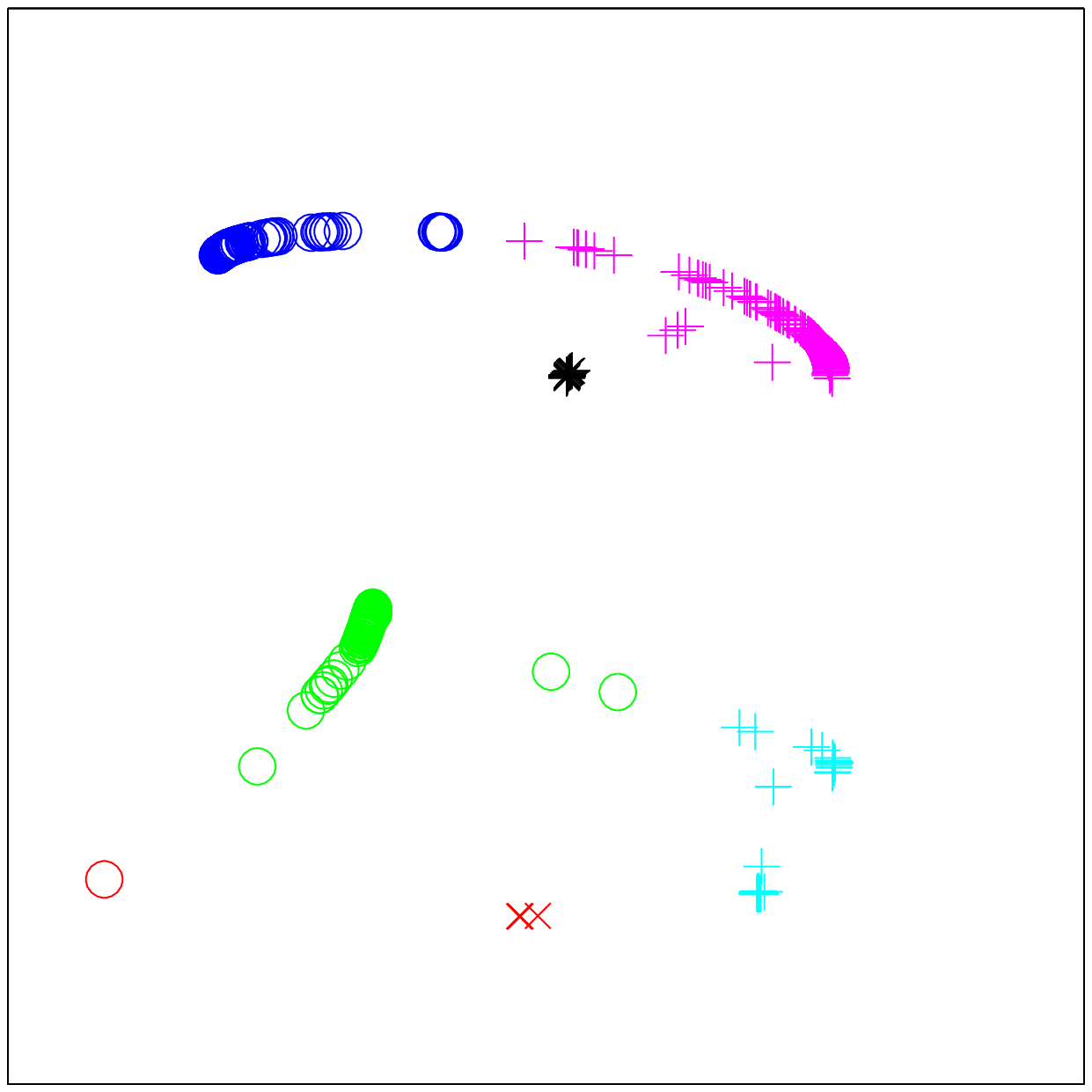} \\
      $\tau = 6$ & $\tau = 7$ & $\tau = 8$ & $\tau = 9$ & $\tau = 10$ & $\tau = 11$ \\
      \includegraphics[width=0.16\textwidth,bb=138 226 494 581,clip]{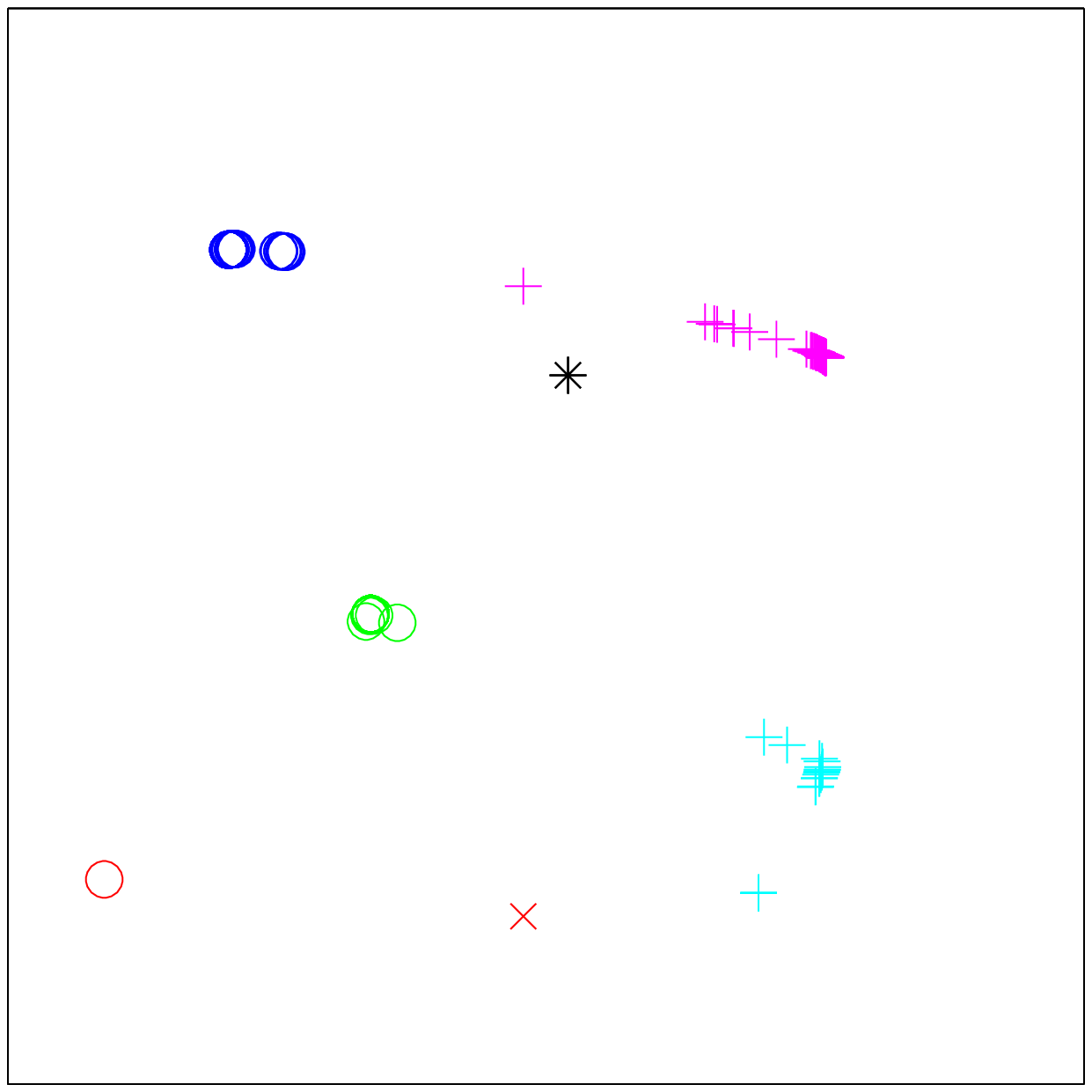} &
      \includegraphics[width=0.16\textwidth,bb=138 226 494 581,clip]{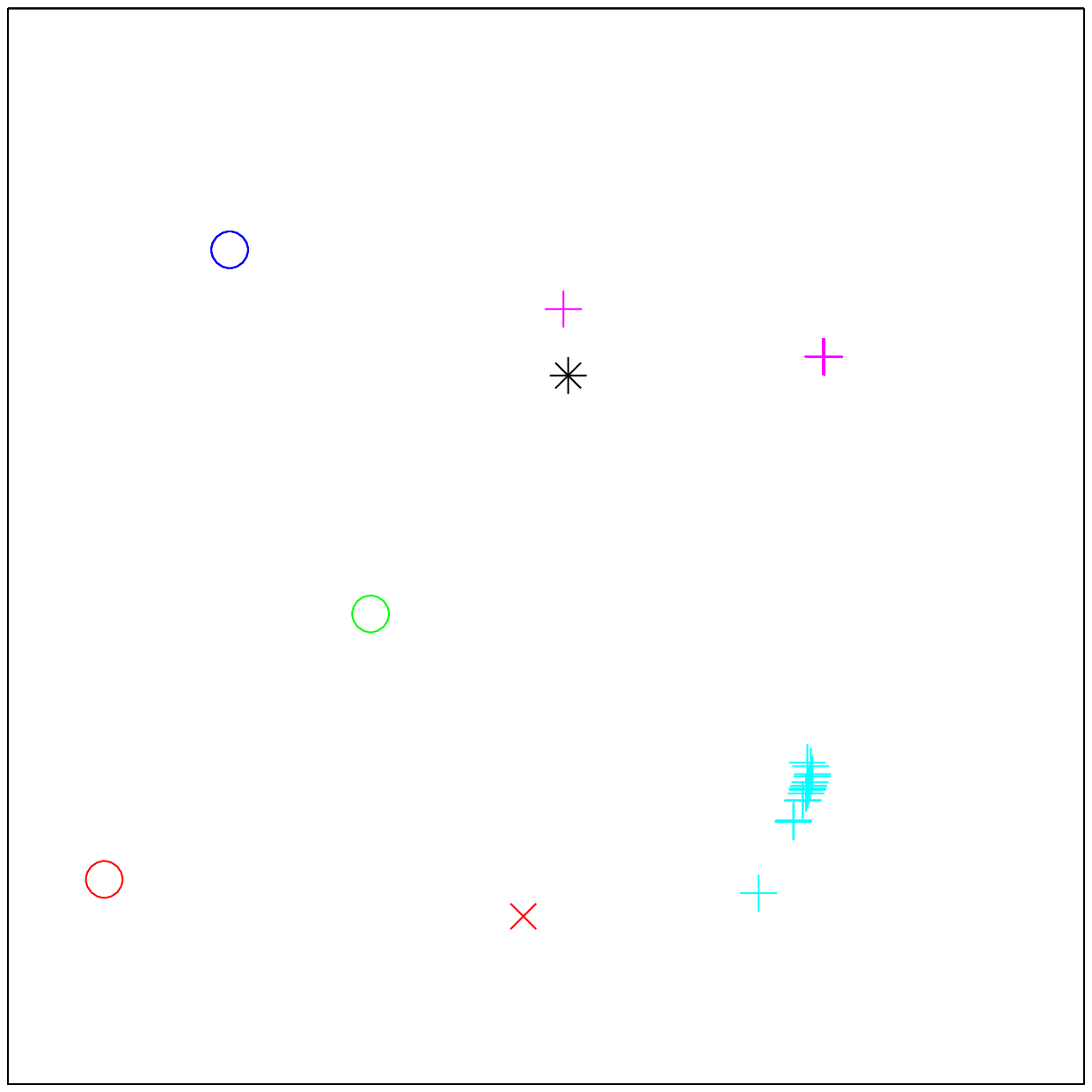} &
      \includegraphics[width=0.16\textwidth,bb=138 226 494 581,clip]{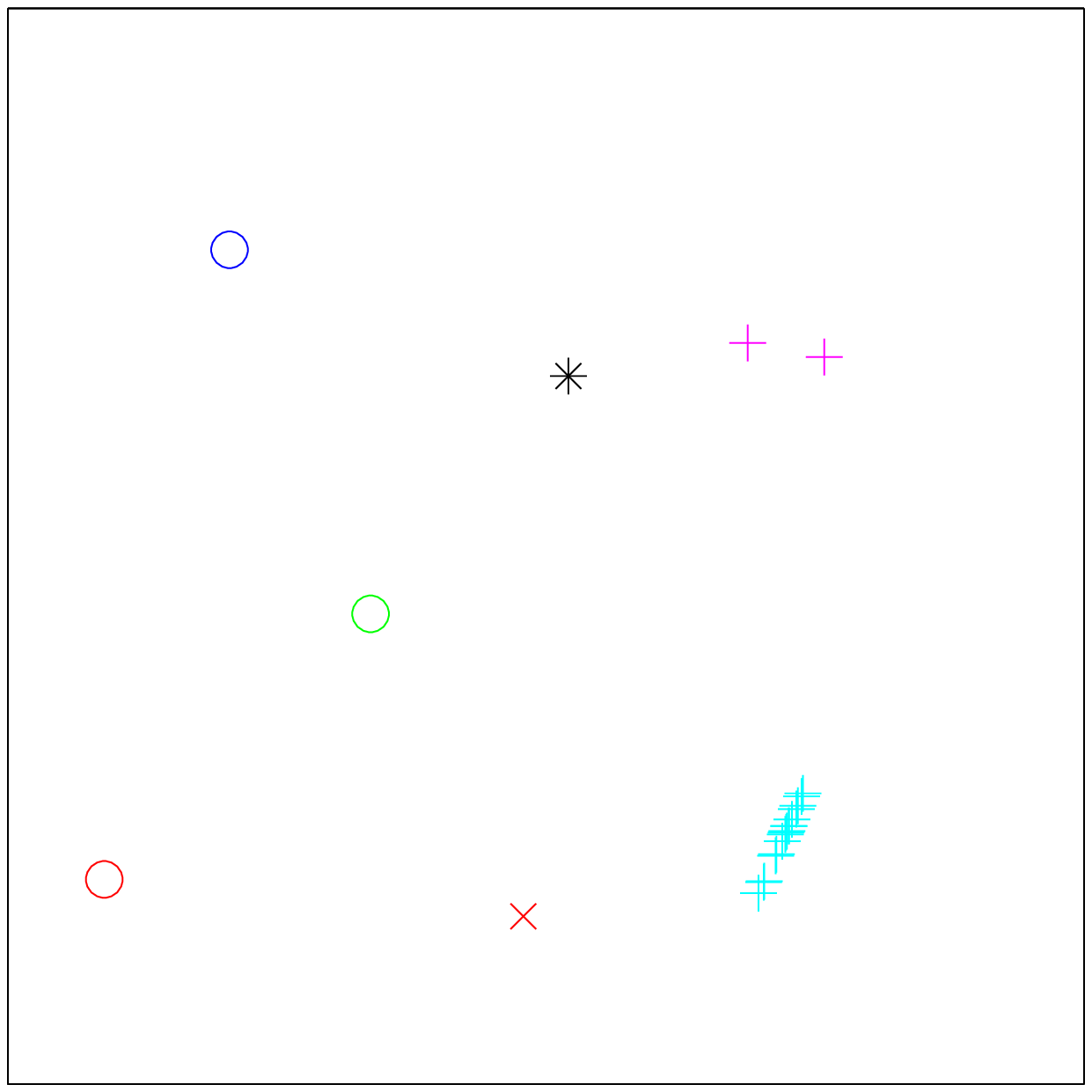} &
      \includegraphics[width=0.16\textwidth,bb=138 226 494 581,clip]{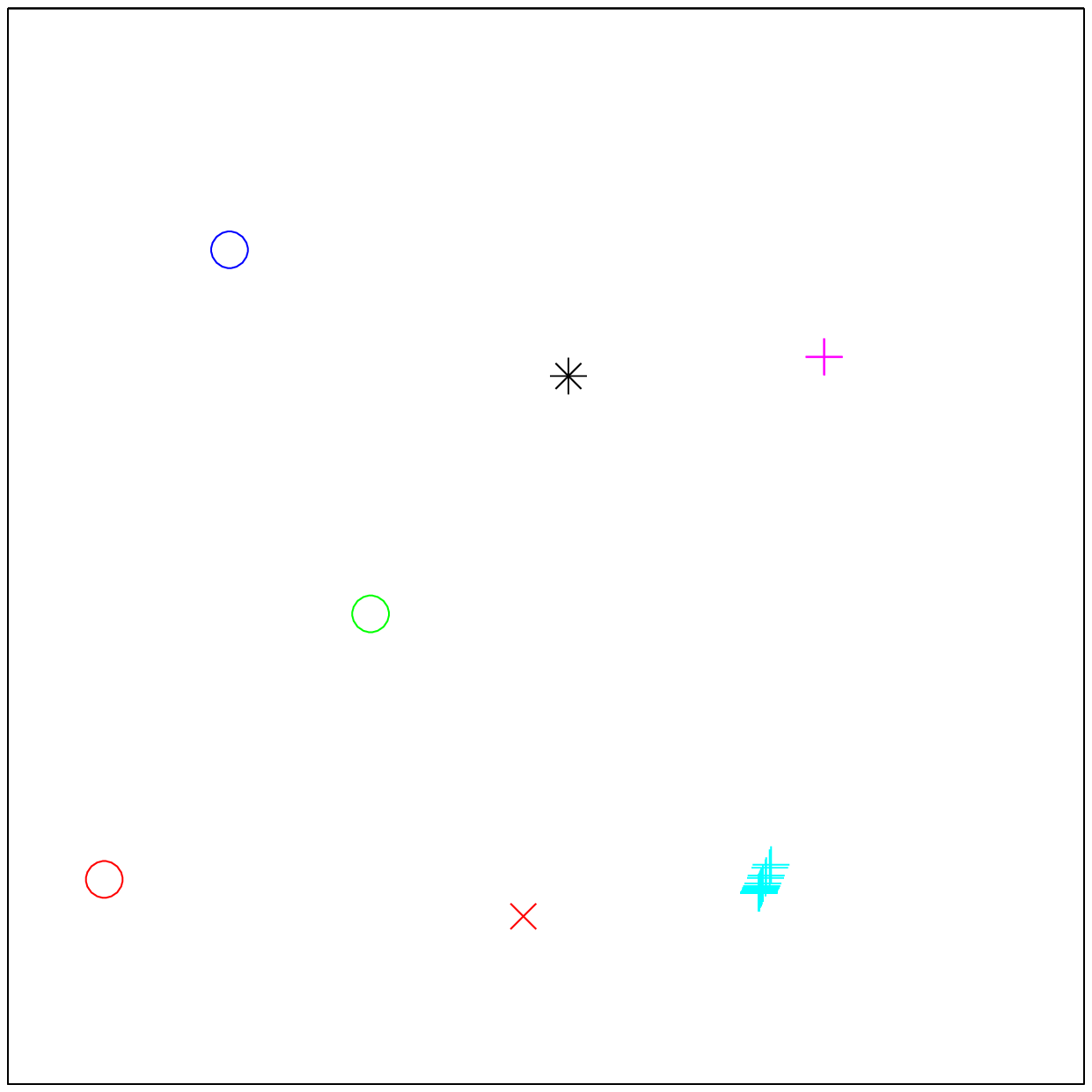} &
      \includegraphics[width=0.16\textwidth,bb=138 226 494 581,clip]{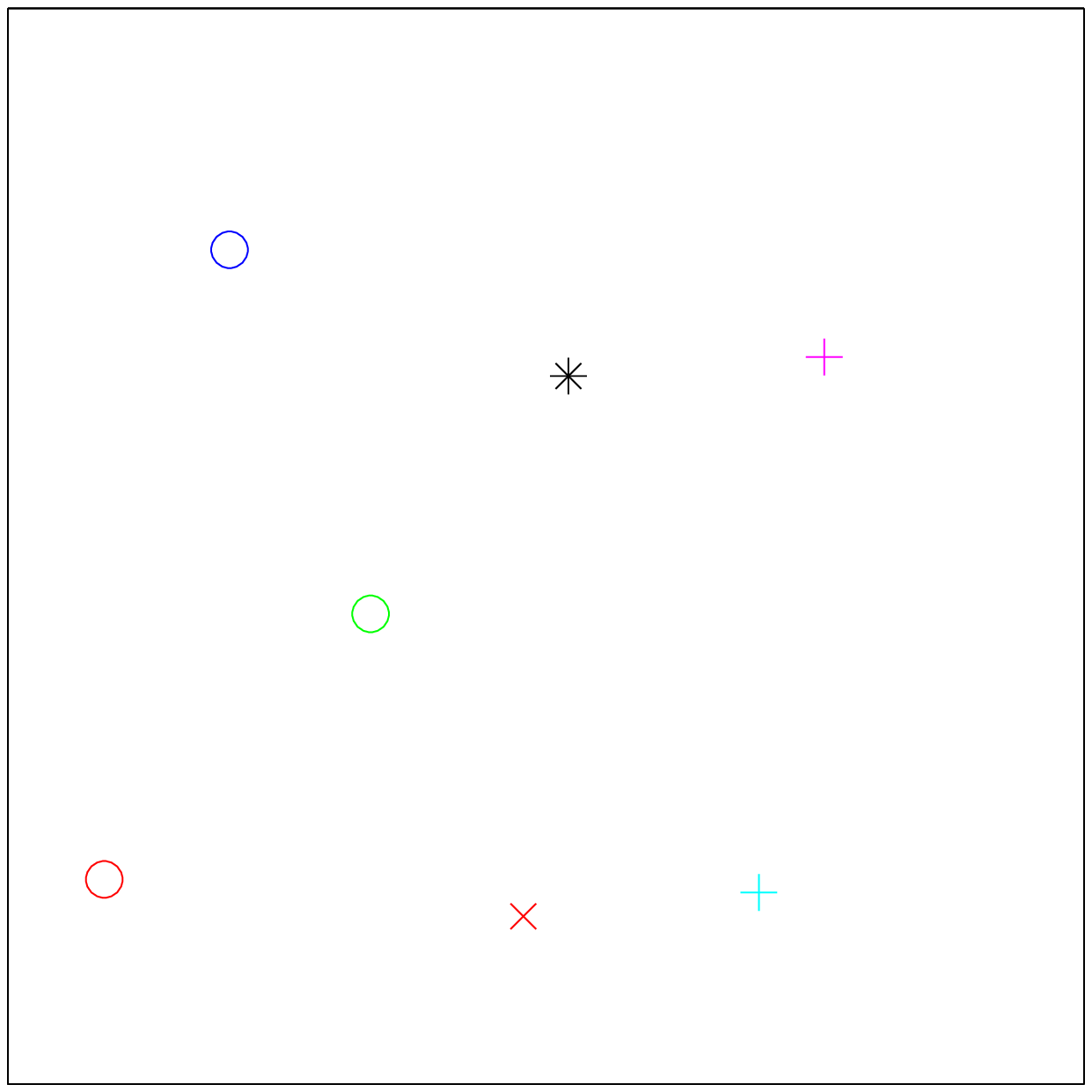} &
      \includegraphics[width=0.16\textwidth,bb=138 226 494 581,clip]{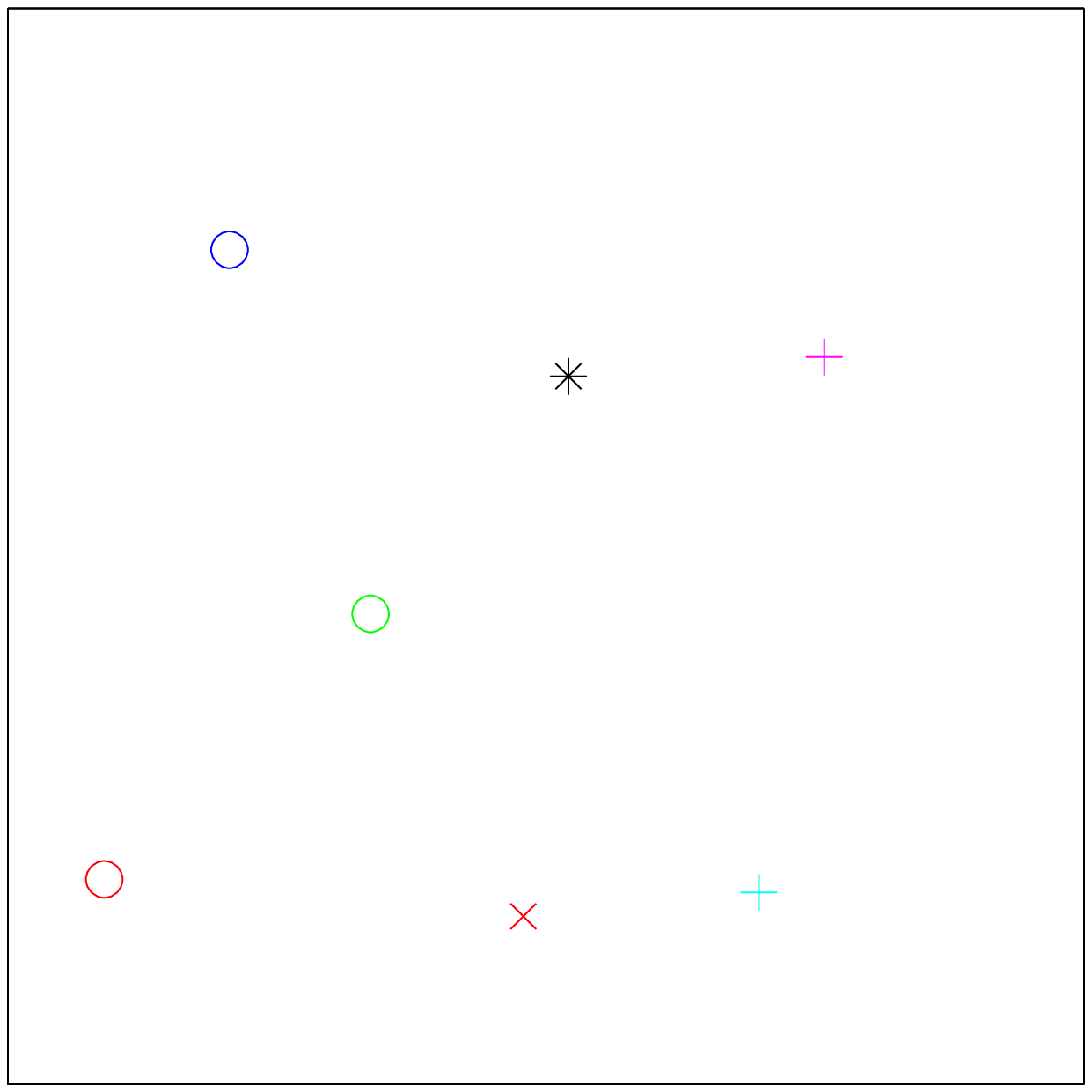}
    \end{tabular}
    \caption{Sequence of datasets $\smash{\X^{(0)},\dots,\X^{(11)}}$ obtained by Gaussian BMS for the \texttt{cameraman} image of fig.~\ref{f:MS_cameraman} with bandwidth $\sigma = 6$, resulting in 7 clusters. We show the 2D projection on the spatial domain of every data point (pixel), colored by cluster, as in fig.~\ref{f:MS_cameraman}\textbf{D}. Note: (1) points very quickly move towards a centroid and collapse into it (clustering property); (2) for each cluster-to-be, the local direction of maximum variance collapses much more slowly than the lower-variance directions, producing linearly shaped clusters (denoising property) that straighten and shorten. The BMS stopping criterion stopped BMS at iteration $\tau = 11$, where the dataset consists of 7 clusters of coincident points. If not stopping, these clusters would keep moving and eventually merge into a single cluster.}
    \label{f:BMS_cameraman}
  \end{center}
\end{figure*}

Fig.~\ref{f:MS_cameraman} shows an example with a grayscale image of $50 \times 50$ pixels. Thus, the dataset contains $N = 2\,500$ points in 3D (location $i$,$j$ and intensity $I$). Fig.~\ref{f:MS_cameraman} shows the result with MS while fig.~\ref{f:BMS_cameraman} shows the result with BMS. As noted earlier, MS and BMS generally give similar clusterings over a range of bandwidths.

With the Gaussian kernel, reasonable clusterings arise for bandwidths around $\frac{1}{5}$ of the image size (vertical or horizontal), although one should explore a range of bandwidths. In some computer vision applications, MS is sometimes used with a very small bandwidth in order to oversegment an image into uniform patches which are then further processed.

\paragraph{Cases when mean-shift clustering does not work well}

\begin{figure}[t]
  \begin{center}
    \begin{tabular}[c]{@{}c@{\hspace{0.05\linewidth}}c@{}}
      \begin{tabular}[c]{@{}c@{}c@{}}
        \psfrag{s}[r][r]{\raisebox{-1.3ex}{$\sigma\rightarrow$}\hspace{2ex}}
        \includegraphics[width=0.25\linewidth]{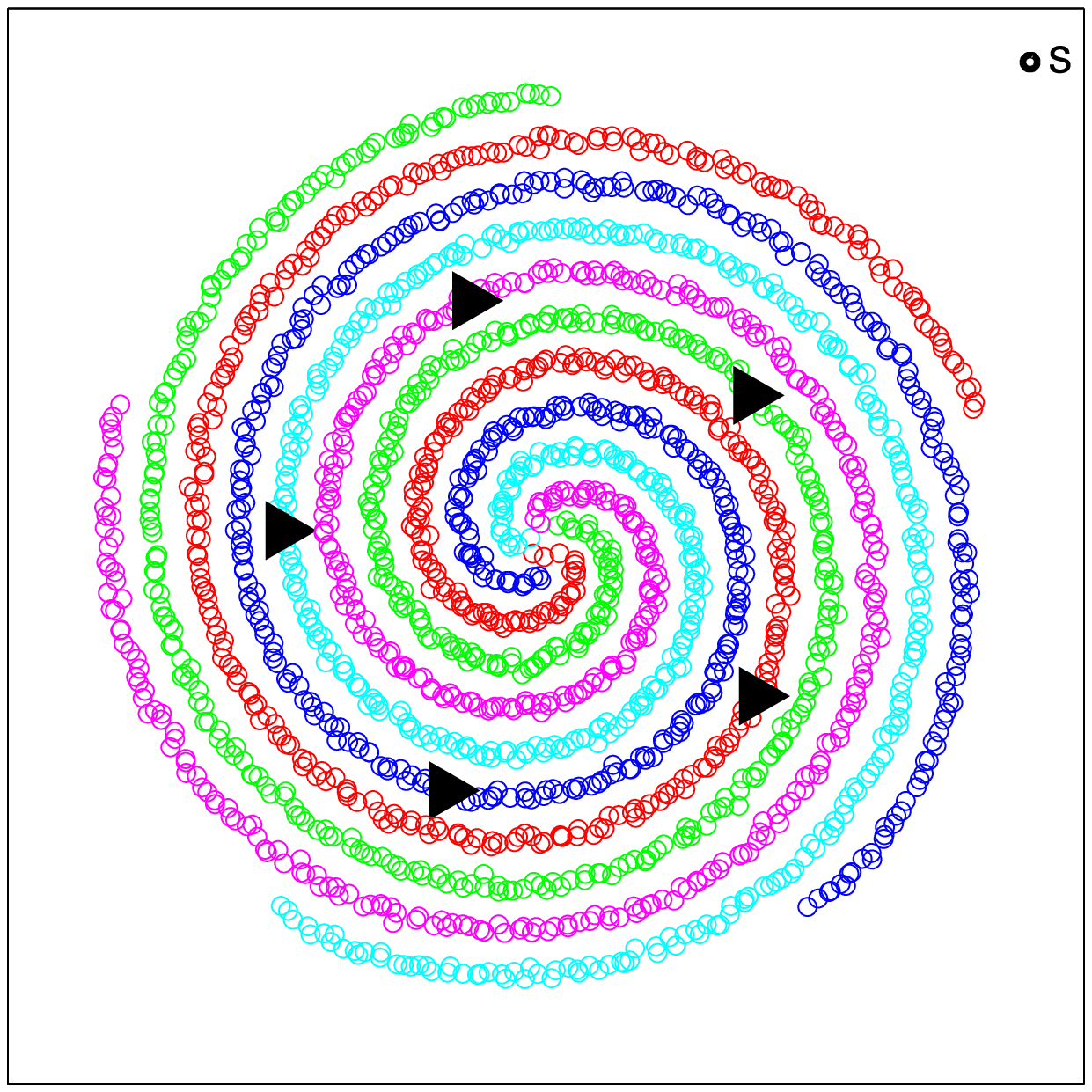} &
        \includegraphics[width=0.25\linewidth]{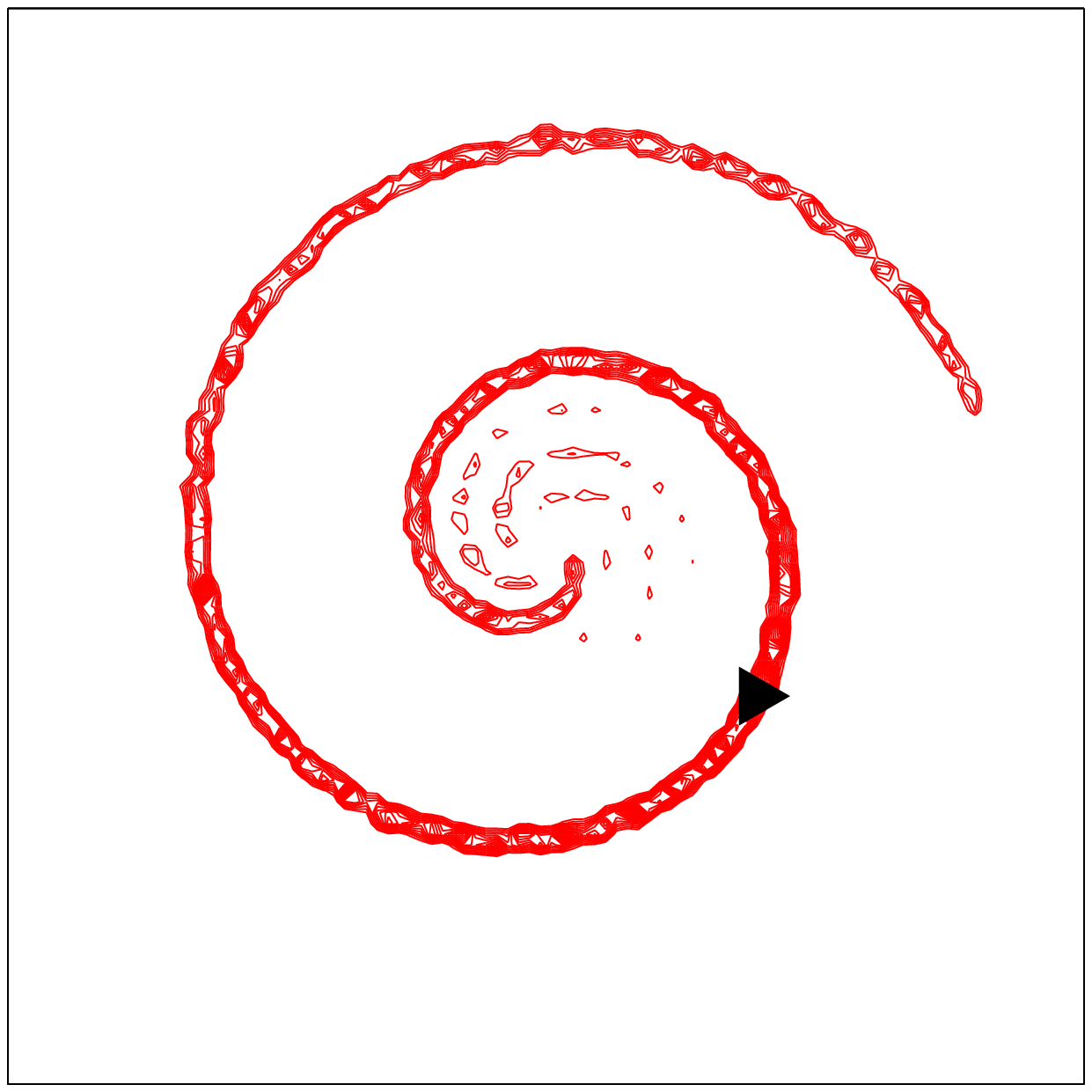}
      \end{tabular} &
      \begin{tabular}[c]{@{}c@{\hspace{0\linewidth}}c@{\hspace{0\linewidth}}c@{\hspace{0\linewidth}}c@{\hspace{0\linewidth}}c@{\hspace{0\linewidth}}c@{\hspace{0\linewidth}}c@{\hspace{0\linewidth}}c@{\hspace{0\linewidth}}c@{\hspace{0\linewidth}}c@{\hspace{0\linewidth}}c@{}}
        \rotatebox{90}{\scriptsize\hspace{1ex}$K$-means} &
        \includegraphics[width=0.080\linewidth,bb=142 226 494 578,clip]{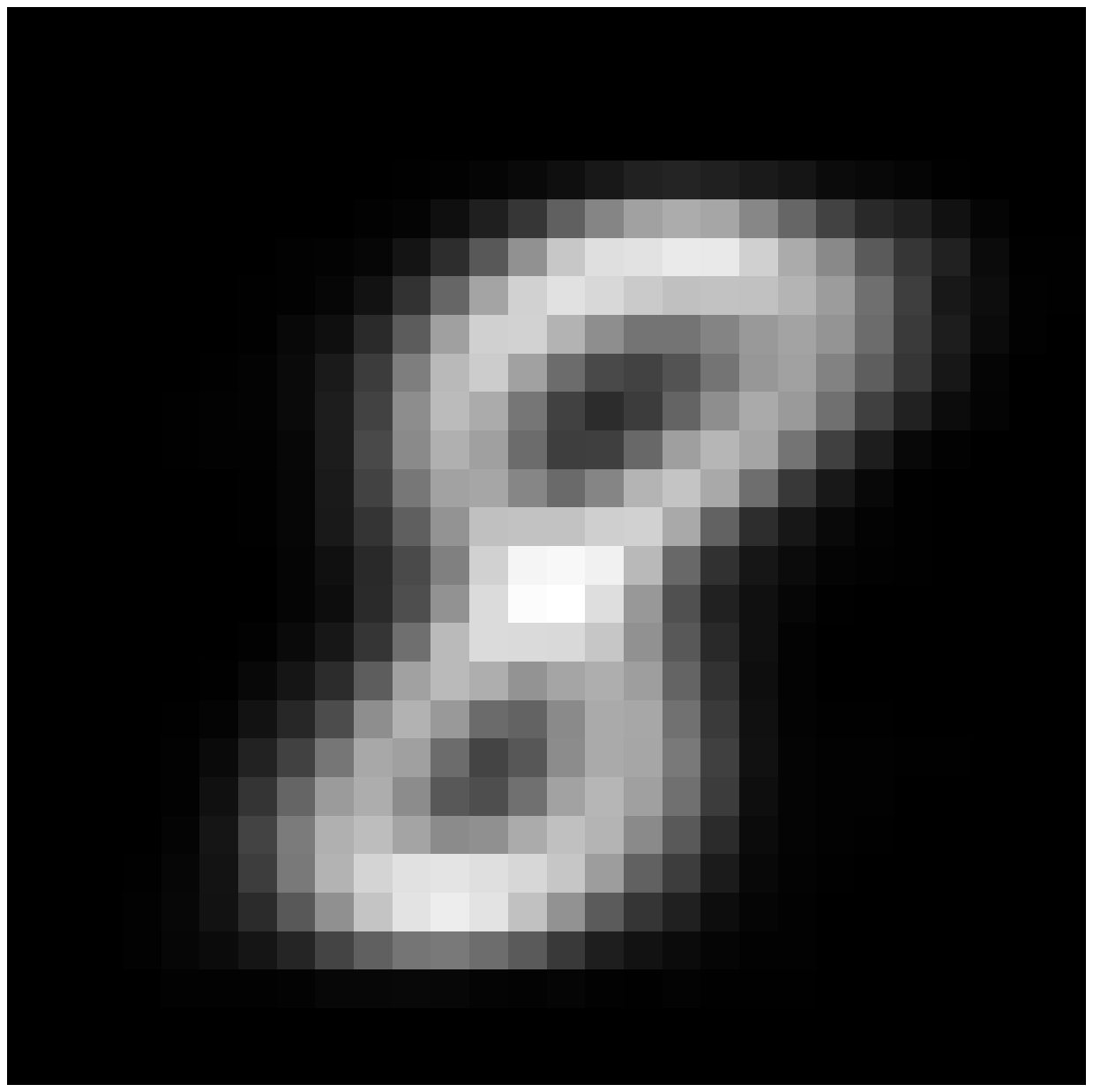} &
        \includegraphics[width=0.080\linewidth,bb=142 226 494 578,clip]{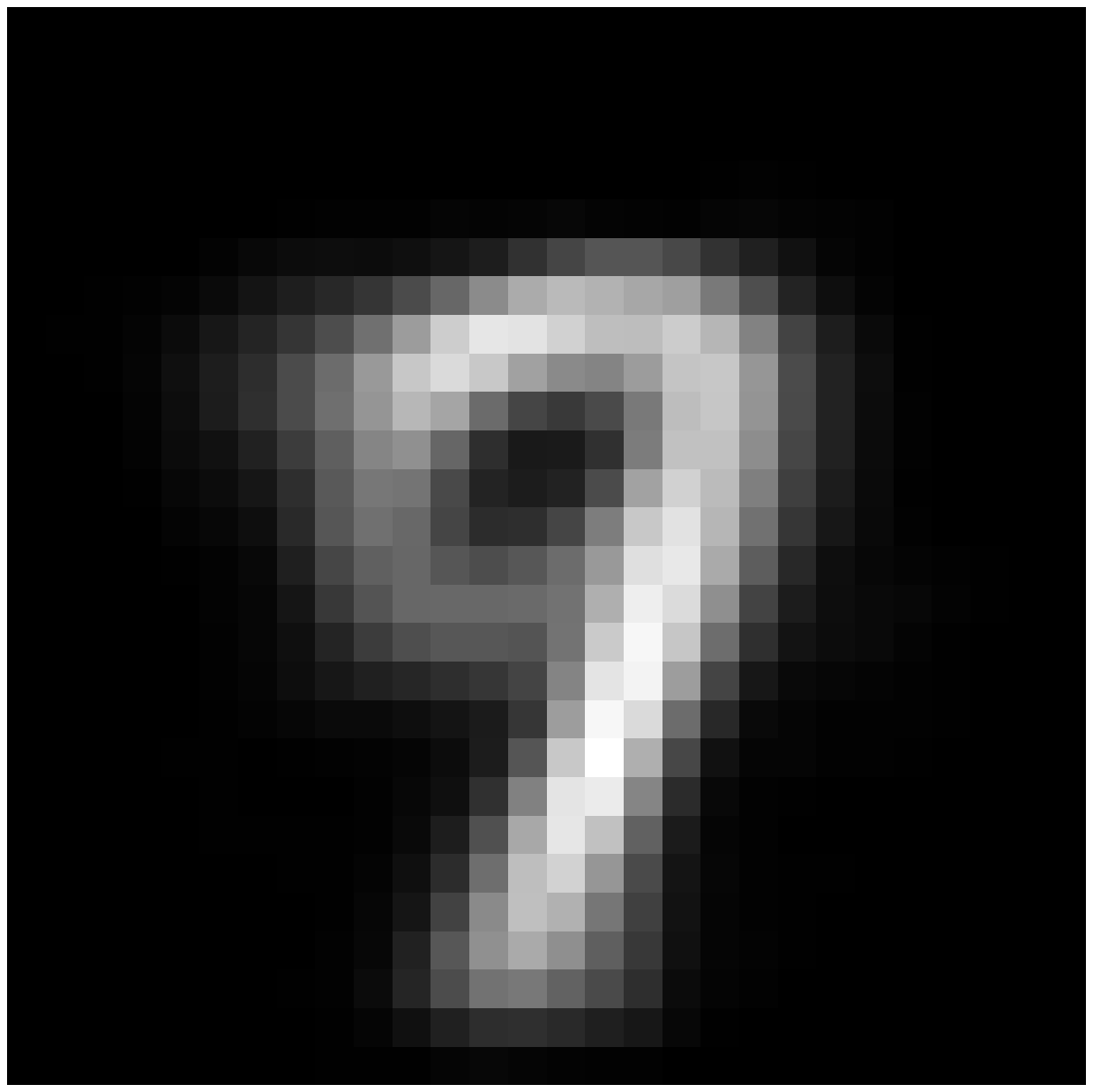} &
        \includegraphics[width=0.080\linewidth,bb=142 226 494 578,clip]{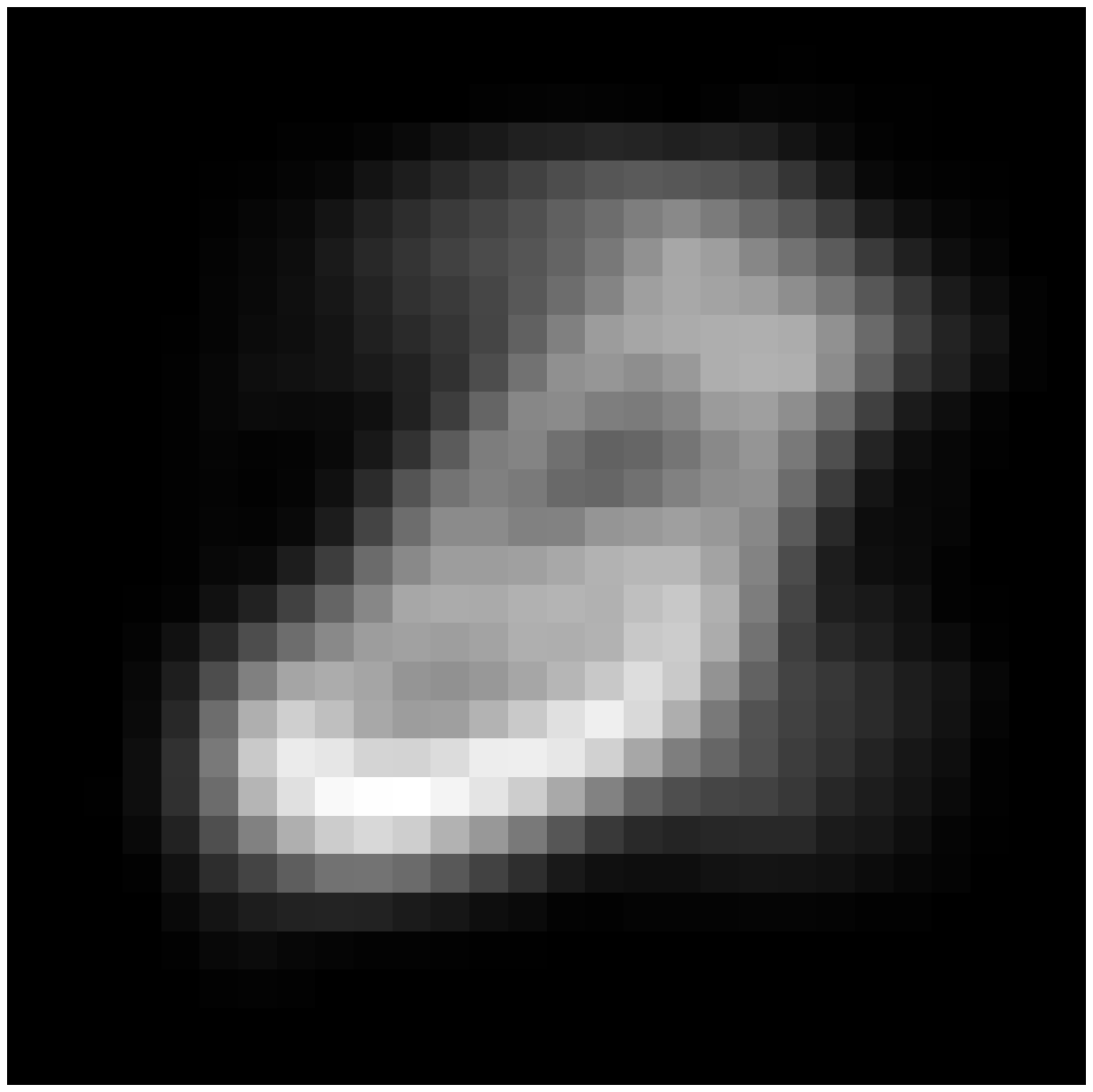} &
        \includegraphics[width=0.080\linewidth,bb=142 226 494 578,clip]{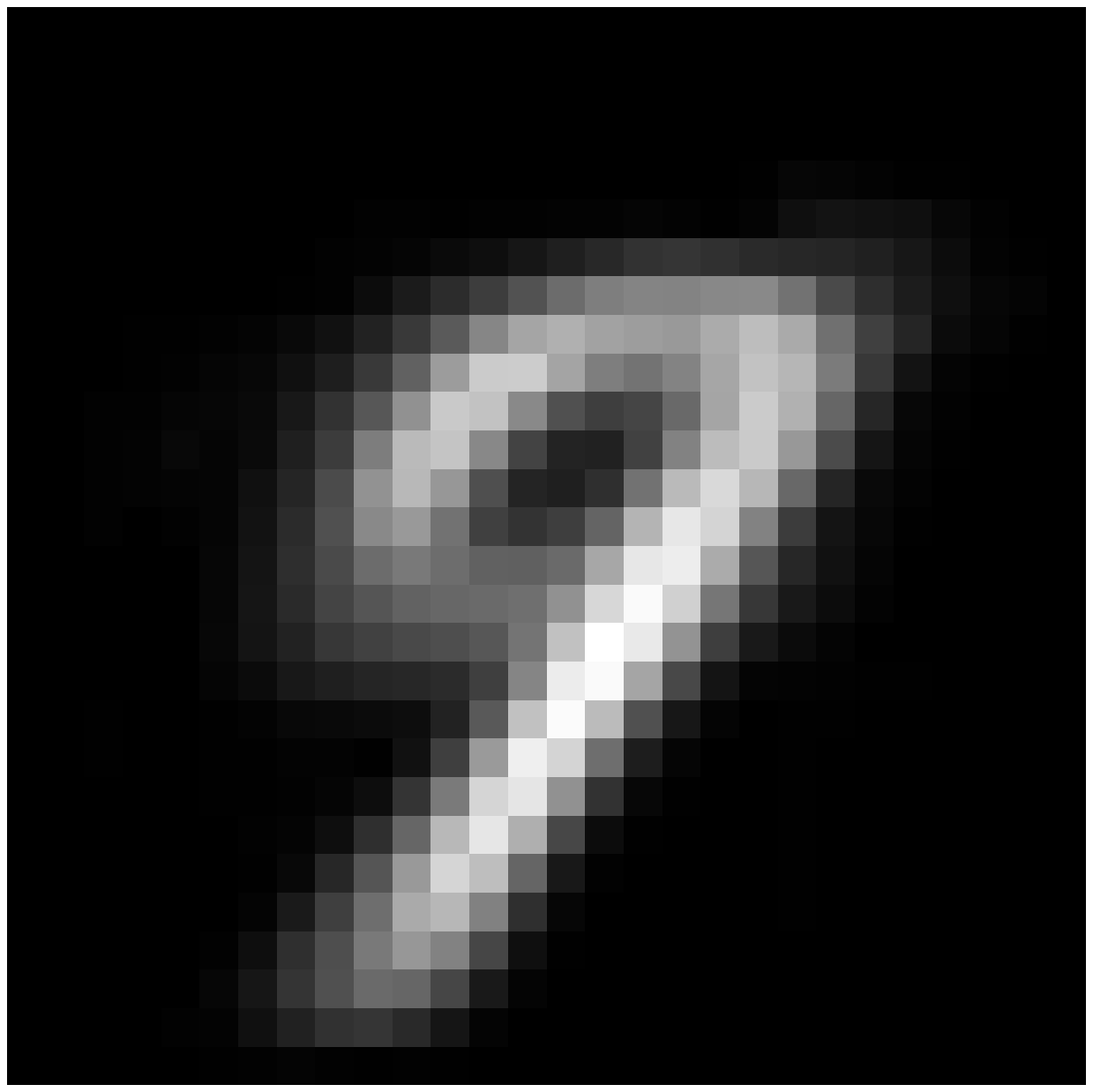} &
        \includegraphics[width=0.080\linewidth,bb=142 226 494 578,clip]{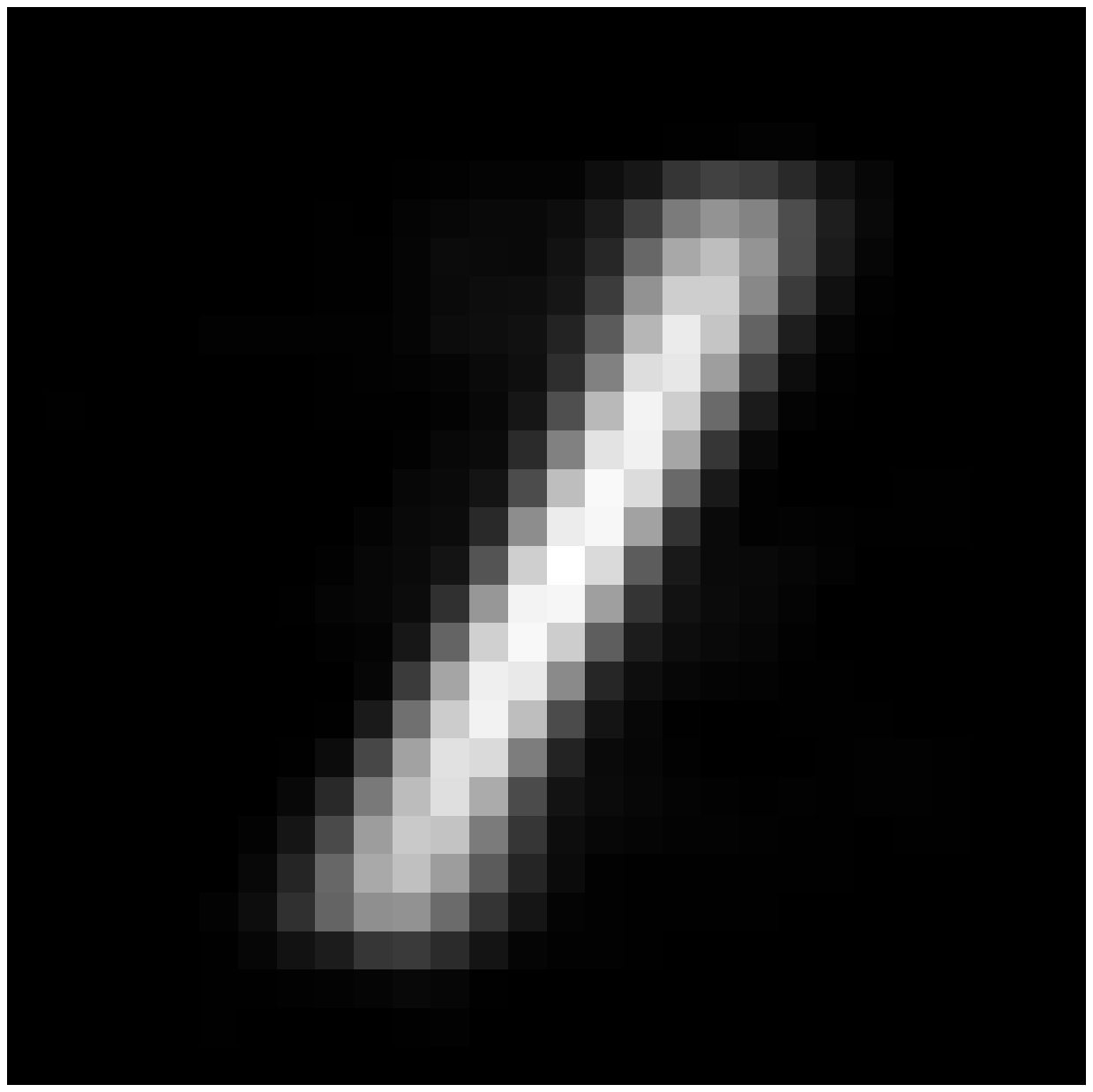} \\[-1ex]
        \rotatebox{90}{\scriptsize\hspace{0.5ex}\caja{c}{c}{Laplacian \\ $K$-modes}} &
        \includegraphics[width=0.080\linewidth,bb=142 226 494 578,clip]{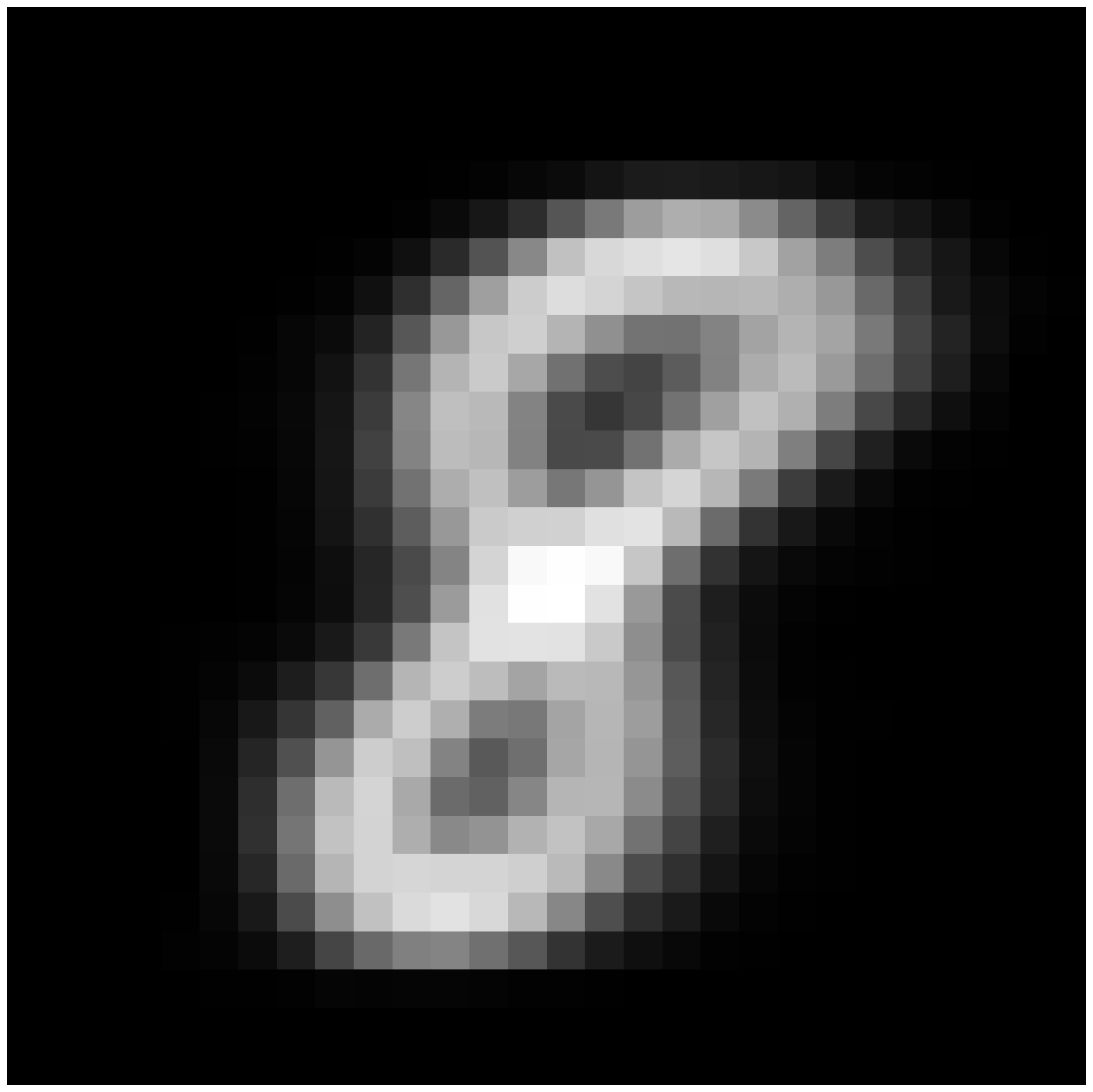} &
        \includegraphics[width=0.080\linewidth,bb=142 226 494 578,clip]{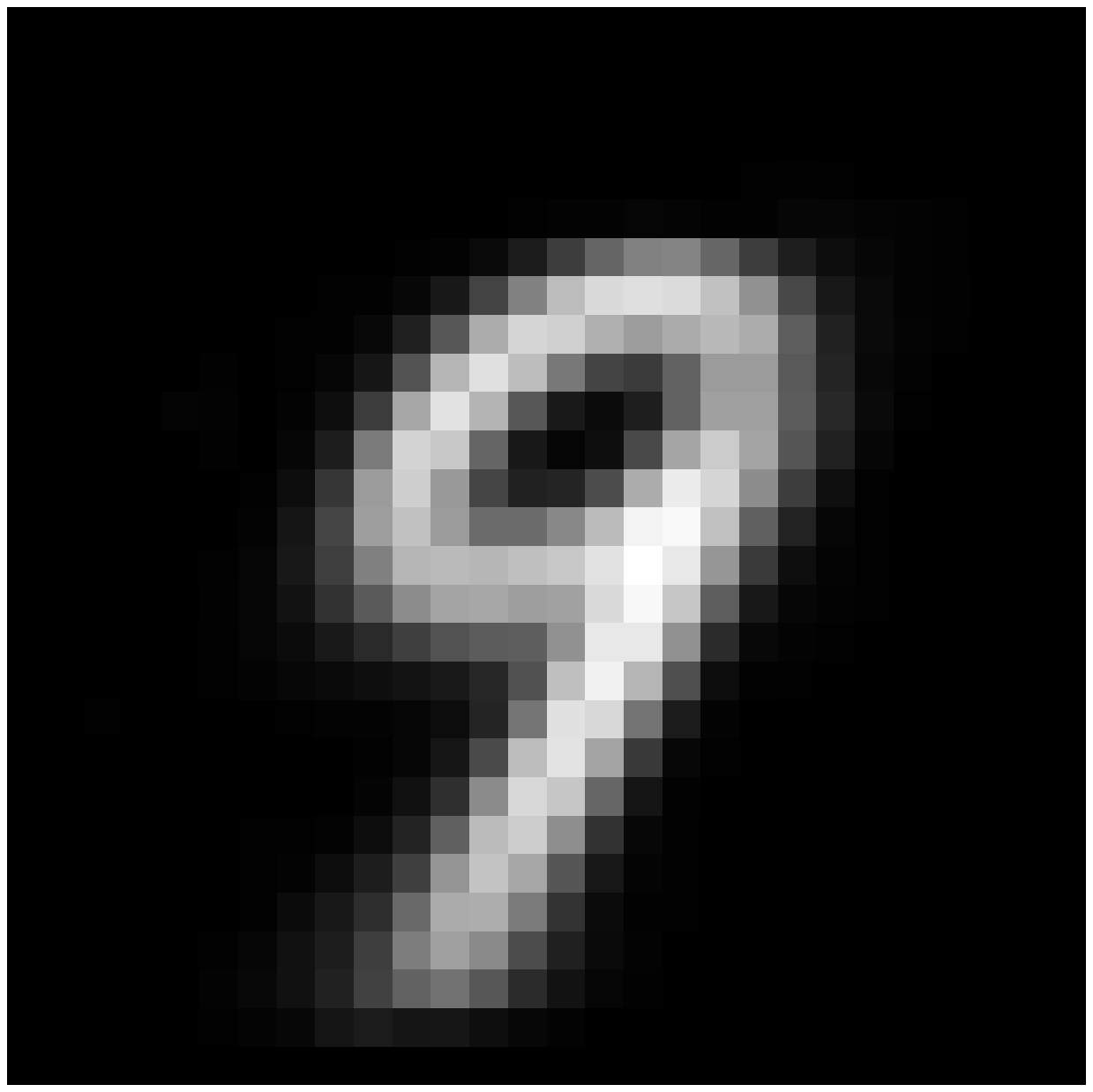} &
        \includegraphics[width=0.080\linewidth,bb=142 226 494 578,clip]{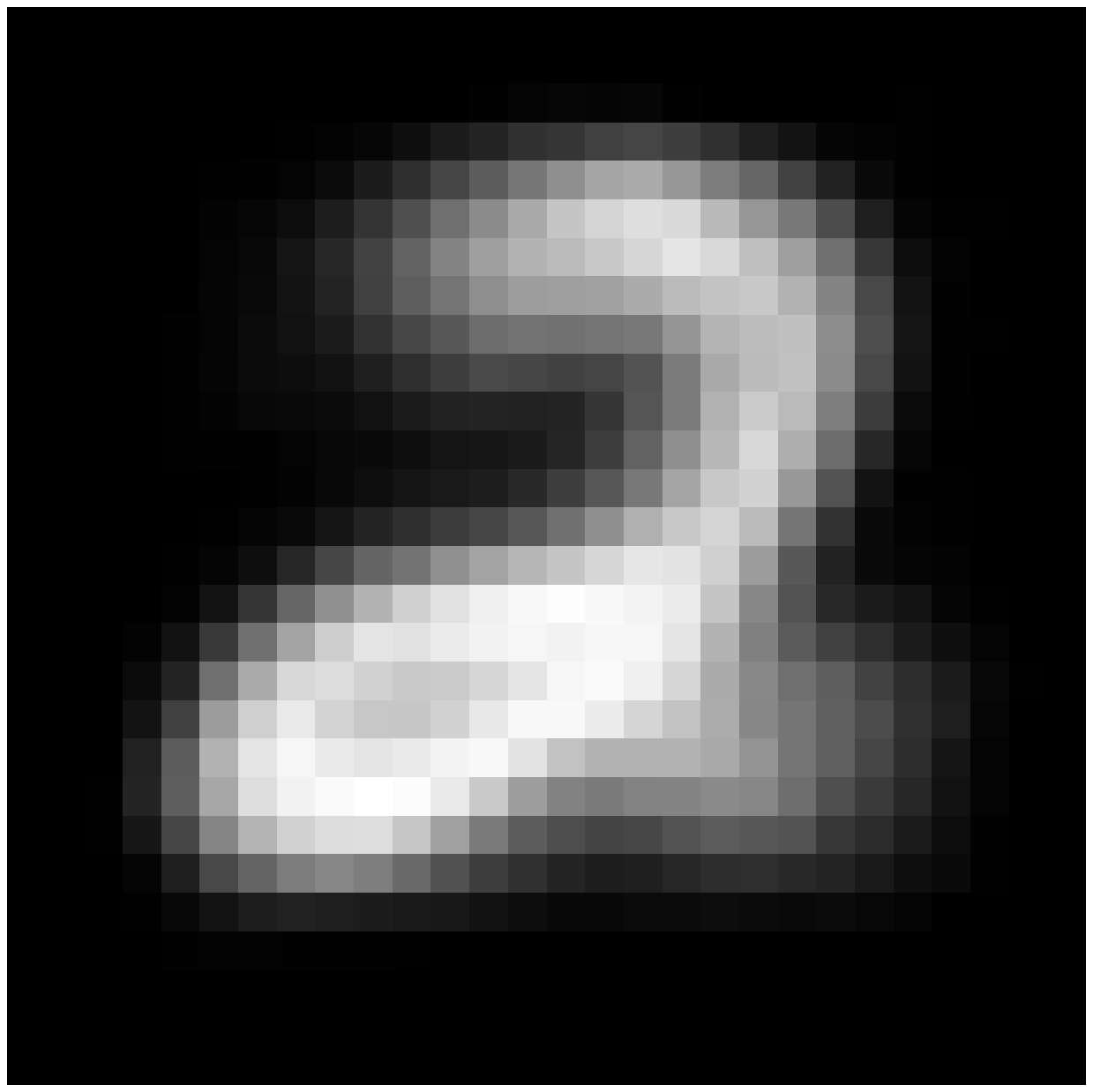} &
        \includegraphics[width=0.080\linewidth,bb=142 226 494 578,clip]{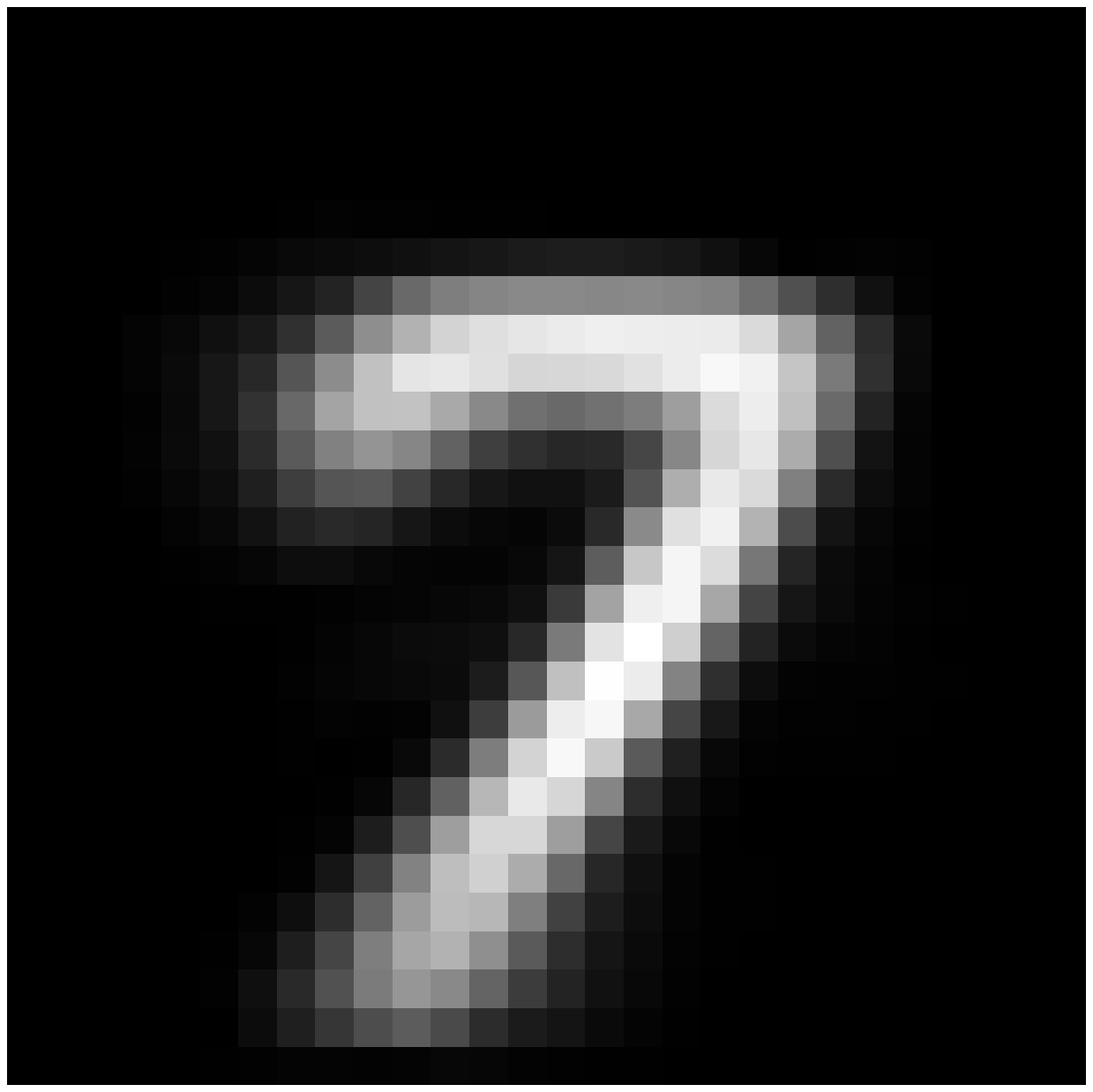} &
        \includegraphics[width=0.080\linewidth,bb=142 226 494 578,clip]{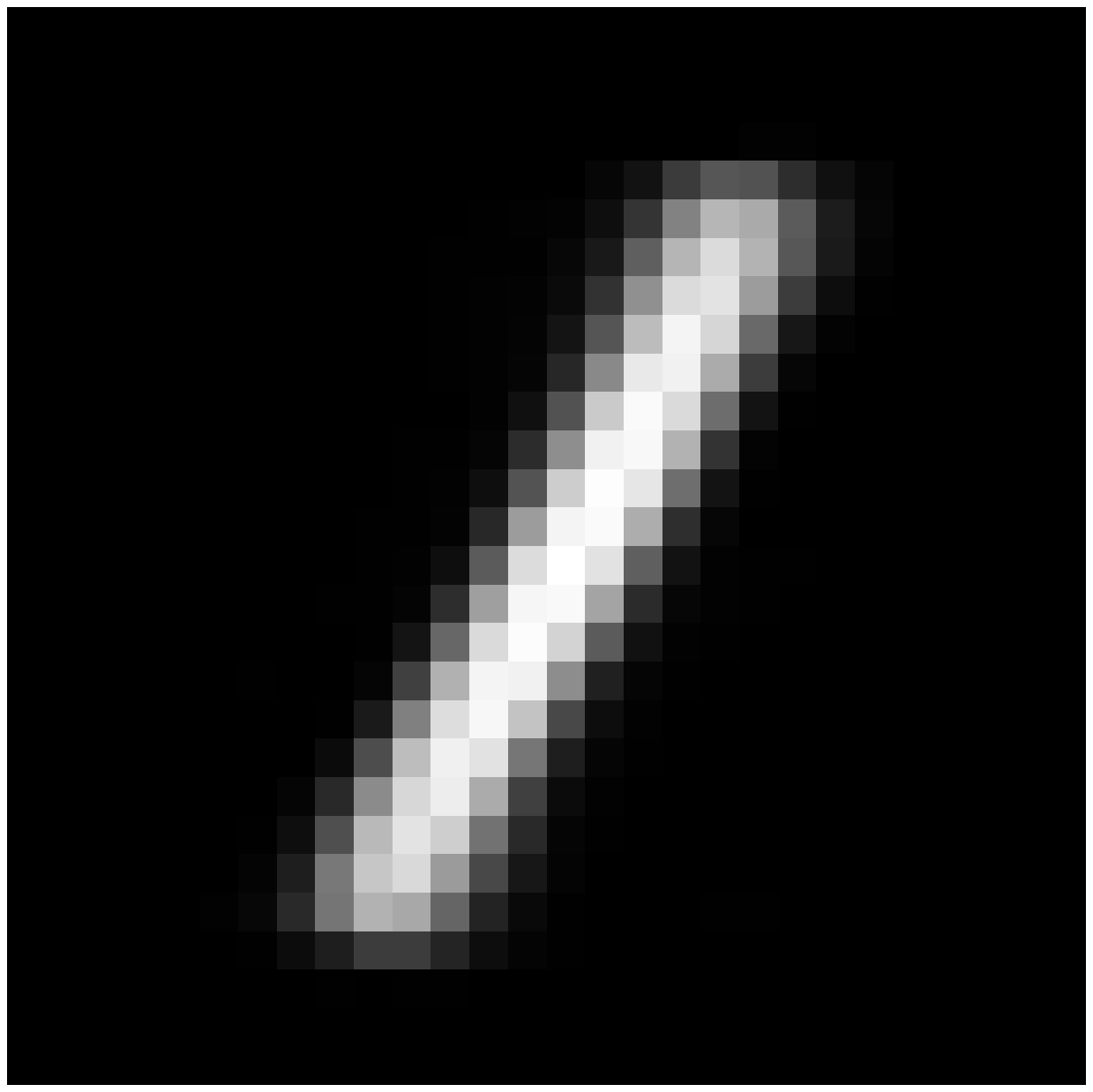} \\[-1ex]
        \rotatebox{90}{\scriptsize\hspace{2ex}\caja{c}{c}{Mean \\ shift}} &
        \includegraphics[width=0.080\linewidth,bb=142 226 494 578,clip]{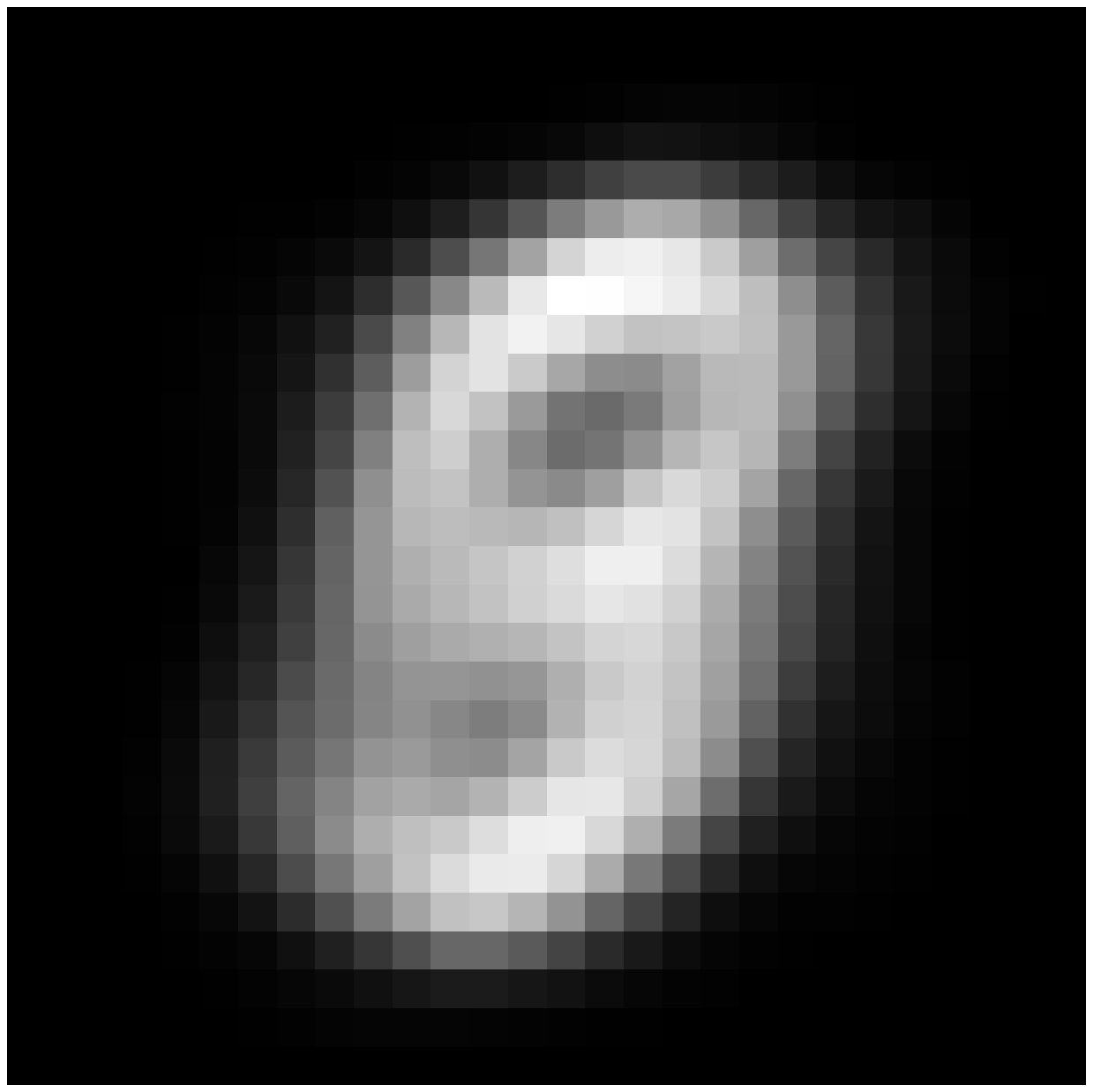} &
        \includegraphics[width=0.080\linewidth,bb=142 226 494 578,clip]{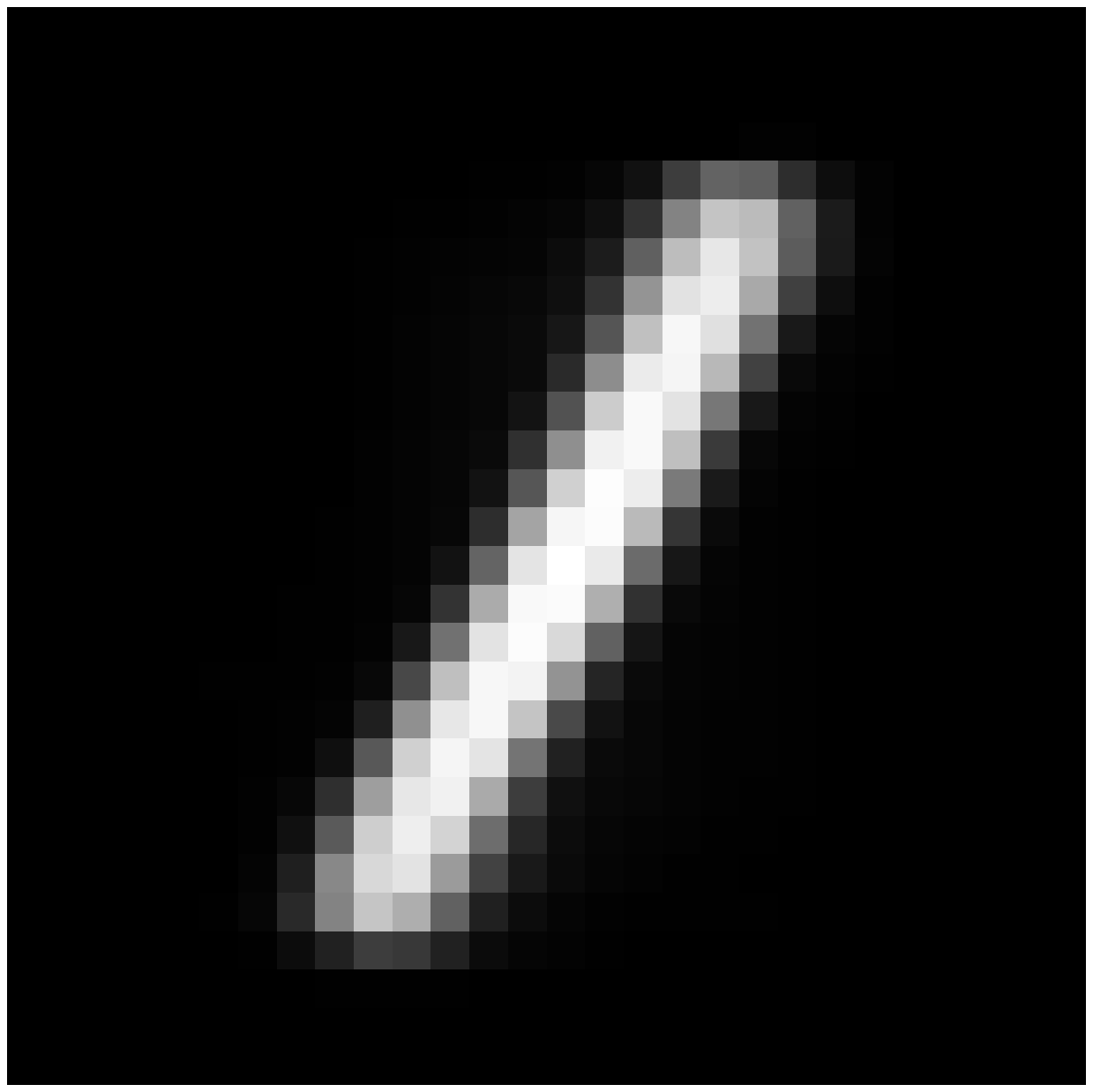} &
        \includegraphics[width=0.080\linewidth,bb=142 226 494 578,clip]{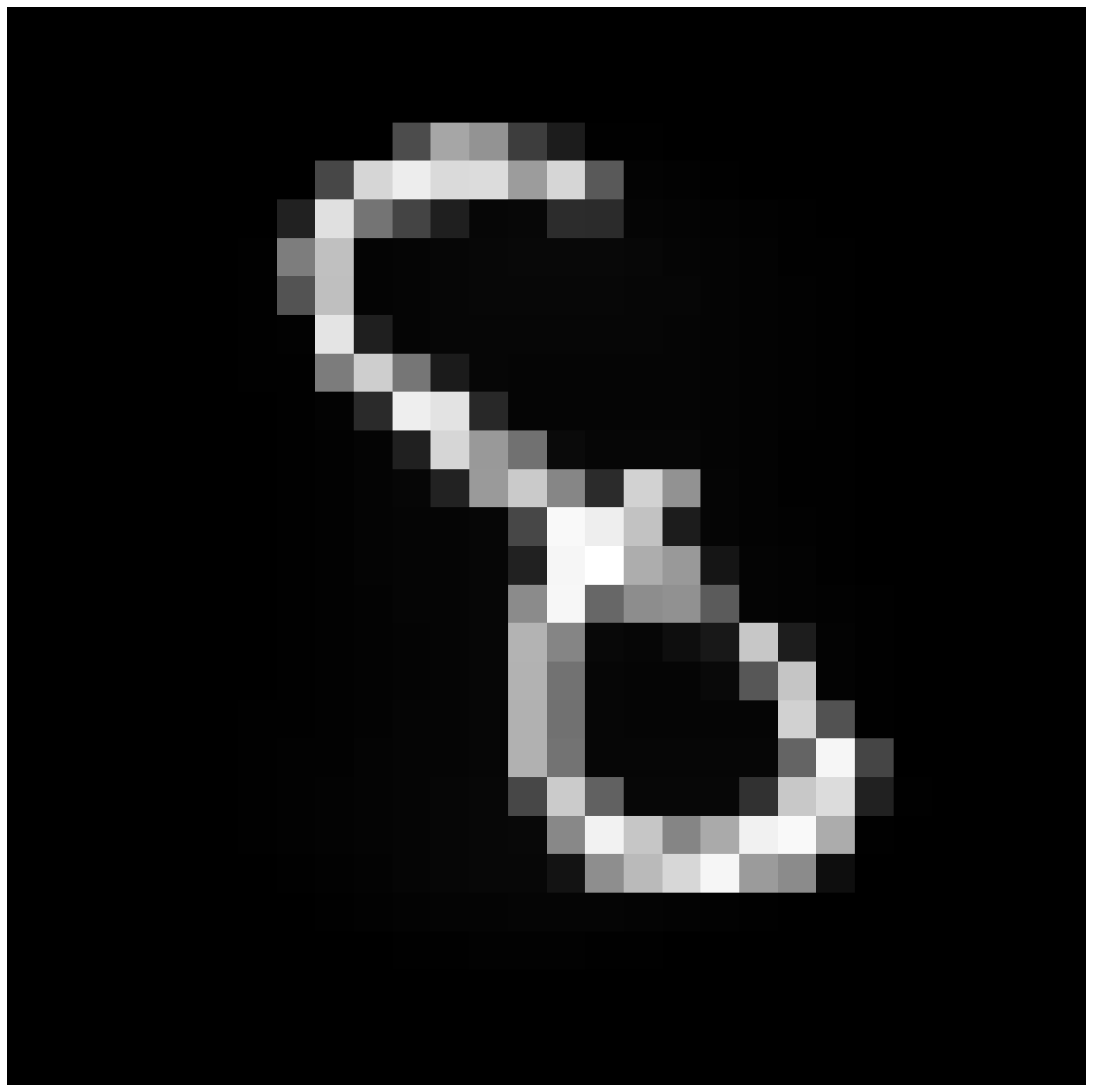} &
        \includegraphics[width=0.080\linewidth,bb=142 226 494 578,clip]{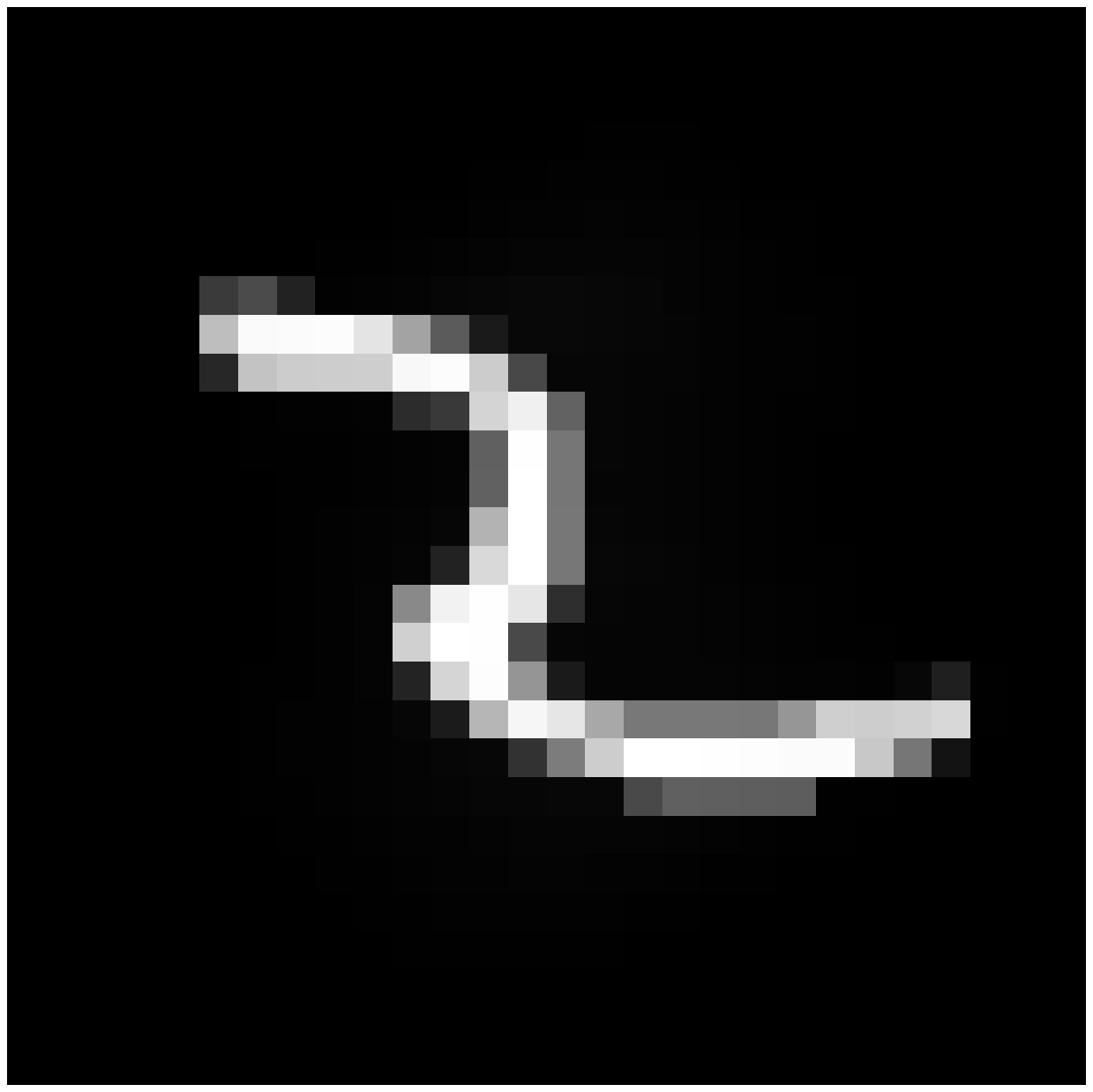} &
        \includegraphics[width=0.080\linewidth,bb=142 226 494 578,clip]{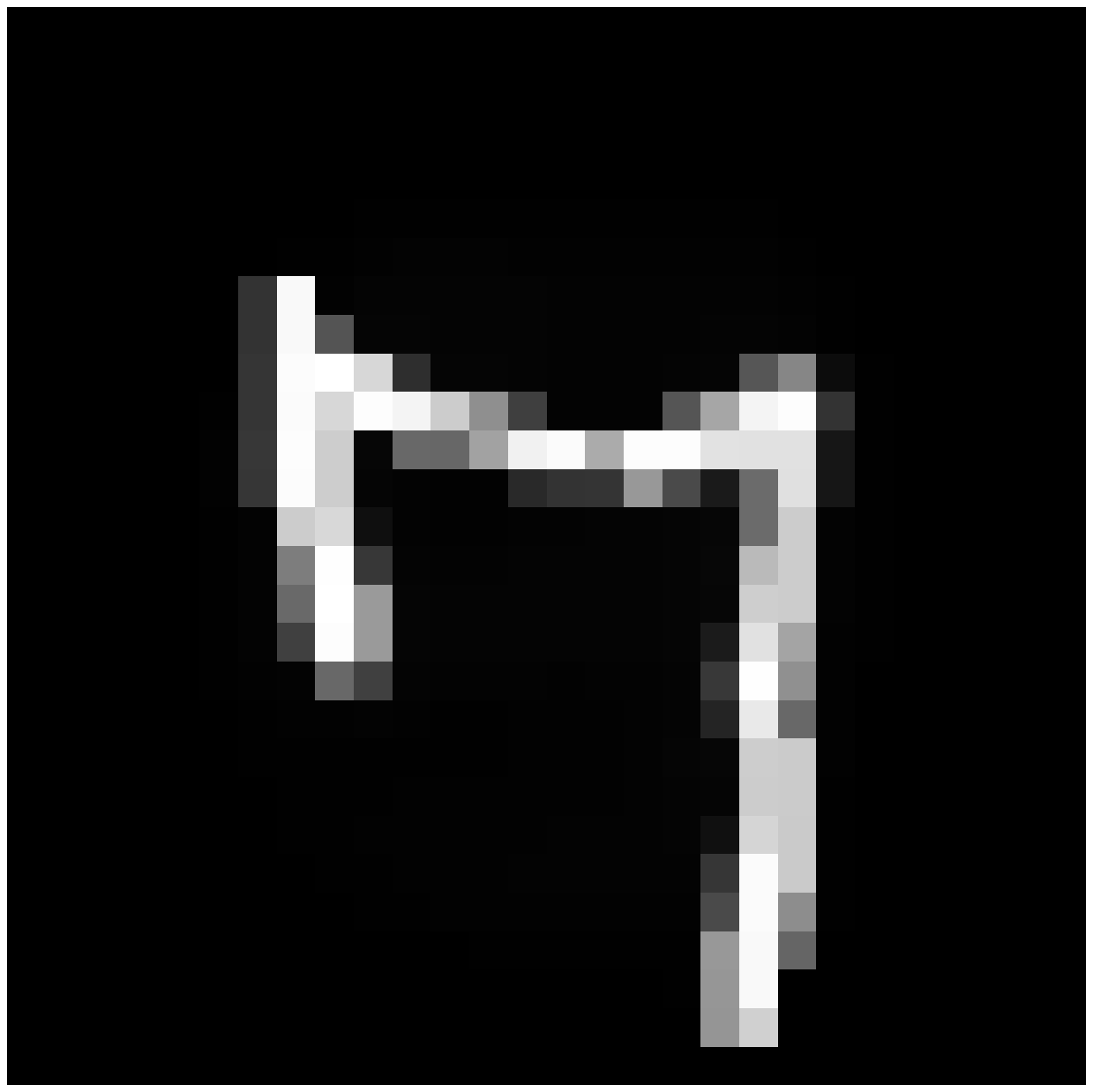}
      \end{tabular}
    \end{tabular}
    \caption{\emph{Left}: clustering of 5 spirals using Laplacian $K$-modes ($K=5$, $\sigma=0.2$, the circle at the top right corner has a radius of $\sigma$), with modes denoted by $\blacktriangleright$, and contours of the KDE of the ``red'' cluster. \emph{Right}: a subset of the $10$ centroids obtained by $K$-means, Laplacian $K$-modes and mean shift for MNIST data.}
    \label{f:LaplacianKmodes}
  \end{center}
\end{figure}

Fig.~\ref{f:LaplacianKmodes} illustrates two situations where MS (and BMS) do not produce good clusterings. The first one is data having manifold structure (of lower dimension than the feature space), as in the 1D spirals in a 2D space shown in the left panel. The KDE will have either many modes spread over the data manifolds (for smaller bandwidths) or few modes that mix manifolds (for larger bandwidths). No bandwidth value will cluster the spirals correctly. The Laplacian $K$-modes algorithm is able to cluster the spirals correctly, while estimating a reasonable KDE and mode for each spiral \cite{WangCarreir14c}.

The second situation is high-dimensional data, as in the MNIST handwritten digits \cite{Lecun_98a}, where each data point is a grayscale image of $28 \times 28$ pixels (i.e., $D = 784$ feature dimensions). In high-dimensional spaces, MS tends to produce either a single mode (for larger bandwidths), which is uninformative, or many modes (for smaller bandwidths), which often sit on small groups of points or outliers. In the results in fig.~\ref{f:LaplacianKmodes} (right panel), the MS bandwidth was tuned to produce $K=10$ clusters, and $K$-means and Laplacian $K$-modes were run with $K=10$. Since the data forms nonconvex clusters, $K$-means produces centroids that average distant data points, of possibly different digit classes, and thus do not look like digits. Again, Laplacian $K$-modes can partition the data into exactly $K$ clusters, while finding a mode for each that looks like a valid digit and is more representative of its class \cite{WangCarreir14c}.

\paragraph{Multivalued regression and inversion}

In (univalued) regression, we assume there exists a mapping $\f\mathpunct{:}\ \x\in\bbR^d\rightarrow\y\in\bbR^D$ which assigns a unique vector \y\ to each possible input \x\ (i.e., \f\ is a function in the mathematical sense). We estimate \f\ given a training set of pairs $\{(\x_n,\y_n)\}^N_{n=1}$. This is the usual regression setting. In \emph{multivalued regression}, the mapping \f\ assigns possibly multiple vectors $\y_1,\y_2\dots$ to each input \x. A classical example is learning the inverse of a (univalued) mapping \g. For example, the inverse of $g(y) = y^2$ is $f(x) = \pm \sqrt{x}$, which has $0$, $1$ or $2$ outputs depending on the value of $x$ (smaller, equal or greater than zero, respectively). Carreira-Perpi{\~n}{\'a}n \cite{Carreir00a,Carreir01a,Carreir04a} proposed to represent a multivalued mapping $\f\mathpunct{:}\ \x\in\bbR^d\rightarrow\y\in\bbR^D$ by the modes of the conditional distribution $p(\y|\x)$. Hence, in this case, we apply mean-shift to a \emph{conditional distribution} $p(\y|\x)$. One advantage of this approach is that the number of inverses is selected automatically for each value of \x; essentially, it is the number of ``clusters'' found by mean-shift in $p(\y|\x)$. Once the modes have been found, one can also estimate error bars for them from the local Hessian at each mode \cite{Carreir00a,Carreir01a}. The conditional distribution may be obtained from a Gaussian KDE or Gaussian mixture $p(\x,\y)$ for the joint distribution of \x\ and \y\ (since the marginal and conditional distributions of any Gaussian mixture are also Gaussian mixtures), but it can also be learned directly, for example using mixture density networks or particle filters \cite{Bishop06a}.

\begin{figure}[t]
  \begin{center}
    \begin{tabular}[c]{@{}c@{\hspace{0.02\textwidth}}c@{\hspace{0.05\textwidth}}c@{}}
    \psfrag{theta1}[t][]{$\theta_1$}
    \psfrag{theta2}[][][1][-90]{$\theta_2$}
    \includegraphics[height=0.28\linewidth]{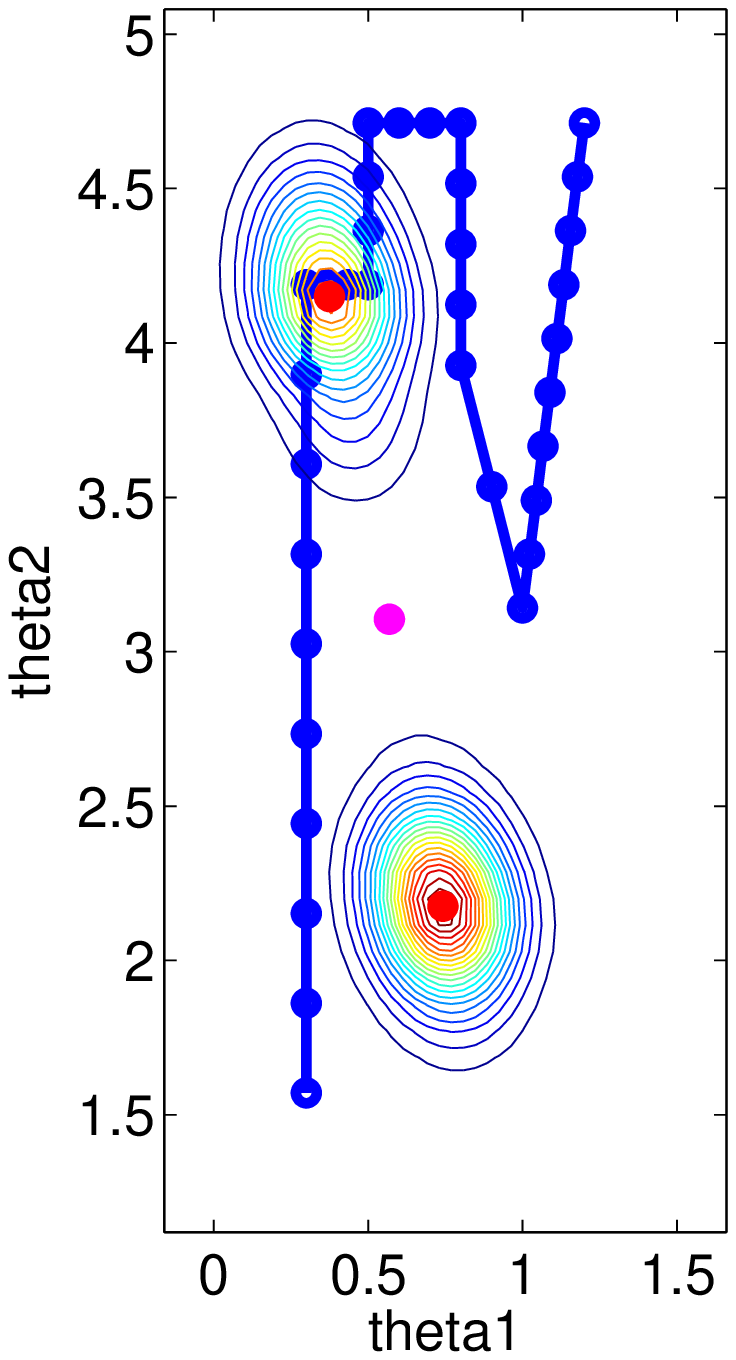} &
    \psfrag{x1}[t][]{$x_1$}
    \psfrag{x2}[][][1][-90]{$x_2$}
    \psfrag{t1}{$\theta_1$}
    \psfrag{t2}{\raisebox{-3ex}{$\theta_2$}}
    \includegraphics[height=0.28\linewidth]{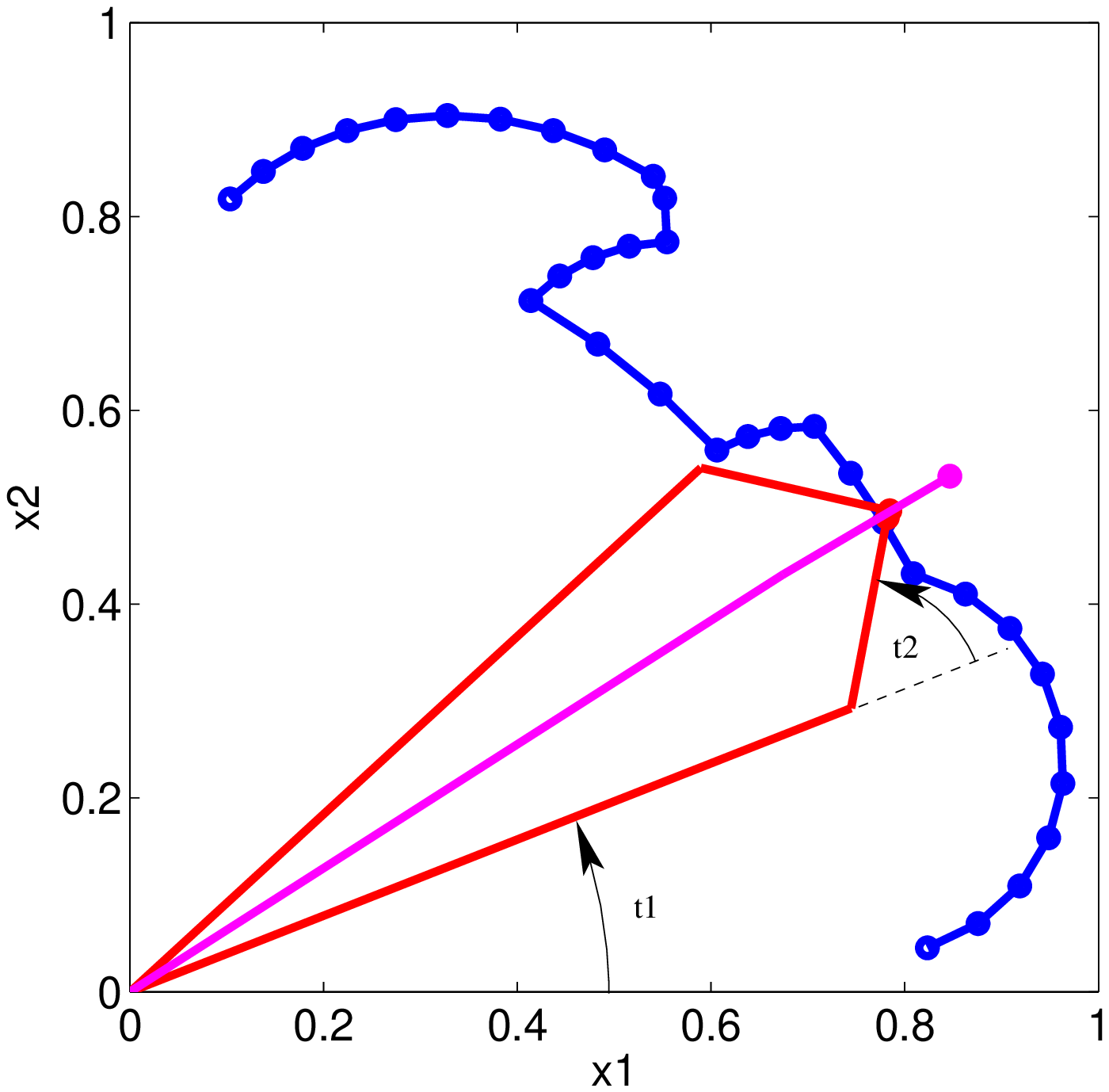} &
    \raisebox{0.5em}{\fbox{\includegraphics[height=0.25\textwidth,bb=130 386 374 517,clip]{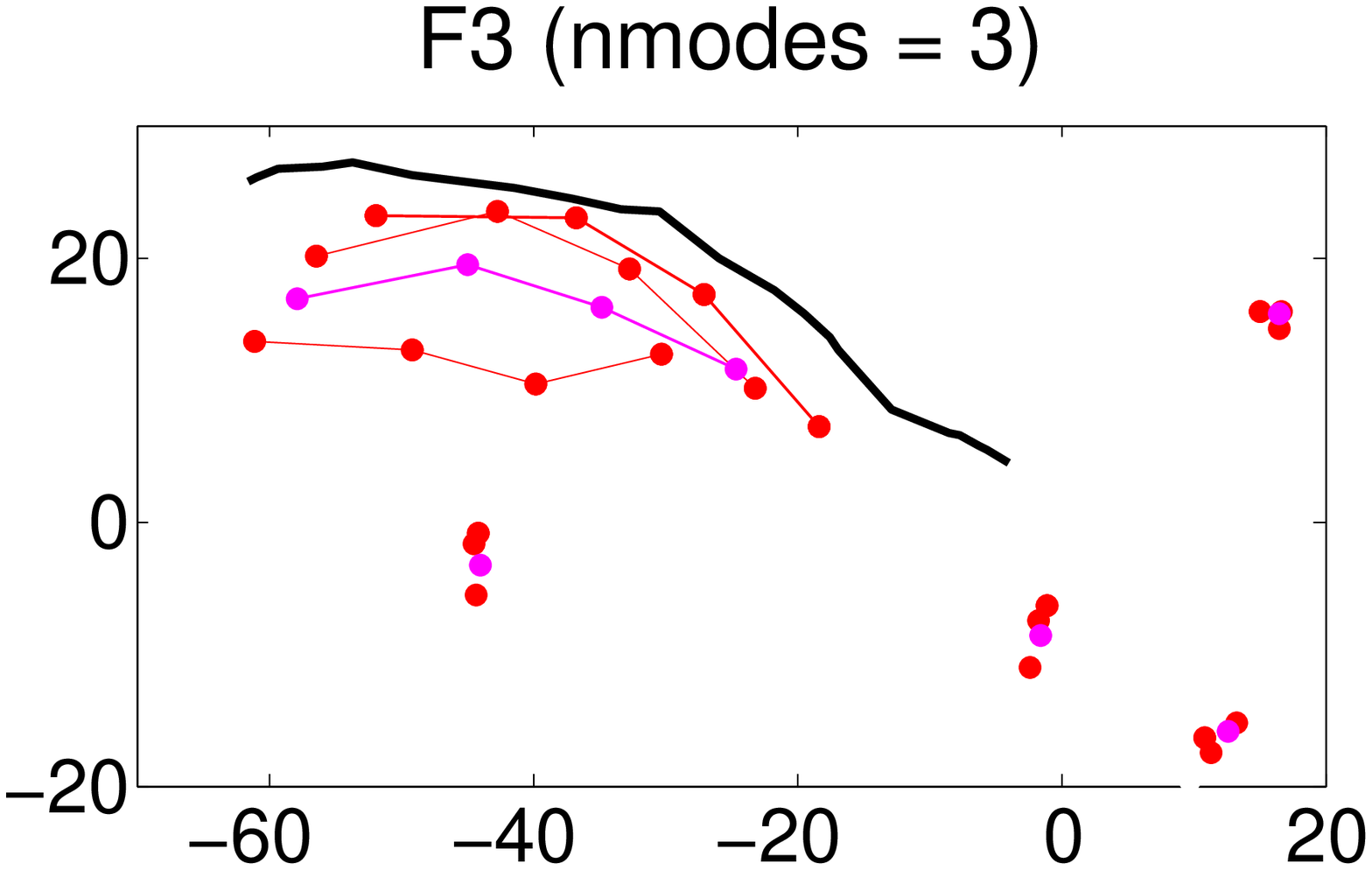}}}
    \end{tabular}
    \caption{\emph{Left panels}: illustration for a 2D robot arm of using conditional modes to represent multiple inverses. The plots correspond to the joints' angle space $\btheta = (\theta_1,\theta_2)$ and the end-effector workspace $\x = (x_1,x_2)$ where we want to reach the point near $(x_1,x_2) = (0.8,0.5)$ (which is part of the desired workspace trajectory in blue). That point \x\ can be reached by two joint angle configurations, elbow up/down, which correspond to the two modes of the conditional distribution $p(\btheta|\x)$ (whose contours are shown). Note that the mean in \btheta-space (magenta dot) does not correspond to a valid inverse. \emph{Right panel}: profile view of the palate (black outline) and several tongue shapes corresponding to the modes (red) and mean (magenta) of the conditional distribution for a speech sound \textipa{/\*r/} in the utterance ``\underline{r}ag''.}
    \label{f:inversion}
  \end{center}
\end{figure}

Fig.~\ref{f:inversion} illustrates two examples. On the left panels, we learn the inverse kinematics mapping of a robot arm \cite{QinCarreir08a,QinCarreir08b}. The forward kinematics mapping $\x = \g(\btheta)$ is univalued and gives the position \x\ in workspace of the arm's end-effector given the angles \btheta\ at the joints. The inverse mapping $\btheta = \f(\x)$ is multivalued, as shown by the elbow-up and elbow-down configurations. In this case, the conditional distribution $p(\btheta|\x)$ was learned using a particle filter. On the right panel, we learn to map speech sounds to tongue shapes (articulatory inversion) \cite{QinCarreir10c}. The American English \textipa{/\*r/} sound (as in ``\underline{r}ag'' or ``\underline{r}oll'') can be produced by one of two tongue shapes: bunched or retroflex. The figure shows, for a specific \textipa{/\*r/} sound, 3 modes in tongue shape space (two bunched and one retroflex). In both inverse kinematics and articulatory inversion, using the conditional mean (shown in magenta), or equivalently doing univalued least-squares regression, leads to invalid inverses.

\paragraph{Other applications}

Algorithms based on mean-shift have also been applied to video segmentation \cite{Dement02a,Wang_04a,ParisDurand07a}, image denoising and discontinuity preserving smoothing \cite{ComanicMeer02a}, and to object tracking \cite{Comanic_03a,Collin03b}, among other problems, as well as for manifold and surface denoising (as described in section~\ref{ch4.1sec:denoising}).

\section{Conclusion and open problems}
\label{ch4.1sec:concl}

Mean-shift algorithms are based on the general idea that locally averaging data results in moving to higher-density, and therefore more typical, regions. Iterating this can be done in two distinct ways, depending on whether the dataset itself is updated: mode finding (MS), or smoothing (BMS), both of which can be used for clustering, but also for manifold denoising, multivalued regression and other tasks.

In practice, mean-shift algorithms are very attractive because---being based on nonparametric kernel density estimates---they make few assumptions about the data and can model nonconvex clusters. The user need only specify the desired scale of the clustering but not the number of clusters itself. Although computationally slow, mean-shift algorithms can be accelerated and scale to large datasets. They are very popular with low-dimensional data having nonconvex clusters, such as image segmentation. The (Laplacian) $K$-modes algorithms remain nonparametric but still perform well with high-dimensional data and outliers.

Mean-shift is intimately related to kernel density estimates and Gaussian mixtures, and to neighborhood graphs. The graph Laplacian or random-walk matrix \PP\ arises as a fundamental object in clustering and dimensionality reduction that encapsulates geometric structure about a dataset. Although not usually seen this way, MS/BMS algorithms are essentially based on \PP, and they alternate between smoothing the data with \PP\ (a power iteration), with possibly modified eigenvalues, and updating \PP\ itself.

Some directions for future research are as follows. (1) Understanding the geometry of Gaussian mixtures and its modes in high dimensions. (2) Finding a meaningful extension of (blurring) mean-shift to directed graphs. (3) Designing, or learning, random-walk or Laplacian matrices that are optimal for clustering in some sense. (4) Finding a meaningful definition of clusters that is based on ``bumps'' of a kernel density estimate, i.e., distinct regions of high probability rather than modes.

\subsubsection*{Acknowledgments}

I thank Weiran Wang and Marina Meil\u{a} for comments on the paper.

\appendix

\section{Connected-components algorithm}
\label{s:conncomp}

Consider an undirected graph $G = (V,E)$ with vertex set $V$ and edge set $E$. A connected component of $G$ is a maximal subset of $V$ such that every pair of vertices in it can be connected through a path using edges in $E$ \cite{Cormen_09a}. Hence, the vertex set $V$ can be partitioned into connected components. The connected component of a given vertex $v\in V$ can be found by depth-first search (DFS), which recursively follows edges adjacent to $v$ until all vertices reachable from $v$ have been found. This can be repeated for the remaining vertices until all connected components have been found, with a total runtime $\calO(\abs{V} + \abs{E})$.

Connected-components is a clustering algorithm in its own right. To use it, one provides a matrix of distances $d_{nm}$ between every pair of vertices (data points $\x_n$ and $\x_m$) and a threshold $\epsilon > 0$, and defines a graph having as vertices the $N$ data points and as edges $(\x_n,\x_m)$ if $d_{nm} < \epsilon$ for $n,m = 1,\dots,N$ (this is sometimes called an $\epsilon$-ball graph). The connected components of this graph give the clusters.

Connected-components gives poor clustering results unless the clusters are tight and clearly separated. However, it can be reliably used as a postprocessing step with mean-shift algorithms, as described in the main text, to merge points that ideally would be identical. For example, with MS, the final iterates for points that in the limit would converge to the same mode are numerically different from each other (by a small amount if using a sufficiently accurate stopping criterion in the mean-shift iteration). Hence, these iterates form a tight cluster around their mode, which is widely separated from the tight clusters corresponding to other modes.

Naively implemented (fig.~\ref{f:conncomp} left), connected-components would run in $\calO(D N^2)$ time, because to construct the graph we have to threshold all pairwise distances, each of which costs $\calO(D)$. However, in the case of tight, clearly separated clusters, we need not construct the graph explicitly, and it runs in $\calO(DNK)$ if there are $K$ connected components. Let the data satisfy the following ``tight clusters'' assumption: there exists $\epsilon > 0$ that is larger than the diameter of each component (the largest distance between every pair of points in a component) but smaller than the distance between any two components (the smallest distance between two points belonging to different components). Then the connected components can be found incrementally by connecting each point to the representative of its component (fig.~\ref{f:conncomp} right), where the representative of component $k$ is $\c_k$, which is a point in that component. This can be done on the fly, as we process each data point in MS, or given the final set of iterates in BMS.

Some heuristics can further accelerate the computation. 1) The distances can be computed incrementally (dimension by dimension), so that as soon as $\epsilon$ is exceeded we exit the distance calculation. In the tight cluster assumption, this could reduce the cost to $\calO(N(K+D))$, since for distances above $\epsilon$ each individual distance dimension will typically exceed $\epsilon$. 2) In image segmentation, we can scan the pixels in raster order, so that for most pixels their component is the same as for the previous pixel (this will not be true when crossing a cluster boundary, but that happens infrequently). Then, when computing the distance, we always try first the last pixels's component. The average cost for the entire connected-components runtime is then $\calO(N D)$, because most pixels compute a single distance.

The value of $\epsilon$ depends on the problem but it can usually be chosen in a wide interval. For example, in image segmentation with features in pixel units, we can set $\epsilon = 0.5$, since we do not need subpixel accuracy to locate a mode, and modes must be at least one pixel apart to be meaningful. In BMS, we can safely set $\epsilon$ to a smaller value (say, $0.01$), since its cubic convergence rate produces extremely tight clusters quickly.

\begin{figure}[t]
  \begin{center}
    \begin{tabular}{@{}c@{\hspace{0.04\linewidth}}c@{}}
      \begin{minipage}[t]{0.47\linewidth}
        \textbf{A}. Naive connected-components algorithm \\
        \setlength{\fboxsep}{4pt}
        \framebox[\textwidth]{%
          \begin{minipage}[c]{0.95\textwidth}
            Define an $\epsilon$-ball graph:
            \begin{itemize}
            \item vertices $\x_1,\dots,\x_N$
            \item edges $(\x_n,\x_m) \Leftrightarrow d(\x_n,\x_m) < \epsilon$, \\
              $\forall n,m = 1,\dots,N$.
            \end{itemize}
            Apply DFS to this graph.
          \end{minipage}%
        }
      \end{minipage} &
      \begin{minipage}[t]{0.47\linewidth}
        \textbf{B}. Efficient connected-components algorithm \\
        \setlength{\fboxsep}{4pt}
        \framebox[\textwidth]{%
          \begin{minipage}[c]{0.95\textwidth}
            \begin{tabbing}
              m \= m \= m \= m \= m \= \kill
              $K \leftarrow 1,\ \c_1 \leftarrow \x_1$ \` {\scriptsize\textsf{first component}} \\
              \underline{\textbf{for}} $n = 2$ to $N$ \+ \\
              \underline{\textbf{for}} $k = 1$ to $K$ \+ \\
              \underline{\textbf{if}} $d(\x_n,\c_k) < \epsilon$ \+ \` {\scriptsize\textsf{old component}} \\
              assign $\x_n$ to component $k$; \underline{\textbf{break}} \- \- \\
              \underline{\textbf{if}} $\x_n$ was not assigned \+ \\
              $K \leftarrow K+1,\ \c_k \leftarrow \x_n$ \` {\scriptsize\textsf{new component}} \-
            \end{tabbing}
          \end{minipage}%
        }
      \end{minipage}
    \end{tabular}
    \caption{Pseudocode for connected-components, implemented naively in $\calO(D N^2)$ (left) and under the ``tight clusters'' assumption in $\calO(D N K)$ (right), where $K$ is the number of components. In all cases, the input is a dataset $\x_1,\dots,\x_N \in \bbR^D$, a distance function $d(\cdot,\cdot)$ applicable to any pair of points, and a threshold $\epsilon > 0$.}
    \label{f:conncomp}
  \end{center}
\end{figure}


\end{document}

%% file: MACPdef.tex
\newcommand{\C}{\ensuremath{\mathbf{C}}}
\newcommand{\D}{\ensuremath{\mathbf{D}}}

\newcommand{\I}{\ensuremath{\mathbf{I}}}
\newcommand{\J}{\ensuremath{\mathbf{J}}}

\newcommand{\LL}{\ensuremath{\mathbf{L}}}

\newcommand{\N}{\ensuremath{\mathbf{N}}}

\newcommand{\PP}{\ensuremath{\mathbf{P}}}
\newcommand{\Q}{\ensuremath{\mathbf{Q}}}

\newcommand{\W}{\ensuremath{\mathbf{W}}}
\newcommand{\X}{\ensuremath{\mathbf{X}}}

\newcommand{\Z}{\ensuremath{\mathbf{Z}}}

\renewcommand{\c}{\ensuremath{\mathbf{c}}}

\newcommand{\f}{\ensuremath{\mathbf{f}}}
\newcommand{\g}{\ensuremath{\mathbf{g}}}

\newcommand{\uu}{\ensuremath{\mathbf{u}}}
\newcommand{\vv}{\ensuremath{\mathbf{v}}}

\newcommand{\x}{\ensuremath{\mathbf{x}}}
\newcommand{\y}{\ensuremath{\mathbf{y}}}
\newcommand{\z}{\ensuremath{\mathbf{z}}}
\newcommand{\0}{\ensuremath{\mathbf{0}}}
\newcommand{\1}{\ensuremath{\mathbf{1}}}

\newcommand{\btheta}{\ensuremath{\boldsymbol{\theta}}}

\newcommand{\bDelta}{\ensuremath{\boldsymbol{\Delta}}}

\newcommand{\bPi}{\ensuremath{\boldsymbol{\Pi}}}

\newcommand{\bSigma}{\ensuremath{\boldsymbol{\Sigma}}}

\newcommand{\bbR}{\ensuremath{\mathbb{R}}}

\newcommand{\calN}{\ensuremath{\mathcal{N}}}
\newcommand{\calO}{\ensuremath{\mathcal{O}}}

\newcommand{\abs}[1]{\left\lvert#1\right\rvert}
\newcommand{\norm}[1]{\left\lVert#1\right\rVert}

\newcommand{\caja}[4][1]{{%
    \renewcommand{\arraystretch}{#1}%
    \begin{tabular}[#2]{@{}#3@{}}%
      #4%
    \end{tabular}%
    }}

{%
\begin{list}{#1}{
\vspace{-\topsep}
\vspace{-\partopsep}
\setlength{\itemindent}{0cm}
\setlength{\rightmargin}{0cm}
\setlength{\listparindent}{0cm}
\settowidth{\labelwidth}{#1}
\setlength{\leftmargin}{\labelwidth}
\addtolength{\leftmargin}{\labelsep}
\setlength{\itemsep}{0cm}
}%
}%
{%
\end{list}
\vspace{-\topsep}
\vspace{-\partopsep}
}

{\begin{enumerate}%
}%
{\end{enumerate}}

%% file: MACPdef-ams.tex
\newcommand{\diagop}{\operatorname{diag}}
\newcommand{\diag}[1]{\ensuremath{\diagop\left(#1\right)}}
\newcommand{\traceop}{\operatorname{tr}}
\newcommand{\trace}[1]{\ensuremath{\traceop\left(#1\right)}}

\theoremstyle{plain}

\newtheorem*{lemma*}{Lemma}

\newtheorem*{prop*}{Proposition}

\theoremstyle{definition}

\newtheorem*{defn*}{Definition}

\newtheorem*{exmp*}{Example}

\newtheorem*{conj*}{Conjecture}

\theoremstyle{remark}

\newtheorem*{rmk*}{Remark}


%% file: mean-shift-review.bbl
\begin{thebibliography}{95}
\providecommand{\natexlab}[1]{#1}
\providecommand{\url}[1]{\texttt{#1}}
\expandafter\ifx\csname urlstyle\endcsname\relax
  \providecommand{\doi}[1]{doi: #1}\else
  \providecommand{\doi}{doi: \begingroup \urlstyle{rm}\Url}\fi

\bibitem[Afsari(2011)]{Afsari11a}
B.~Afsari.
\newblock Riemannian $l^p$ center of mass: Existence, uniqueness, and
  convexity.
\newblock \emph{Proc. Amer. Math. Soc.}, 139:\penalty0 655--673, 2011.

\bibitem[Andoni and Indyk(2008)]{AndoniIndyk08a}
A.~Andoni and P.~Indyk.
\newblock Near-optimal hashing algorithms for approximate nearest neighbor in
  high dimensions.
\newblock \emph{Comm. ACM}, 51\penalty0 (1):\penalty0 117--122, Jan. 2008.

\bibitem[Babaud et~al.(1986)Babaud, Witkin, Baudin, and Duda]{Babaud_86a}
J.~Babaud, A.~P. Witkin, M.~Baudin, and R.~O. Duda.
\newblock Uniqueness of the {Gaussian} kernel for scale-space filtering.
\newblock \emph{IEEE Trans. Pattern Analysis and Machine Intelligence},
  8\penalty0 (1):\penalty0 26--33, Jan. 1986.

\bibitem[Barash and Comaniciu(2004)]{BarashComanic04a}
D.~Barash and D.~Comaniciu.
\newblock A common framework for nonlinear diffusion, adaptive smoothing,
  bilateral filtering and mean shift.
\newblock \emph{Image and Vision Computing Journal}, 22\penalty0 (1):\penalty0
  73--81, Jan. 2004.

\bibitem[Belkin and Niyogi(2003)]{BelkinNiyogi03b}
M.~Belkin and P.~Niyogi.
\newblock Laplacian eigenmaps for dimensionality reduction and data
  representation.
\newblock \emph{Neural Computation}, 15\penalty0 (6):\penalty0 1373--1396, June
  2003.

\bibitem[Berger(2003)]{Berger03a}
M.~Berger.
\newblock \emph{A Panoramic View of {Riemannian} Geometry}.
\newblock Springer-Verlag, 2003.

\bibitem[Bishop(2006)]{Bishop06a}
C.~M. Bishop.
\newblock \emph{Pattern Recognition and Machine Learning}.
\newblock Springer Series in Information Science and Statistics.
  Springer-Verlag, Berlin, 2006.

\bibitem[Carreira-Perpi{\~n}{\'a}n(2000{\natexlab{a}})]{Carreir00a}
M.~{\'A}. Carreira-Perpi{\~n}{\'a}n.
\newblock Reconstruction of sequential data with probabilistic models and
  continuity constraints.
\newblock In S.~A. Solla, T.~K. Leen, and K.-R. M{\"u}ller, editors,
  \emph{Advances in Neural Information Processing Systems (NIPS)}, volume~12,
  pages 414--420. MIT Press, Cambridge, MA, 2000{\natexlab{a}}.

\bibitem[Carreira-Perpi{\~n}{\'a}n(2000{\natexlab{b}})]{Carreir00b}
M.~{\'A}. Carreira-Perpi{\~n}{\'a}n.
\newblock Mode-finding for mixtures of {Gaussian} distributions.
\newblock \emph{IEEE Trans. Pattern Analysis and Machine Intelligence},
  22\penalty0 (11):\penalty0 1318--1323, Nov. 2000{\natexlab{b}}.

\bibitem[Carreira-Perpi{\~n}{\'a}n(2001)]{Carreir01a}
M.~{\'A}. Carreira-Perpi{\~n}{\'a}n.
\newblock \emph{Continuous Latent Variable Models for Dimensionality Reduction
  and Sequential Data Reconstruction}.
\newblock PhD thesis, Dept. of Computer Science, University of Sheffield, UK,
  2001.

\bibitem[Carreira-Perpi{\~n}{\'a}n(2004)]{Carreir04a}
M.~{\'A}. Carreira-Perpi{\~n}{\'a}n.
\newblock Reconstruction of sequential data with density models.
\newblock arXiv:1109.3248 [cs.LG], Jan.~27 2004.

\bibitem[Carreira-Perpi{\~n}{\'a}n(2006{\natexlab{a}})]{Carreir06a}
M.~{\'A}. Carreira-Perpi{\~n}{\'a}n.
\newblock Acceleration strategies for {Gaussian} mean-shift image segmentation.
\newblock In C.~Schmid, S.~Soatto, and C.~Tomasi, editors, \emph{Proc. of the
  2006 IEEE Computer Society Conf. Computer Vision and Pattern Recognition
  (CVPR'06)}, pages 1160--1167, New York, NY, June~17--22 2006{\natexlab{a}}.

\bibitem[Carreira-Perpi{\~n}{\'a}n(2006{\natexlab{b}})]{Carreir06b}
M.~{\'A}. Carreira-Perpi{\~n}{\'a}n.
\newblock Fast nonparametric clustering with {Gaussian} blurring mean-shift.
\newblock In W.~W. Cohen and A.~Moore, editors, \emph{Proc. of the 23rd Int.
  Conf. Machine Learning (ICML'06)}, pages 153--160, Pittsburgh, PA,
  June~25--29 2006{\natexlab{b}}.

\bibitem[Carreira-Perpi{\~n}{\'a}n(2007)]{Carreir07a}
M.~{\'A}. Carreira-Perpi{\~n}{\'a}n.
\newblock Gaussian mean shift is an {EM} algorithm.
\newblock \emph{IEEE Trans. Pattern Analysis and Machine Intelligence},
  29\penalty0 (5):\penalty0 767--776, May 2007.

\bibitem[Carreira-Perpi{\~n}{\'a}n(2008)]{Carreir08a}
M.~{\'A}. Carreira-Perpi{\~n}{\'a}n.
\newblock Generalised blurring mean-shift algorithms for nonparametric
  clustering.
\newblock In \emph{Proc. of the 2008 IEEE Computer Society Conf. Computer
  Vision and Pattern Recognition (CVPR'08)}, Anchorage, AK, June~23--28 2008.

\bibitem[Carreira-Perpi{\~n}{\'a}n(2010)]{Carreir10a}
M.~{\'A}. Carreira-Perpi{\~n}{\'a}n.
\newblock The elastic embedding algorithm for dimensionality reduction.
\newblock In J.~F{\"u}rnkranz and T.~Joachims, editors, \emph{Proc. of the 27th
  Int. Conf. Machine Learning (ICML 2010)}, pages 167--174, Haifa, Israel,
  June~21--25 2010.

\bibitem[Carreira-Perpi{\~n}{\'a}n and Wang(2013)]{CarreirWang13a}
M.~{\'A}. Carreira-Perpi{\~n}{\'a}n and W.~Wang.
\newblock The {$K$}-modes algorithm for clustering.
\newblock arXiv:1304.6478 [cs.LG], Apr.~23 2013.

\bibitem[Carreira-Perpi{\~n}{\'a}n and
  Williams(2003{\natexlab{a}})]{CarreirWilliam03a}
M.~{\'A}. Carreira-Perpi{\~n}{\'a}n and C.~K.~I. Williams.
\newblock On the number of modes of a {Gaussian} mixture.
\newblock Technical Report \mbox{EDI--INF--RR--0159}, School of Informatics,
  University of Edinburgh, Feb. 2003{\natexlab{a}}.

\bibitem[Carreira-Perpi{\~n}{\'a}n and
  Williams(2003{\natexlab{b}})]{CarreirWilliam03b}
M.~{\'A}. Carreira-Perpi{\~n}{\'a}n and C.~K.~I. Williams.
\newblock On the number of modes of a {Gaussian} mixture.
\newblock In L.~Griffin and M.~Lillholm, editors, \emph{Scale Space Methods in
  Computer Vision}, number 2695 in Lecture Notes in Computer Science, pages
  625--640. Springer-Verlag, 2003{\natexlab{b}}.

\bibitem[Carreira-Perpi{\~n}{\'a}n and
  Williams(2003{\natexlab{c}})]{CarreirWilliam03c}
M.~{\'A}. Carreira-Perpi{\~n}{\'a}n and C.~K.~I. Williams.
\newblock An isotropic {Gaussian} mixture can have more modes than components.
\newblock Technical Report \mbox{EDI--INF--RR--0185}, School of Informatics,
  University of Edinburgh, Dec. 2003{\natexlab{c}}.

\bibitem[Chakravarthy and Ghosh(1996)]{ChakravGhosh96a}
S.~V. Chakravarthy and J.~Ghosh.
\newblock Scale-based clustering using the radial basis function network.
\newblock \emph{IEEE Trans. Neural Networks}, 7\penalty0 (5):\penalty0
  1250--1261, Sept. 1996.

\bibitem[Chaturvedi et~al.(2001)Chaturvedi, Green, and Caroll]{Chatur_01a}
A.~Chaturvedi, P.~E. Green, and J.~D. Caroll.
\newblock $k$-modes clustering.
\newblock \emph{Journal of Classification}, 18\penalty0 (1):\penalty0 35--55,
  Jan. 2001.

\bibitem[Chazal et~al.(2013)Chazal, Guibas, Oudot, and Skraba]{Chazal_13a}
F.~Chazal, L.~J. Guibas, S.~Y. Oudot, and P.~Skraba.
\newblock Persistence-based clustering in {Riemannian} manifolds.
\newblock \emph{Journal of the ACM}, 60\penalty0 (6):\penalty0 41, Nov. 2013.

\bibitem[Cheng(1995)]{Cheng95a}
Y.~Cheng.
\newblock Mean shift, mode seeking, and clustering.
\newblock \emph{IEEE Trans. Pattern Analysis and Machine Intelligence},
  17\penalty0 (8):\penalty0 790--799, Aug. 1995.

\bibitem[Chung(1997)]{Chung97a}
F.~R.~K. Chung.
\newblock \emph{Spectral Graph Theory}.
\newblock Number~92 in CBMS Regional Conference Series in Mathematics. American
  Mathematical Society, Providence, RI, 1997.

\bibitem[Ciollaro et~al.(2014)Ciollaro, Genovese, Lei, and
  Wasserman]{Ciollar_14a}
M.~Ciollaro, C.~Genovese, J.~Lei, and L.~Wasserman.
\newblock The functional mean-shift algorithm for mode hunting and clustering
  in infinite dimensions.
\newblock arXiv:1408.1187 [stat.ME], Aug.~6 2014.

\bibitem[Collins(2003)]{Collin03b}
R.~T. Collins.
\newblock Mean-shift blob tracking through scale space.
\newblock In \emph{Proc. of the 2003 IEEE Computer Society Conf. Computer
  Vision and Pattern Recognition (CVPR'03)}, pages 234--240, Madison,
  Wisconsin, June~16--22 2003.

\bibitem[Comaniciu(2003)]{Comanic03a}
D.~Comaniciu.
\newblock An algorithm for data-driven bandwidth selection.
\newblock \emph{IEEE Trans. Pattern Analysis and Machine Intelligence},
  25\penalty0 (2):\penalty0 281--288, Feb. 2003.

\bibitem[Comaniciu and Meer(2002)]{ComanicMeer02a}
D.~Comaniciu and P.~Meer.
\newblock Mean shift: A robust approach toward feature space analysis.
\newblock \emph{IEEE Trans. Pattern Analysis and Machine Intelligence},
  24\penalty0 (5):\penalty0 603--619, May 2002.

\bibitem[Comaniciu et~al.(2003)Comaniciu, Ramesh, and Meer]{Comanic_03a}
D.~Comaniciu, V.~Ramesh, and P.~Meer.
\newblock Kernel-based object tracking.
\newblock \emph{IEEE Trans. Pattern Analysis and Machine Intelligence},
  25\penalty0 (5):\penalty0 564--577, May 2003.

\bibitem[Cormen et~al.(2009)Cormen, Leiserson, Rivest, and Stein]{Cormen_09a}
T.~H. Cormen, C.~E. Leiserson, R.~L. Rivest, and C.~Stein.
\newblock \emph{Introduction to Algorithms}.
\newblock MIT Press, Cambridge, MA, third edition, 2009.

\bibitem[{DeMenthon}(2002)]{Dement02a}
D.~{DeMenthon}.
\newblock Spatio-temporal segmentation of video by hierarchical mean shift
  analysis.
\newblock In \emph{Statistical Methods in Video Processing Workshop (SMVP
  2002)}, Copenhagen, Denmark, June~1--2 2002.

\bibitem[Desbrun et~al.(1999)Desbrun, Meyer, Schr{\"o}der, and
  Barr]{Desbrun_99a}
M.~Desbrun, M.~Meyer, P.~Schr{\"o}der, and A.~H. Barr.
\newblock Implicit fairing of irregular meshes using diffusion and curvature
  flow.
\newblock In W.~Waggenspack, editor, \emph{Proc. of the 26th Annual Conference
  on Computer Graphics and Interactive Techniques (SIGGRAPH 1999)}, pages
  317--324, Los Angeles, CA, Aug.~8--13 1999.

\bibitem[Edelsbrunner et~al.(2002)Edelsbrunner, Letscher, and
  Zomorodian]{Edelsb_02a}
H.~Edelsbrunner, D.~Letscher, and A.~Zomorodian.
\newblock Topological persistence and simplification.
\newblock \emph{Discrete \& Computational Geometry}, 28\penalty0 (4):\penalty0
  511--533, Nov. 2002.

\bibitem[Fashing and Tomasi(2005)]{FashinTomasi05a}
M.~Fashing and C.~Tomasi.
\newblock Mean shift is a bound optimization.
\newblock \emph{IEEE Trans. Pattern Analysis and Machine Intelligence},
  27\penalty0 (3):\penalty0 471--474, Mar. 2005.

\bibitem[Forsyth and Ponce(2003)]{ForsytPonce03a}
D.~A. Forsyth and J.~Ponce.
\newblock \emph{Computer Vision. {A} Modern Approach}.
\newblock Prentice-Hall, Upper Saddle River, N.J., 2003.

\bibitem[Fukunaga and Hostetler(1975)]{FukunagHostet75a}
K.~Fukunaga and L.~D. Hostetler.
\newblock The estimation of the gradient of a density function, with
  application in pattern recognition.
\newblock \emph{IEEE Trans. Information Theory}, IT--21\penalty0 (1):\penalty0
  32--40, Jan. 1975.

\bibitem[Georgescu et~al.(2003)Georgescu, Shimshoni, and Meer]{Georges_03a}
B.~Georgescu, I.~Shimshoni, and P.~Meer.
\newblock Mean shift based clustering in high dimensions: A texture
  classification example.
\newblock In \emph{Proc. 9th Int. Conf. Computer Vision (ICCV'03)}, pages
  456--463, Nice, France, Oct.~14--17 2003.

\bibitem[Golub and van Loan(1996)]{GolubLoan96a}
G.~H. Golub and C.~F. van Loan.
\newblock \emph{Matrix Computations}.
\newblock Johns Hopkins University Press, Baltimore, third edition, 1996.

\bibitem[Greengard and Strain(1991)]{GreengStrain91a}
L.~Greengard and J.~Strain.
\newblock The fast {Gauss} transform.
\newblock \emph{SIAM J. Sci. Stat. Comput.}, 12\penalty0 (1):\penalty0 79--94,
  Jan. 1991.

\bibitem[Grillenzoni(2013)]{Grillen13a}
C.~Grillenzoni.
\newblock Detection of tectonic faults by spatial clustering of earthquake
  hypocenters.
\newblock \emph{Spatial Statistics}, 7:\penalty0 62--78, Feb. 2013.

\bibitem[Hall and Minnotte(2002)]{HallMinnot02a}
P.~Hall and M.~C. Minnotte.
\newblock High order data sharpening for density estimation.
\newblock \emph{Journal of the Royal Statistical Society, B}, 64\penalty0
  (1):\penalty0 141--157, Jan. 2002.

\bibitem[Hall et~al.(2006)Hall, M{\"u}ller, and Wu]{Hall_06a}
P.~Hall, H.-G. M{\"u}ller, and P.-S. Wu.
\newblock Real-time density and mode estimation with application to
  time-dynamic mode tracking.
\newblock \emph{Journal of Computational and Graphical Statistics}, 15\penalty0
  (1):\penalty0 82--100, Mar. 2006.

\bibitem[Hein and Maier(2007)]{HeinMaier07a}
M.~Hein and M.~Maier.
\newblock Manifold denoising.
\newblock In B.~Sch{\"o}lkopf, J.~Platt, and T.~Hofmann, editors,
  \emph{Advances in Neural Information Processing Systems (NIPS)}, volume~19,
  pages 561--568. MIT Press, Cambridge, MA, 2007.

\bibitem[Hinton and Roweis(2003)]{HintonRoweis03a}
G.~Hinton and S.~T. Roweis.
\newblock Stochastic neighbor embedding.
\newblock In S.~Becker, S.~Thrun, and K.~Obermayer, editors, \emph{Advances in
  Neural Information Processing Systems (NIPS)}, volume~15, pages 857--864. MIT
  Press, Cambridge, MA, 2003.

\bibitem[Ho-Le(1988)]{Ho-Le88a}
K.~Ho-Le.
\newblock Finite element mesh generation methods: A review and classification.
\newblock \emph{Computer-Aided Design}, 20\penalty0 (1):\penalty0 27--38,
  Jan.--Feb. 1988.

\bibitem[Horn and Johnson(1986)]{HornJohnson86a}
R.~A. Horn and C.~R. Johnson.
\newblock \emph{Matrix Analysis}.
\newblock Cambridge University Press, Cambridge, U.K., 1986.

\bibitem[Huang(1998)]{Huang98a}
Z.~Huang.
\newblock Extensions to the $k$-means algorithm for clustering large data sets
  with categorical values.
\newblock \emph{Data Mining and Knowledge Discovery}, 2\penalty0 (2):\penalty0
  283--304, Sept. 1998.

\bibitem[Hyv{\"a}rinen(2005)]{Hyvaer05a}
A.~Hyv{\"a}rinen.
\newblock Estimation of non-normalized statistical models by score matching.
\newblock \emph{J. Machine Learning Research}, 6:\penalty0 695--708, Apr. 2005.

\bibitem[Koenderink(1984)]{Koender84a}
J.~J. Koenderink.
\newblock The structure of images.
\newblock \emph{Biol. Cybern.}, 50\penalty0 (5):\penalty0 363--370, Aug. 1984.

\bibitem[Koontz et~al.(1976)Koontz, Narendra, and Fukunaga]{Koontz_76a}
W.~L.~G. Koontz, P.~M. Narendra, and K.~Fukunaga.
\newblock Graph-theoretic approach to nonparametric cluster-analysis.
\newblock \emph{IEEE Trans. Computers}, C--25\penalty0 (9):\penalty0 936--944,
  1976.

\bibitem[{LeCun} et~al.(1998){LeCun}, Bottou, Bengio, and Haffner]{Lecun_98a}
Y.~{LeCun}, L.~Bottou, Y.~Bengio, and P.~Haffner.
\newblock Gradient-based learning applied to document recognition.
\newblock \emph{Proc. IEEE}, 86\penalty0 (11):\penalty0 2278--2324, Nov. 1998.

\bibitem[Lifshitz and Pizer(1990)]{LifshitPizer90a}
L.~M. Lifshitz and S.~M. Pizer.
\newblock A multiresolution hierarchical approach to image segmentation based
  on intensity extrema.
\newblock \emph{IEEE Trans. Pattern Analysis and Machine Intelligence},
  12\penalty0 (6):\penalty0 529--540, June 1990.

\bibitem[Lindeberg(1994)]{Lindeb94a}
T.~Lindeberg.
\newblock \emph{Scale-Space Theory in Computer Vision}.
\newblock Kluwer Academic Publishers Group, Dordrecht, The Netherlands, 1994.

\bibitem[{McLachlan} and Krishnan(1997)]{MclachKrishn97a}
G.~J. {McLachlan} and T.~Krishnan.
\newblock \emph{The {EM} Algorithm and Extensions}.
\newblock Wiley Series in Probability and Mathematical Statistics. John Wiley
  \& Sons, New York, 1997.

\bibitem[{McLachlan} and Krishnan(2008)]{MclachKrishn08a}
G.~J. {McLachlan} and T.~Krishnan.
\newblock \emph{The {EM} Algorithm and Extensions}.
\newblock Wiley Series in Probability and Mathematical Statistics. John Wiley
  \& Sons, second edition, 2008.

\bibitem[Meil\u{a} and Bao(2010)]{MeilaBao10a}
M.~Meil\u{a} and L.~Bao.
\newblock An exponential model for infinite rankings.
\newblock \emph{J. Machine Learning Research}, 11:\penalty0 3481--3518, Dec.
  2010.

\bibitem[Meil\u{a} and Shi(2001)]{MeilaShi01a}
M.~Meil\u{a} and J.~Shi.
\newblock Learning segmentation by random walks.
\newblock In T.~K. Leen, T.~G. Dietterich, and V.~Tresp, editors,
  \emph{Advances in Neural Information Processing Systems (NIPS)}, volume~13,
  pages 873--879. MIT Press, Cambridge, MA, 2001.

\bibitem[Minnotte and Scott(1993)]{MinnotScott93a}
M.~C. Minnotte and D.~W. Scott.
\newblock The mode tree: A tool for visualization of nonparametric density
  features.
\newblock \emph{Journal of Computational and Graphical Statistics}, 2\penalty0
  (1):\penalty0 51--68, Mar. 1993.

\bibitem[Minnotte et~al.(1998)Minnotte, Marchette, and Wegman]{Minnot_98a}
M.~C. Minnotte, D.~J. Marchette, and E.~J. Wegman.
\newblock The bumpy road to the mode forest.
\newblock \emph{Journal of Computational and Graphical Statistics}, 7\penalty0
  (2):\penalty0 239--251, June 1998.

\bibitem[Nocedal and Wright(2006)]{NocedalWright06a}
J.~Nocedal and S.~J. Wright.
\newblock \emph{Numerical Optimization}.
\newblock Springer Series in Operations Research and Financial Engineering.
  Springer-Verlag, New York, second edition, 2006.

\bibitem[Paris and Durand(2007)]{ParisDurand07a}
S.~Paris and F.~Durand.
\newblock A topological approach to hierarchical segmentation using mean shift.
\newblock In \emph{Proc. of the 2007 IEEE Computer Society Conf. Computer
  Vision and Pattern Recognition (CVPR'07)}, Minneapolis, MN, June~18--23 2007.

\bibitem[Paris et~al.(2008)Paris, Kornprobst, Tumblin, and Durand]{Paris_08a}
S.~Paris, P.~Kornprobst, J.~Tumblin, and F.~Durand.
\newblock Bilateral filtering: Theory and applications.
\newblock \emph{Foundations and Trends in Computer Graphics and Vision},
  4\penalty0 (1):\penalty0 1--73, 2008.

\bibitem[Pennec et~al.(2006)Pennec, Fillard, and Ayache]{Pennec_06a}
X.~Pennec, P.~Fillard, and N.~Ayache.
\newblock A riemannian framework for tensor computing.
\newblock \emph{Int. J. Computer Vision}, 66\penalty0 (1):\penalty0 41--66,
  Jan. 2006.

\bibitem[Qin and Carreira-Perpi{\~n}{\'a}n(2008{\natexlab{a}})]{QinCarreir08a}
C.~Qin and M.~{\'A}. Carreira-Perpi{\~n}{\'a}n.
\newblock Trajectory inverse kinematics by conditional density modes.
\newblock In \emph{Proc. of the 2008 IEEE Int. Conf. Robotics and Automation
  (ICRA'08)}, pages 1979--1986, Pasadena, California, May~19--23
  2008{\natexlab{a}}.

\bibitem[Qin and Carreira-Perpi{\~n}{\'a}n(2008{\natexlab{b}})]{QinCarreir08b}
C.~Qin and M.~{\'A}. Carreira-Perpi{\~n}{\'a}n.
\newblock Trajectory inverse kinematics by nonlinear, nongaussian tracking.
\newblock In \emph{Proc. of the IEEE Int. Conf. Acoustics, Speech and Sig.
  Proc. (ICASSP'08)}, pages 2057--2060, Las Vegas, Nevada, Mar.~31 -- Apr.~4
  2008{\natexlab{b}}.

\bibitem[Qin and Carreira-Perpi{\~n}{\'a}n(2010)]{QinCarreir10c}
C.~Qin and M.~{\'A}. Carreira-Perpi{\~n}{\'a}n.
\newblock Articulatory inversion of {American} {English} /r/ by conditional
  density modes.
\newblock In T.~Kobayashi, K.~Hirose, and S.~Nakamura, editors, \emph{Proc. of
  Interspeech'10}, pages 1998--2001, Makuhari, Japan, Sept.~26--30 2010.

\bibitem[Roberts(1997)]{Robert97a}
S.~J. Roberts.
\newblock Parametric and non-parametric unsupervised cluster analysis.
\newblock \emph{Pattern Recognition}, 30\penalty0 (2):\penalty0 261--272, Feb.
  1997.

\bibitem[Rose(1998)]{Rose98a}
K.~Rose.
\newblock Deterministic annealing for clustering, compression, classification,
  regression, and related optimization problems.
\newblock \emph{Proc. IEEE}, 86\penalty0 (11):\penalty0 2210--2239, Nov. 1998.

\bibitem[Samet(2006)]{Samet06a}
H.~Samet.
\newblock \emph{Foundations of Multidimensional and Metric Data Structures}.
\newblock Morgan Kaufmann, 2006.

\bibitem[Sasaki et~al.(2014)Sasaki, Hyv{\"a}rinen, and Sugiyama]{Sasaki_14a}
H.~Sasaki, A.~Hyv{\"a}rinen, and M.~Sugiyama.
\newblock Clustering via mode seeking by direct estimation of the gradient of a
  log-density.
\newblock In T.~Calders, F.~Esposito, E.~H{\"u}llermeier, and R.~Meo, editors,
  \emph{Proc. of the 25th European Conf. Machine Learning (ECML--14)}, pages
  19--34, Nancy, France, Sept.~15--19 2014.

\bibitem[Sch{\"o}lkopf et~al.(1999)Sch{\"o}lkopf, Mika, Burges, Knirsch,
  M{\"u}ller, R{\"a}tsch, and Smola]{Schoel_99b}
B.~Sch{\"o}lkopf, S.~Mika, C.~J.~C. Burges, P.~Knirsch, K.-R. M{\"u}ller,
  G.~R{\"a}tsch, and A.~Smola.
\newblock Input space vs. feature space in kernel-based methods.
\newblock \emph{IEEE Trans. Neural Networks}, 10\penalty0 (5):\penalty0
  1000--1017, Sept. 1999.

\bibitem[Shamir et~al.(2006)Shamir, Shapira, and Cohen-Or]{Shamir_06a}
A.~Shamir, L.~Shapira, and D.~Cohen-Or.
\newblock Mesh analysis using geodesic mean shift.
\newblock \emph{The Visual Computer}, 22\penalty0 (2):\penalty0 99--108, Feb.
  2006.

\bibitem[Shapira et~al.(2009)Shapira, Avidan, and Shamir]{Shapir_09a}
L.~Shapira, S.~Avidan, and A.~Shamir.
\newblock Mode-detection via median-shift.
\newblock In \emph{Proc. 12th Int. Conf. Computer Vision (ICCV'09)}, pages
  1909--1916, Kyoto, Japan, Sept.~29 -- Oct.~2 2009.

\bibitem[Sheikh et~al.(2007)Sheikh, Khan, and Kanade]{Sheikh_07a}
Y.~A. Sheikh, E.~A. Khan, and T.~Kanade.
\newblock Mode-seeking via medoidshifts.
\newblock In \emph{Proc. 11th Int. Conf. Computer Vision (ICCV'07)}, Rio de
  Janeiro, Brazil, Oct.~14--21 2007.

\bibitem[Shi and Malik(2000)]{ShiMalik00a}
J.~Shi and J.~Malik.
\newblock Normalized cuts and image segmentation.
\newblock \emph{IEEE Trans. Pattern Analysis and Machine Intelligence},
  22\penalty0 (8):\penalty0 888--905, Aug. 2000.

\bibitem[Silverman(1986)]{Silver86a}
B.~W. Silverman.
\newblock \emph{Density Estimation for Statistics and Data Analysis}.
\newblock Number~26 in Monographs on Statistics and Applied Probability.
  Chapman \& Hall, London, New York, 1986.

\bibitem[Subbarao and Meer(2009)]{SubbarMeer09a}
R.~Subbarao and P.~Meer.
\newblock Nonlinear mean shift over {Riemannian} manifolds.
\newblock \emph{Int. J. Computer Vision}, 84\penalty0 (1):\penalty0 1--20, Aug.
  2009.

\bibitem[Taubin(1995)]{Taubin95a}
G.~Taubin.
\newblock A signal processing approach to fair surface design.
\newblock In S.~G. Mair and R.~Cook, editors, \emph{Proc. of the 22nd Annual
  Conference on Computer Graphics and Interactive Techniques (SIGGRAPH 1995)},
  pages 351--358, Los Angeles, CA, Aug.~6--11 1995.

\bibitem[Taubin(2000)]{Taubin00a}
G.~Taubin.
\newblock Geometric signal processing on polygonal meshes.
\newblock In \emph{Eurographics'2000: State of the Art Reports}, 2000.

\bibitem[Thompson and Tapia(1990)]{ThompsTapia90a}
J.~R. Thompson and R.~A. Tapia.
\newblock \emph{Nonparametric Function Estimation, Modeling, and Simulation}.
\newblock Other Titles in Applied Mathematics. SIAM Publ., 1990.

\bibitem[Vladymyrov and Carreira-Perpi{\~n}{\'a}n(2013)]{VladymCarreir13a}
M.~Vladymyrov and M.~{\'A}. Carreira-Perpi{\~n}{\'a}n.
\newblock Entropic affinities: Properties and efficient numerical computation.
\newblock In S.~Dasgupta and D.~{McAllester}, editors, \emph{Proc. of the 30th
  Int. Conf. Machine Learning (ICML 2013)}, pages 477--485, Atlanta, GA,
  June~16--21 2013.

\bibitem[Wand and Jones(1994)]{WandJones94a}
M.~P. Wand and M.~C. Jones.
\newblock \emph{Kernel Smoothing}.
\newblock Number~60 in Monographs on Statistics and Applied Probability.
  Chapman \& Hall, London, New York, 1994.

\bibitem[Wang et~al.(2004)Wang, Xu, Shum, and Cohen]{Wang_04a}
J.~Wang, Y.~Xu, H.-Y. Shum, and M.~F. Cohen.
\newblock Video tooning.
\newblock \emph{ACM Trans. Graphics}, 23\penalty0 (3):\penalty0 574--583, Aug.
  2004.

\bibitem[Wang et~al.(2007)Wang, Lee, Gray, and Rehg]{Wang_07a}
P.~Wang, D.~Lee, A.~Gray, and J.~Rehg.
\newblock Fast mean shift with accurate and stable convergence.
\newblock In M.~Meil\u{a} and X.~Shen, editors, \emph{Proc. of the 11th Int.
  Conf. Artificial Intelligence and Statistics (AISTATS 2007)}, San Juan,
  Puerto Rico, Mar.~21--24 2007.

\bibitem[Wang and Carreira-Perpi{\~n}{\'a}n(2010)]{WangCarreir10a}
W.~Wang and M.~{\'A}. Carreira-Perpi{\~n}{\'a}n.
\newblock Manifold blurring mean shift algorithms for manifold denoising.
\newblock In \emph{Proc. of the 2010 IEEE Computer Society Conf. Computer
  Vision and Pattern Recognition (CVPR'10)}, pages 1759--1766, San Francisco,
  CA, June~13--18 2010.

\bibitem[Wang and Carreira-Perpi{\~n}{\'a}n(2014)]{WangCarreir14c}
W.~Wang and M.~{\'A}. Carreira-Perpi{\~n}{\'a}n.
\newblock The {Laplacian} {$K$}-modes algorithm for clustering.
\newblock arXiv:1406.3895, June~15 2014.

\bibitem[Wang et~al.(2011)Wang, Carreira-Perpi{\~n}{\'a}n, and Lu]{Wang_11b}
W.~Wang, M.~{\'A}. Carreira-Perpi{\~n}{\'a}n, and Z.~Lu.
\newblock A denoising view of matrix completion.
\newblock In J.~Shawe-Taylor, R.~S. Zemel, P.~Bartlett, F.~Pereira, and K.~Q.
  Weinberger, editors, \emph{Advances in Neural Information Processing Systems
  (NIPS)}, volume~24, pages 334--342. MIT Press, Cambridge, MA, 2011.

\bibitem[Weickert(1998)]{Weicker98a}
J.~Weickert.
\newblock \emph{Anisotropic Diffusion in Image Processing}.
\newblock ECMI Series. Teubner, Stuttgart, Leipzig, 1998.

\bibitem[Wilson and Spann(1990)]{WilsonSpann90a}
R.~Wilson and M.~Spann.
\newblock A new approach to clustering.
\newblock \emph{Pattern Recognition}, 23\penalty0 (12):\penalty0 1413--1425,
  1990.

\bibitem[Witkin(1983)]{Witkin83a}
A.~P. Witkin.
\newblock Scale-space filtering.
\newblock In \emph{Proc. of the 8th Int. Joint Conf. Artificial Intelligence
  (IJCAI'83)}, pages 1019--1022, Karlsruhe, Germany, Aug. 1983.

\bibitem[Wong(1993)]{Wong93a}
Y.~Wong.
\newblock Clustering data by melting.
\newblock \emph{Neural Computation}, 5\penalty0 (1):\penalty0 89--104, Jan.
  1993.

\bibitem[Yang et~al.(2003)Yang, Duraiswami, Gumerov, and Davis]{Yang_03a}
C.~Yang, R.~Duraiswami, N.~A. Gumerov, and L.~Davis.
\newblock Improved fast {Gauss} transform and efficient kernel density
  estimation.
\newblock In \emph{Proc. 9th Int. Conf. Computer Vision (ICCV'03)}, pages
  464--471, Nice, France, Oct.~14--17 2003.

\bibitem[Yuan et~al.(2010)Yuan, Hu, and He]{Yuan_10a}
X.~Yuan, B.-G. Hu, and R.~He.
\newblock Agglomerative mean-shift clustering.
\newblock \emph{IEEE Trans. Knowledge and Data Engineering}, 24\penalty0
  (2):\penalty0 209--219, Feb. 2010.

\bibitem[Yuille and Poggio(1986)]{YuillePoggio86a}
A.~L. Yuille and T.~A. Poggio.
\newblock Scaling theorems for zero crossings.
\newblock \emph{IEEE Trans. Pattern Analysis and Machine Intelligence},
  8\penalty0 (1):\penalty0 15--25, Jan. 1986.

\end{thebibliography}
